%% file: mive.tex
\documentclass[review]{elsarticle}

\usepackage{lineno,hyperref}
\modulolinenumbers[5]

\usepackage{textcomp}

\makeatletter
\def\ps@pprintTitle{%
 \let\@oddhead\@empty
 \let\@evenhead\@empty
 \def\@oddfoot{\vbox{\vskip 1pc\hbox{\scriptsize\it
 Published in: Neural Networks (Elsevier), Volume 126, June 2020, Pages 275-299.}
 \vskip -1.5pc
 \hbox{\scriptsize  \it DOI: \url{https://doi.org/10.1016/j.neunet.2020.03.013}}
 \vskip -1.3pc
 \hbox{\tiny \it
 {\textcopyright}  2020. This manuscript version is made available under the CC-BY-NC-ND 4.0
 license \url{http://creativecommons.org/licenses/by-nc-nd/4.0/}
 }
 }%
 \let\@evenfoot\@oddfoot}}
 \makeatother

\usepackage{amsmath}
\usepackage{amssymb}
\usepackage{mathrsfs}
\usepackage{times}

\usepackage{rotating}
\usepackage{bbm}
\usepackage{latexsym}
\usepackage{url}

\usepackage{graphicx}
\usepackage{mathrsfs}
\usepackage{appendix}

\usepackage{array}
\usepackage{multirow}
\usepackage{tabu}
\usepackage{booktabs}       
\usepackage[export]{adjustbox}

\usepackage{amsthm}

\usepackage{color}
\newcommand{\bbR}{{\mathbb{R}}}
\def\feat{\Phi}

\def\dts{\mathinner{\ldotp\ldotp}} 
\def\dash---{\thinspace---\hskip.16667em\relax} 
\let\del=\partial
\def\tw{h} 

\def\(#1){[\hbox{$\mkern1mu\thickmuskip=\thinmuskip#1\mkern1mu$}]} 
\def\R{{\bf R}}
\def\maxlim{\mathop{\rm max\,lim}}
\def\Vec{\mathop{\rm vec}}
\def\Tr{\mathop{\rm Tr}}
\def\rev#1{#1}

\newtheorem{theorem}{Theorem}
\newtheorem{lemma}{Lemma}
\newtheorem{proposition}{Proposition}

\begin{document}

\begin{frontmatter}

\title{Learning Visual Features Under Motion Invariance}



\author[mysecondaryaddress]{Alessandro Betti\corref{mycorrespondingauthor}}
\ead{alessandro.betti@unifi.it}

\author[mysecondaryaddress]{Marco Gori}
\ead{marco@diism.unisi.it}

\author[mysecondaryaddress]{Stefano Melacci}
\ead{mela@diism.unisi.it}



\cortext[mycorrespondingauthor]{Corresponding author}

\address[mysecondaryaddress]{Department of Information Engineering and Mathermatics, University of Siena}

\begin{abstract}
Humans are continuously exposed to a stream of visual data with a 
natural temporal structure.  
However, most successful computer vision algorithms work at image level,
completely discarding the precious information carried by motion.
In this paper, we claim that processing visual streams naturally leads to formulate the motion
invariance principle, which enables the construction of a new
theory of learning that originates from variational principles, just 
like in physics. 
Such principled approach is well
suited for a discussion on a number of interesting questions
that arise in vision, and it offers a well-posed computational scheme
for the discovery of convolutional filters over the
retina.
Differently from traditional convolutional networks, which need massive supervision,
the proposed theory offers a truly new scenario for the  unsupervised processing 
of  video signals, where features are extracted in a multi-layer architecture with
motion invariance. 
While the theory enables 
the implementation of novel computer vision systems, 
it also sheds light on the role of information-based
principles to drive possible  biological solutions.
\end{abstract}


\end{frontmatter}


\section{Introduction}
\input intro.tex

\section{Inquiring about visual features}
\label{inq-vis}
\input inq.tex

\section{Main results}
\label{main-res}

\input res.tex

\section{Driving principles}
\label{vis-prin}
\input princ.tex

\section{Laws of visual features}
\label{Cognitive-action-section}
\input continuum.tex

\section{Neural interpretation in the retina}
\label{DiscreteRetina}
\input discrete.tex

\section{Visual features in the light of the theory}
\label{light-vis}
\input lig.tex

\section{Experiments}
\label{ExpResults}
\input exp.tex

\section*{Conclusion}
\label{ConcSec}
\input conc.tex

\subsection*{Acknowledgments}
The initial intuition on the approach used in this paper comes from a few very fruitful interactions with Tomaso Poggio. We thank Marco Lippi, Marco Maggini, Giovanni Bellettini for constructive discussions on different issues raised in the paper.

This work was partly supported by the PRIN 2017 project RexLearn, funded by the Italian Ministry of Education, University and Research (grant no. 2017TWNMH2).

\begin{appendices}
\section{Variation of the motion invariance term}
\input motion

\section{Gaussian Green Functions}
\input gauss

\section{Vectorization}
\input vect
\end{appendices}

\section*{References}

\bibliography{corr,nn}

\end{document}

%% file: intro.tex
For many years, the pioneering work on  vision by David Marr~\citep{Marr82}, 
has evolved without a systematic exploration of foundations in machine learning. When the target is moved to unrestricted visual environments and the emphasis is shifted from huge labelled databases to a human-like protocol of interaction, we need to go beyond the current peaceful interlude that we are experimenting in vision and machine learning. A fundamental question a good theory is expected to answer is why children can learn to recognize objects and actions from a few supervised examples, whereas nowadays supervised learning approaches strive to achieve this task. In particular, why are they so thirsty for supervised examples? This fundamental difference seems to be deeply rooted in the different communication protocol at the basis of the acquisition of visual skills in children and machines. 

So far, the semantic labeling of pixels of a given video stream has been mostly carried out at frame level. This seems to be the natural outcome of well-established pattern recognition methods working on images, which have given rise to nowadays emphasis on collecting big labelled image databases (e.g.~\cite{imagenet_cvpr09}) 
with the purpose of devising and testing challenging machine learning algorithms. While this framework is the one in which most of nowadays state-of-the art object recognition approaches have been developing, we argue that there are strong arguments to start exploring the more natural visual interaction that animals experiment in their own environment.

This leads to process a video signal instead of image collections, that naturally leads to a paradigm-shift
in the associated processes of learning to see. The idea of shifting to video is very much 
related to the growing interest of {\em learning in the wild} that has been explored in the
last few years\footnote{See e.g.~\url{https://sites.google.com/site/wildml2017icml/}.}.
The learning processes that take place in this kind of environments has a different
nature with respect to those that are typically considered in machine learning. 
Learning convolutional nets on ImageNet typically consists of updating the weights
from the processing of temporally unrelated images, whereas a video carries out
information when we pass through contiguous frames  by smooth changes. 
While ImageNet is a collection of unrelated images, a video supports information
only when motion is involved. In presence of fixed images that last for awhile, the corresponding
stream of equal frames basically supports only the information of a single image.
As a consequence,  visual environments diffuse information only when
motion is involved. There is no transition from one image to the next one---like in
ImageNet--- but, as time goes by, the information is only carried out by motion. Once we 
deeply capture this fundamental feature of visual environments, we early realize that
we need a different theory of machine learning that must deal with video instead of
a collection of independent images anymore.

A crucial problem that has been recognized by Poggio and Anselmi~\cite{Poggio:2016:VCD} is the need to incorporate visual invariances
into deep nets that go beyond simple translation invariance that is currently characterizing
convolutional networks. They propose an elegant mathematical framework 
on visual invariance and enlighten some intriguing neurobiological connections. 
Overall, the ambition of extracting distinctive features from vision poses a challenging task. 
While we are typically concerned with feature extraction that is independent of classic geometric transformation, it looks like we are still missing the fantastic human skill of capturing, for example, distinctive features to recognize ironed and rumpled shirts. There is no apparent difficulty to recognize shirts by keeping the recognition coherence in case we roll up the sleeves, or we simply curl them up into a ball for the laundry basket. Of course, there are neither rigid transformations, like translations and rotation, nor scale maps, that transforms an ironed shirt into the same shirt thrown into the laundry basket. 
Is there any natural invariance?

In this paper, we claim that
motion invariance is in fact the only invariance that we need.
Translation, rotation, and scale invariance, that have been the subject of many studies \cite{Lowe:2004}, are in
fact examples of invariances that can be fully gained whenever we develop the 
ability to detect features that are invariant under motion. 
Consider the simple example of your inch that moves closer and closer to your eyes. Any of its representing features 
that is motion invariant will also be scale invariant. 
Clearly, translation, rotation, and complex deformation invariances  derive from
motion invariance. Humans life always experiments motion, so as the gained visual
invariances naturally arise from motion invariance. Animals with foveal eyes also
move quickly to focus  attention on informative areas of the retina, which means that they
continually experiment motion. Hence, also in case of fixed images, 
conjugate, vergence, saccadic, smooth pursuit, and vestibulo-ocular movements lead to acquire visual information from relative motion.  We claim that the 
production of such a continuous visual stream naturally drives
the extraction of feature that are supposed to be useful for object and action recognition.
The enforcement of this consistency condition creates a mine of visual data
during animal life. Interestingly, the same can happen for machines. 
Of course, we need to compute the optical flow at pixel level so as to enforce
the consistency of all the extracted features. Early studies 
on this problem~\cite{HornAI1981}, along with recent related 
improvements (see e.g.~\cite{Baker:2011})
suggests to determine the velocity field by enforcing brightness invariance. 
As the optical flow is gained, it is used to enforce motion consistency on the
visual features. Interestingly, the theory we propose is quite related to the
variational approach that is used to determine the optical flow in~\cite{HornAI1981}.
In addition to the importance of motion invariance, it is worth mentioning that an effective visual system should also develop 
features that do not follow such invariance. These kind of features can be 
conveniently combined with those that are discussed in this paper with the
purpose of carrying out high level visual tasks.

This work is somewhat inspired by the research activity reported in~\cite{DBLP:journals/cviu/GoriLMM16},
where the authors propose the extraction of visual features as a
constraint satisfaction problem, mostly based on information-based principles
and  early ideas on motion invariance. 
However, we incorporate motion invariance in the framework of the principle of least
cognitive action~\cite{DBLP:journals/tcs/BettiG16}, which gives rise  to a
time-variant differential equation, where the Lagrangian coordinates
corresponds with the values of the convolutional filters. 
Unsupervised development of features
from temporally coherent data has already been investigated
in Slow Feature Analysis (SFA)~\cite{wiskott2002slow,wiskott2003slow}, with more recent applications to high-level tasks, such as action recognition~\cite{sun2014dl}. The basic idea is to extract features that are ``slowly varying'' with respect to the ``quickly varying'' input signal.
SFA has been applied in several contexts, and also in the case of motion estimation in video signals.
Other unsupervised learning algorithms 
have been mostly applied to image datasets~\cite{
huang2007unsupervised, NIPS2010_4133}. More recent approaches embraces the idea of
exploiting some notions of motion
coherence with unsupervised learning of image-level features or with object segmentation
\cite{Wang_2015_ICCV, goroshin2015unsupervised, li2016unsupervised,
pathak2017learning}. However to the best of our knowledge,
none of the cited works proposed a learning theory for pixel-level visual features directly formulated in
the time domain and based on motion.

The paper is organized as follows. In the next section we begin discussing
the emerge of visual features along with a number of desiderata for a
a good theory on vision. In Section~\ref{main-res} we show the main results of the paper,
while in  Section~\ref{vis-prin} we present the driving principles of the theory that gives 
rise to a computational model on the emergence of visual features, discussed in
Section~\ref{Cognitive-action-section}, that is inspired by variational laws of analytic mechanics.
The discretization of this model on the retina are described in~\ref{DiscreteRetina}, while 
Section~\ref{light-vis} sheds light on the general questions raised in Section~\ref{inq-vis}.
Some experimental results are given in Section~\ref{ExpResults} and,
finally, some conclusions are driven in Section~\ref{ConcSec}.


%% file: inq.tex
\label{inq-vis}

The theory proposed in this paper offers a computational perspective on the emergence
of visual features regardless of the ``body'' which sustains the processing. 
The theory is rooted on the need to address some fundamental questions
that involve vision in animals, and that are likely to be very important in order to 
construct an effective and efficient computational model for computers. 
As it will become early clear, the need of visual features that support
the property of motion invariance plays a central role in most of the
questions outlined below. 

\begin{enumerate}

\item [$Q1$] {\em How can humans conquer visual skills without requiring ``intensive supervision''?} \\
Recent remarkable achievements in  computer vision are mostly based
on tons of supervised examples\dash--- of the order of millions! 
This does not explain how can humans conquer visual skills with  scarse
``supervision'' from the environment. 
Hence, there is plenty of evidence and motivations 
for invoking a theory strongly rooted in unsupervised learning that
can be  
capable of explaining the emergence of features from 
visual data collections. While the need for theories of unsupervised learning 
in computer vision has been advocated in a number 
of papers (see e.g.~\cite{DBLP:journals/corr/TavanaeiMM16},
\cite{Lee:2009},\cite{DBLP:conf/cvpr/RanzatoHBL07},
\cite{DBLP:conf/iccv/GoroshinBTEL15}),
so far,  because of many recent successful applications, 
the powerful representations that arise from supervised learning,
seem to attract much more interest. 
While information-based principles could themselves suffice to construct 
visual features, the absence of any feedback from the environment make those
methods quite limited with respect to supervised learning. Interestingly, 
one of the claim of this paper is that motion invariance  
inherently offers a huge amount of ``free supervisions'' from the visual environment,
thus explaining the reason why humans do not need the massive supervision
process that is dominating feature extraction in convolutional neural networks.

\item [$Q2$] {\em How can animals gradually conquer visual skills in 
a truly temporal-based visual environment?} \\
Animals, including primates, conquer visual skills by living in their own visual environment. 
This is gradually achieved without needing to separate learning from test environments. 
At any stage of their evolution, it looks like they acquire the skills that are
required to face the current tasks. On the opposite, most approaches to computer
vision do not really grasp the notion of time. The typical ideas behind on-line learning
do not necessarily capture the natural temporal structure of the visual tasks. Time plays a crucial role in any cognitive process. One might believe that this is restricted to human life, but more careful analyses lead us to conclude that the temporal dimension plays a crucial role in the well-positioning of most challenging cognitive tasks, regardless of whether they are supported by humans or machines. 
Interestingly, nowadays dominating trend leads to struggle for the acquisition of huge labeled databases, while
the truly incorporation of time might led to a paradigm shift in the interpretation of the learning and test environment and construct visual features without needing any labeling.
The theory proposed in this paper is framed in the context of agent life characterized
by the ordinary notion of time, which emerges in all its facets. We are not concerned
with huge supervised visual data repositories, but merely with the agent life in its own visual environments.
The extraction of features in such a temporal-based visual environment 
is the main objective of this paper.

\item [$Q3$] {\em Can animals see in a world of shuffled frames?}\\
One might figure out what human life could have been in a world of visual 
information with shuffled frames. Could  children really acquire visual skills in 
such an artificial world, which is the one we are presenting to machines?
Notice that in a world of shuffled frames, for a  video to be recorded, we require a space that is significantly larger than the space required to store the 
corresponding temporally coherent visual stream. 
This is a serious warning that is typically neglected. As a consequence,
any recognition process is likely to be remarkably more difficult when shuffling
frames, which clearly indicates the importance of keeping the  
spatiotemporal structurethat is offered by nature. This calls for the formulation of a theory 
of learning capable of capturing spatiotemporal structures. Basically, we need to 
abandon the indisputable issue  of restricting computer vision to the processing of images.
The reason for formulating a theory of learning on video instead of on images
is not only rooted in the curiosity of grasping the computational mechanisms 
that take place in nature.
 It looks like that, while ignoring the crucial role of 
temporal coherence, learning visual features leads to tackling a problem that is  
remarkably more difficult than the one nature has prepared for humans! 
In a sense, the very good results that we already can experiment nowadays on the 
extraction of visual features are quite surprising, but they are mostly due to 
the stress of the computational power and the artificial framework of supervised learning. 
The theory proposed in this paper relies on the choice of capturing
temporal structures in natural visual environments, which is claimed to 
simplify dramatically the problem at hand, and to give rise to a reduce 
dramatically the computational burden.

\item [$Q4$] {\em How can humans attach semantic labels at pixel level?}\\
Humans provide scene interpretation thanks to linguistic descriptions. 
This requires a deep integration of visual and linguistic skills, that are
required to come up with compact, yet effective visual descriptions. 
However, amongst these high level visual skills, it is worth mentioning
that humans can attach semantic labels to a single pixel in the retina. 
While this decision process is 
inherently interwound  with a certain degree of ambiguity, it is remarkably
effective. The linguistic attributes that are extracted are related to 
the context of the pixel that is taken into account for label attachment, while
the ambiguity seems to be mostly a linguistic more than a visual issue. 
The theory proposed in this paper addresses directly this visual skill
since the hidden labels can be extracted for a given pixel at different levels of abstraction. 
The bottom line is that human-like linguistic descriptions of visual scenes 
is gained on top of pixel-based feature descriptions that, as a byproduct,
must allow us to perform semantic labeling. Interestingly, there is more;
as it will be shown in the following, there are in fact computational issues
that lead us to promote the idea of carrying out the feature extraction 
process while focussing attention on salient pixels.

\item [$Q5$] {\em What could drive the functional difference between the
ventral and dorsal mainstream in the visual cortex?}\\
It has been pointed out that 
the visual cortex of humans and other primates is composed of two main information pathways
that are referred to as the  ventral stream and dorsal stream~\cite{GoodaleMilner92}.
The ventral ``what'' and the dorsal ``where/how'' 
visual pathways are traditionally distinguished,
so as the ventral stream is devoted to perceptual analysis of the visual input, such as object recognition, whereas the dorsal stream is concerned with  motion ability in the interaction with the environment. The enforcement of motion invariance is clearly conceived for extracting 
features that are useful for object recognition to assolve the ``what'' task. 
Of course, neurons with built-in motion invariance are not adeguate to make spatial estimations. 
The model behind the learning of the filters indicates the need to access to velocity estimation,
which is consistent with neuroanatomical evidence. Interestingly, we will see that the 
theory also advocates the need of hierarchical structures for the dorsal mainstream, but 
there is one more reason for those structures in the ventral stream. 

\item [$Q6$] {\em Why do we need a hierarchical architecture with receptive fields?} \\
Beginning from early studies by Hubel and Wiesel~\cite{Hubel62}, neuroscientists have
gradually gained evidence  that the visual cortex presents a hierarchical 
structure, and that the neurons process the visual information on the basis of inputs 
restricted to receptive field. Is there any reason why this solution has been
developed? We can promptly realize that, even though the neurons are restricted
to compute over receptive fields, deep structures easily conquer the
possibility of taking large contexts into account for their decision.  
Is this biological solution driven by computational laws of 
vision? In this paper we provide evidence of the fact that receptive fields do 
favor the acquisition of motion invariance which, as already stated, is the 
fundamental invariance of vision. Since hierarchical architectures is the natural
solution for developing more abstract representations by using receptive fields, it turns
out that motion invariance is in fact at the basis of the biological structure of the visual 
cortex. The computation at different layers yields features with progressive degree
of abstraction, so as higher computational processes are expected to use all the
information extracted in the layers.

\item [$Q7$] {\em Why do animals focus attention?}\\
The retina of animals with well-developed visual system is organized
in such a way that there are very high resolution receptors in a restricted
area, whereas lower resolution receptors are present in the rest of the retina.
Why is this convenient? One can easily argue that any action typically takes place
in a relatively small zone in front of the animals, which suggests that the 
evolution has led to develop high resolution in a limited portion of the retina. 
On the other hand, this leads to the detriment of the peripheral vision, that 
is also very important. In addition, this could apply for the dorsal system whose
neurons are expected to provide information that is useful to support movement
and actions in the visual environment. At a first glance, 
the ventral mainstream, with neurons 
involved in the ``what'' function, does not seem to benefit from foveal eyes.
The theory proposed in this paper strongly supports the need for
foveal retinas, when we need to achieve an efficient construction of visual features
delegated to sustain object recognition. 
However, it will be argued that the most important reason for focussing attention is
that of  dramatically simplifying the computation and limit the ambiguities 
that come from the need to sustaining a parallel computation over each frame. 

\item  [$Q8$]  {\em Why do foveal animals perform eye movements?}\\
Human eyes make jerky saccadic movements during ordinary visual 
acquisition.  One reason for these movements is that the fovea
provides high-resolution in portions of about $1,2$ degrees.
Because of such a small high resolution portions, the overall sensing of a scene
does require intensive movements of the fovea. Hence, the foveal movements
do represent a good alternative to eyes with uniformly high resolution 
retina. On the other hand, the preference 
of the solution of foveal eyes with saccadic movements is arguable;
while a uniformly high resolution retina is more complex to achieve
than foveal retina, saccadic movements in this case are less important. The information-based
theory presented in this paper makes it possible to conclude that foveal
retina with saccadic movements is in fact a solution that is 
computationally sustainable and very effective.

\item [$Q9$]  {\em Why does it take 8-12 months for newborns to achieve adult visual acuity?}\\ 
There are surprising results that come from developmental psychology 
on what a newborn see. Charles Darwin came up with the following remark: 
\begin{quote}
	It was surprising how slowly he acquired the power of following with his eyes 
	an object if swinging at all rapidly; for he could not do this well when seven 
	and a half months old.
\end{quote} 
At the end of the seventies, this early remark was given a technically sound
basis~\cite{DobsonVR-1978}.   In the paper, three techniques, \dash--- optokinetic nystagmus (OKN), preferential looking (PL), and the visually evoked potential (VEP) \dash--- were used to 
assess visual acuity in infants between birth and 6 months of age.
More recently, the survey by  Braddick and Atkinson~\cite{BraddickVR-2011} provides 
an in-depth discussion on the state of the art in the field. It is clearly stated that 
for  newborns to gain adult visual acuity, depending on the specific visual test, 
several months are required.
Is the development of adult visual acuity a biological issue or does it come 
from higher level computational laws? 
This paper provides evidence to conclude that the blurring process taking place
in newborns is in fact a natural strategy to optimize the cognitive action defined
by Eq.~\ref{CognitiveActionEq} under causality requirements.
Moreover, the strict limitations both in terms of spatial and temporal resolution of the
video signal, according to the theory, help conquering
visual skills.  

\item [$Q10$] {\em Causality and Non Rapid Eye Movements (NREM) sleep phases}\\  
Computer vision is mostly based on huge training sets of images, whereas
humans use video streams for learning visual skills. Notice that because of 
the alternation of the biological rhythm of sleep, humans somewhat process collections of
visual streams pasted with relaxing segments composed of ``null'' video signal.
This happens mostly during NREM phases of sleep,
in which also eye movements and connection with visual memory 
are nearly absent. Interestingly, the
Rapid Eye Movements (REM) phase is, on the opposite, similar to ordinary 
visual processing, the only difference being that the construction of visual features
during the dream is based on the visual internal memory representations ~\cite{AndrillonNature2014}. 
As a matter of fact, the process of learning the filters experiments an alternation 
of visual information with the reset of the signal. 
We provide evidence to claim that 
such a relaxation coming from the reset of the signal nicely fits  the 
overall objective of the visual agent. 

In particular, throughout the paper, we will see that the reset of the visual 
information favors the optimization under causality requirements.
Hence, the theory offers an intriguing interpretation
of the role of eye movement and of sleep for the optimal development of visual features. In a sense, the theory also offers a general framework for interpreting
the importance of the day-night rhythm in the development of
visual features. 

\end{enumerate}

Throughout the paper we will address the above questions during the development 
of the main results on visual features.

%% file: res.tex
%
We are given a retina ${\cal X} \subset \bbR^{2}$, \rev{that} can \rev{be} 
formally  regarded as a compact subset of the plane; 
for the moment we will not assume any specific 
shape. 
The purpose of this paper is that of analyzing the mechanisms that give rise
to the construction of local features for any pixel $x \in {\cal X}$ of the retina, at any
time $t$. These features, along with the video itself, can be regarded as 
visual fields, that are defined on the retina and on a given time horizon
$[0\dts T]$; clearly
the analysis of on-line learning of visual features leads to regard 
the horizon as $[0\dts \infty)$. 
As it will be clear in the remainder of the paper, a set of symbols are extracted
at any layer of a deep architecture, so as any pixel\dash---along with its
context\dash---turns
out to be represented by the list of symbols extracted at each layer.
The computational process that we define involves the video
as well as appropriate vector fields that are used to express
a set of pixel-based features properly used to capture contextual
information.  The video, as well as all the involved
fields, are defined on the domain ${\cal D}={\cal X} \times [0\dts T]$.
In what follows, points on the retina will be represented
with two dimensional vectors $x=(x_1,x_2)$ on a defined 
coordinate system on the retina.
The temporal coordinate is usually denoted by
$t$, and, therefore, the video signal on the pair $(x,t)$ is $\textrm{C}(x,t)$. For
further convenience we also define the map $\textrm{C}_t\colon {\cal X}\to \bbR^m$ so that
$\textrm{C}_t(x)\equiv \textrm{C}(x,t)$.
The color field can be thought of as a special field that is characterized by the
RGB color components of any single pixel; in this case $m=3$. 
\begin{figure}
		\centering
		\includegraphics{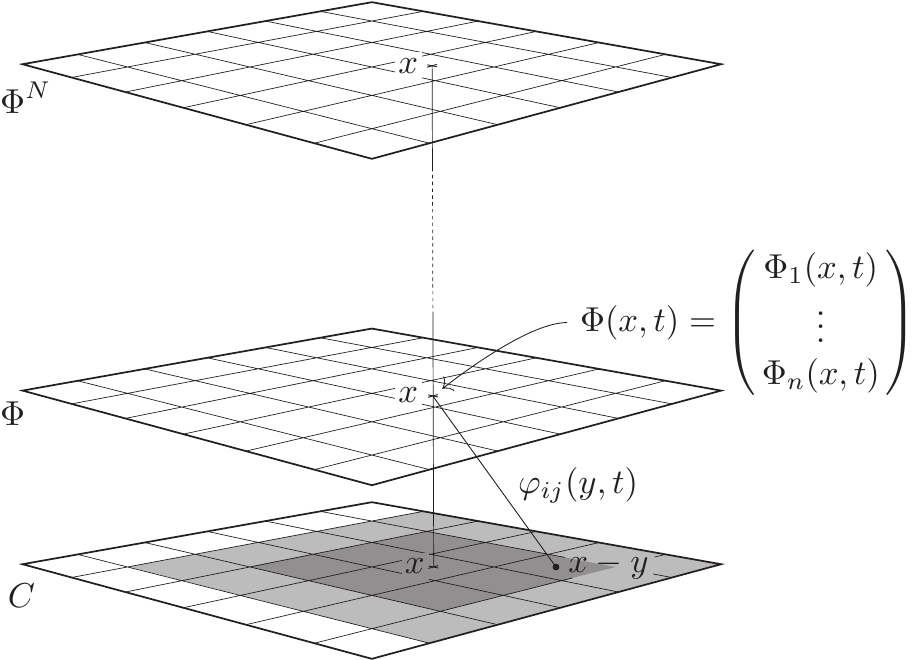}
		\caption{Convolutional computation in a deep network.
		The input is processed by convolutional filters which 
		transform $\textrm{C}$   to $\Phi$.
                Notice that the features
		are extracted at different level on the same pixel $x$.}
	\label{ConvFig}
\end{figure}

Now, we are concerned with the problem of extracting visual features that, unlike
the components of the video, express the information associated with 
the pair $(x,t)$ and with its spatial context. Basically, one would like to extract
visual features that characterize the information in the neighborhood of pixel $x$.
A possible way of constructing this kind of features is to construct the 
map\footnote{Throughout the
paper we use the Einstein convention to simplify the equations.}
\begin{equation}
	\Phi_i(x,t)=
	\sum_{j=1}^{m}\int_{\cal X} dy \ \varphi_{ij}(x,y,t,\textrm{C}_j(y,t)).
\label{Kernel-contex-def2}
\end{equation}
Here, the feature defined by index $i=1,\ldots,n$, that is 
denoted by $\Phi_i(x,t)$ presents a spatial dependence
on any pixel $y \in {\cal X}$. 
Here we assume that $n$ symbols are generated from the $m$
components of the video.
In the special case in which
such a dependence only involves the distance from
the pixel of coordinates $x$ on which we want to determine
the feature, the above equation reduces to 
\begin{equation}
	\Phi_i(x,t)=
	\sum_{j=1}^{m}\int_{\cal X} dy \ \varphi_{ij}(x-y,t,\textrm{C}_j(y,t)),
\label{Kernel-contex-def3}
\end{equation}
which becomes the convolutional computation in case of
linear filters $\varphi_{ij}$, that is
\begin{equation}
	\Phi_i(x,t)=
	\sum_{j=1}^{m}\int_{\cal X} dy \ \varphi_{ij}(x-y,t)\textrm{C}_j(y,t)
\label{Kernel-contex-def}
\end{equation} 
Notice that $\varphi(z,t)$ is responsible of expressing the spatial dependencies,
and that one could also extend the context in the temporal dimension. 
However, the immersion in the temporal dimension that arises from the
formulation given in this paper makes it reasonable to begin restricting the 
contextual information to spatial dependencies on the the retina.
In addition, it is worth mentioning that the agent is expected to return a decision 
also in case of fixed images, which represents a further element for considering 
features defined by Eq.~(\ref{Kernel-contex-def}). 
In general, the kernel $\varphi$ can be
regarded as a map from ${\cal X} \times {\cal X}\times [0\dts T] \to \bbR^{n,m}$. 
Whenever $\varphi(x,y,t) \leadsto \varphi(x-y,t)$ the above definition
reduces to an ordinary spatial convolution. 
Notice that while the kernel $\varphi(x,y,t)$ can handle the ambiguities that arise from the
the presence of strong visual deformations of the same features in the same frame at time $t$, 
the same does not hold for $\varphi(x-y,t)$, that only reasonably deals with those deformations
while focusing attention on $x$ at time $t$. This issue will be widely covered in the following, but
it is already clear that the convolutional filter $\varphi(x-y,t)$ can face strong visual 
deformation only when supported by focus of attention driven computation. The presence
of multiple deformations in the same frame yields inconsistent decisions, so as only 
an ``averaging solution'' can be discovered.
The computation of   $\Phi(x,t)$ yields a field with $n$ features, instead of
the three components of color in the video signal.  However, 
Eq.~(\ref{Kernel-contex-def}) can be used for carrying out a piping scheme
where a new set of features $\Phi^2$ is computed from $\Phi$
and so forth (see Fig.~\ref{ConvFig}). Of course,
this process can be continued according to a deep computational structure 
with a homogeneous convolutional-based computation, 
which yields the features 
at the $p$ convolutional layer. 
The theory proposed in this paper focuses on the construction of any of these
convolutional layers which are expected to provide higher and higher degree of  abstraction
as we increase the number of layers. 
The {\it filters\/}  $\varphi$ completely determine the
features $\Phi(x,t)$. In this paper we formulate a theory for the discovery of $\varphi$
that is based on three driving principles:

\begin{itemize}
\item	{\em Optimization of information-based indices};
\item {\em Motion invariance};
\item {\em Parsimony principle}.
\end{itemize}
%
%
\begin{figure}
		\centering
		\includegraphics{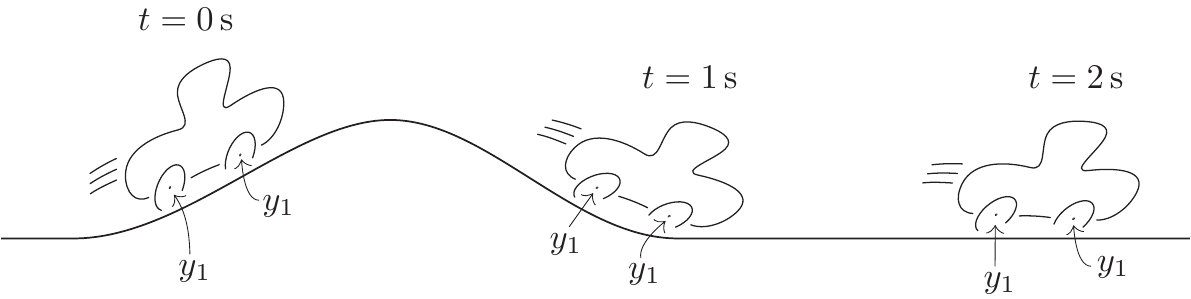}
		\caption{\small Motion invariance in the feature extraction process.
		The symbol $y_{1}$, that defines a features at the beginning of
		motion ($t=0$ s), must be coherently extracted during the 
		movement\dash---see the enforcement of the coherence requirement at $t=1,2$ s.}
	\label{MotionInvarianceFig}
\end{figure}

While the first and third principles are typically adopted in classic unsupervised learning, 
motion invariance does characterize the approach followed in this paper.
Of course, there are visual features that do not obey the motion invariance principle. 
Animals easily estimate of the distance to the objects in the environment, a property that clearly
indicates the need for features whose value do depend on motion. The perception of
vertical visual cues, as well as a reasonable estimation of the angle with respect to the
vertical line also suggests the need for features that are motion dependent. 

Now, we provide arguments to support the principled framework of this paper. 
Like for human interaction, visual concepts are expected to be acquired by the agents solely by processing their own visual stream along with human supervisions on selected pixels, instead of relying on huge labelled databases. In this new learning environment based on a video stream, any intelligent agent willing to attach semantic labels to a moving pixel is expected to take coherent decisions with respect to its motion. Basically, any label attached to a moving pixel has to be the same during its motion. Hence, video streams provide a huge amount of information just coming from imposing coherent labeling, which is likely to be the primary information associated with visual perception experienced by any animal. Roughly speaking, once a pixel has been labeled, the constraint of coherent labeling virtually offers tons of other supervisions, that are essentially ignored in most machine learning approaches working on big databases of labeled images.
It turns out that most of the visual information to perform semantic labeling comes from the motion coherence constraint, which might explain the reason why children learn to recognize objects from a few supervised examples. The linguistic process of attaching symbols to objects takes place at a later stage of children development, when he has already developed strong pattern regularities. We conjecture that, regardless of biology, the enforcement of motion coherence constraint is a high level computational principle that plays a fundamental role for discovering pattern regularities. 
%
%

As for the adoption of the parsimony principle
in visual environments, we can use appropriate 
functionals to enforce both the spatial and temporal smoothness of the solution. 
While the spatial smoothness can be gained by penalizing solutions with 
high spatial derivatives\dash---including the zero-order
derivatives\dash---temporal smoothness arises from the introduction of
kinetic energy terms which penalizes high velocity and, more generally, 
high temporal derivatives.

\rev{The agent behavior turns out to be driven by the minimization of an 
appropriate functional ${\cal A}$ that combines the all above principles.}
Since the optimization is generally formulated over arbitrarily large time horizons, 
all terms are properly weighted by a {\em discount factor} that leads to ``forget''
very old information in the agent life. As it will be shown,
this contributes to a well-position of the
optimization problem and gives rise to dissipation processes~\cite{DBLP:journals/tcs/BettiG16}.






The main result in this paper is that this optimization can be interpreted in terms 
of laws of nature expressed by a temporal differential equation. 
When regarding the retina as a discrete structure, we can compute the
probability that at time $t$, in pixel $x$, the emitted symbol is $y_{i}$
by\footnote{We use Einstein's notation.} 
\[
	\Phi_{ix}(t)=\varphi_{iky}(t) \textrm{C}_{k(x-y)}(t).
\] 
Here, for any pair of symbols
$y_k$, $y_i$, and for any pixel with position $z$, in the coordinate system defined by $x$,
the filter $\varphi_{ikz}$ is the temporal function that the agent is expected
to learn from the visual environment. Basically, the process of learning consists
of determining 
\begin{equation}
\hat{\varphi} = \arg \min_{\varphi} {\cal A}(\varphi).
\end{equation}
Here we overload the symbol ${\cal A}$ to denote both the case of
the cognitive action defined on spatially continuous filters and the 
case of a discrete retina. 

In this paper we show how we can get the filters $\hat{\varphi}$
by addressing the problem of determining stationary points of the action  ${\cal A}$
and, moreover, we discuss the existence of $\hat{\varphi}$.
The filters are determined by imposing 
\begin{equation}
	\delta {\cal A}(\varphi)=0,
\label{NullVariationOfAction}
\end{equation}
that is the nullification of the variation of the action, which corresponds with
the stationarity condition on  ${\cal A}$. 
It is worth mentioning that this does not correspond
with the classic gradient flow used in machine learning, since in that case
the filters are updated by using the gradient heuristics towards the stationary
condition. The consequences of imposing condition~\eqref{NullVariationOfAction}
is mostly discussed in Section~\ref{Cognitive-action-section}, where
we prove that, when considering
the continuous setting of computation in which $\varphi_{ij}(z,t)$ are the 
unknown filters,  there is no local solution
to this problem, since any stationary point of this functional turns out to be characterized
by the integro-differential equation~\eqref{IntegroDiffEL}. 
Interestingly, we show that we can naturally gain 
a local solution when introducing introducing the classic notion of
receptive field. This issue turns out to be relevant also in
case we deal with a discrete retina. In that case
we prove that the minimum of the cognitive action 
corresponds with the discovery of the filters $\varphi_{ijy}$
that satisfy the forth-order time-variant differential equation~\eqref{Full-EL},
where $q$ is the linearized vector of $\varphi_{ijx}$.
The equation contains coefficients which inherits by the time-variance from 
the video. 
The analysis carried out in the paper shows how can we attack the problem
either in the case in which the agent is expected to learn from a given video
stream with the purpose to work on subsequent text collections, or in the
case in which the agent lives in a certain visual environment, where there is
no distinction between learning and test phases. Basically, it is pointed out
that only the second case leads to a truly interesting and novel result.

It is shown that the solution of the above differential equation is strongly
facilitated when performing an initial blurring of the video that lasts
until all the visual statistical cues are likely been presented to the agent. 
This very much resembles 
early stages of developments in newborns~\cite{BraddickVR-2011}.  
It is shown that the given differential equations of learning lead to
conclude that only a very slow dynamics takes place, which means that
all the derivatives of $q$ are nearly null and, consequently, $q$ is 
nearly constant. This strongly facilitates the numerical solutions and,
in general, the computational model turns out to be very robust, a property
that is clearly welcome also in nature. As time goes by, while the 
blurring process increases the visual acuity the coefficients of the
differential equation begin to change with velocity that is connected
with motion. However, in the meantime, the values of the filters 
have reached a nearly-constant value. Basically, the learning trajectories
are characterized by the mentioned nearly-null derivatives, a condition
that, again strongly facilitates the well-position of the problem. 

%
%
%
A further intuitive reason for a slow dynamics of $q(t)$ is also a consequence of visual invariant features.
For example, when considering a moving car and another one of the same type parked somewhere in
the same frame, during the motion interval, 
the processing over the parked car would benefit from a nearly constant 
solution.  This suggests also searching for the 
same constant solution on the corresponding moving pixel. 
When regarding the problem of learning in a truly on-line mode,
the previous differential equation can be considered as the model
for computing $\varphi_{ijy}$ given the Cauchy conditions.
Of course, the solution is affected by these initial conditions. Moreover,
as it will be clear in the reminder of the paper, the previous differential equations
yield the minimization of the action under appropriate border conditions that 
correspond with forcing a trajectory that satisfies the condition of 
nearly-null of the first, second, and third derivatives of $q$. When 
joined with the blurring process this leads to a causal dynamics
driven by initial conditions that are compatible with boundary conditions
imposed at any time of the agent's life.

The puzzle of extracting robust  cues from visual scenes
has only been partially faced by nowadays successful approaches to computer 
vision. The remarkable achievements  of the last few years have been
mostly based on the accumulation of huge visual collections gathered by
crowdsourcing. An appropriate set up of convolutional networks trained
in the framework of deep learning has given rise to very effective 
internal representations of visual features. They have been successfully used
by facing a number of relevant classification problems by transfer learning. 
Clearly, this approach has been stressing the power of deep learning when 
combining huge supervised collections with massive parallel computation.
In this paper, we argue that while stressing this issue we 
have been facing artificial problems that, from a pure computational point of view, 
are likely to be significantly more complex than natural visual tasks 
that are daily faced by animals. In humans, the emergence of cognition from visual
environments is interwound with language. This often leads to attack
the interplay between visual and linguistic skills by simple models
that, like for supervised learning, strongly rely on linguistic attachment.
However, when observing the spectacular skills of the eagle that catches the pray,
one promptly realizes that for an in-depth understanding
of vision, that likely yields also an impact in computer implementation,
one should begin with a neat separation with language! 
This paper is mostly motivated by the curiosity of addressing a number
of questions that arise when looking at natural visual processes.
While they come from natural observation, they are mostly regarded
as general issues strongly rooted in information-based principles, that we
conjecture are of primary importance also in computer vision.

\begin{figure}
\includegraphics{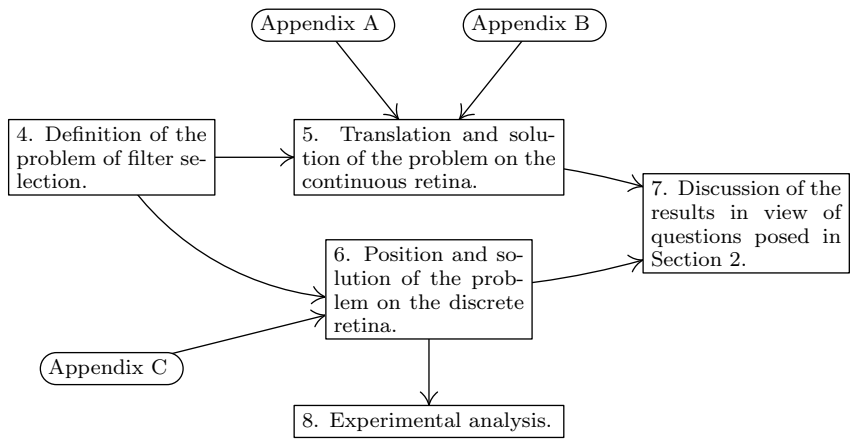}
\rev{\caption{Conceptual map of Sections~4--5. Each arrow indicate the main
dependencies of each section.}}
\label{conceptual-map}
\end{figure}

\rev{Figure~\ref{conceptual-map} shows how the theory and results that we
have described above have been organized in the remaining sections on the
paper.}

%% file: princ.tex
%
%
\medskip\noindent
We can provide an interpretation of the processing carried out
by our visual agent in the framework of information theory. 
The basic idea is that the agent produces a set of symbols from a given alphabet while processing the video. 
Unlike traditional approaches to computer vision, we begin considering
that maps on the retina are refined with the final purpose of transforming
the color field, which reports pixel-based information, into visual features
that take the pixel context into account. As such, one could expect 
each pixel be associated with a remarkable number of features that
somehow express the visual information in its neighborhood. 
A similar map of features, $\Phi(x,t)$ 
is clearly reporting an enriched color field that, 
just like $\textrm{C}(x,t)$, still operates at pixel level.
In doing so, all subsequent cognitive tasks that relies on video can benefit
of the processing on $\Phi(x,t)$ that, unlike $\textrm{C}(x,t)$, is
expected to express relevant visual features that emerge from
the context. It will be shown that 
the search for an appropriate enrichment of the color field
leads to important architectural conclusions that address some of
questions raised in the previous section and very much 
support nowadays emphasis on the deep networks.

\rev{	We use an information-based approach  to determine
	$\varphi$. Beginning from the color field $\textrm{C}$, we attach
	symbol $y_i \in \Sigma$ of a discrete vocabulary
	to pixel $(x,t)$ with probability $\Phi_i(x,t)$. 
	The principle of Maximum Mutual Information (MMI) is a natural way 
	of maximizing the transfer of information from the visual source, expressed
	in terms of mixtures of colors, to the source of  symbols $y_i \in \Sigma$.
	Clearly, the same idea can be extended to any layer in the hierarchy.
	Once we are given a certain visual environment over a certain 
	time horizon $[0\dts T]$\dash---which can  be extended to $[0\dts +\infty)$\dash---and once the filters $\varphi$ have been defined,
	the mutual information turns out to be a functional of $\varphi$, that is denoted
	as ${\cal I}(\varphi)$. However, in the following, it will be shown that the more general view behind the
	the maximum entropy principle (MaxEnt) offers a better framework for the 
	formulation of the theory}

\medskip\noindent
\textbf{MMI principle.\enspace} 
The purpose of the visual agent is to generate symbols from the video. 
We will make use of the Maximum Mutual Information principle (MMI),
according to which we want to maximize the transfer of information from the 
input to the generated symbols. As it will be shown later, this can also be
reformulated within the framework of the 
Maximum Entropy principle~\cite{Jaynes1957}. 

Let us  define random variables $X$ and $T$, which  take into account the spatiotemporal
probability distribution, while $Y$ is used to specify the probability distribution 
over the possible symbols, and $F$ to specify the video frame.
Basically, the realization of these of $(X,T,F)$ is the triple $(x,t,f)$,
which describes the spatiotemporal pair $(x,t)$ (pixel-time) at 
frame $f$, that is clearly characterized by the given video signal at time $t$.
In order to assess the information transfer from $(X,T,F)$ to $Y$ 
we consider the corresponding mutual information $I$. Clearly, it is zero
whenever random variable $Y$ is independent of $X$, $T$ and $F$.
The mutual information can be expressed by
\begin{equation}
	I(Y;X,T,F)=S(Y)-S(Y\mid X,T,F).
\label{MI-defition}
\end{equation}
The conditional entropy $S(Y\mid X,T,F)$ is given by 
\begin{equation}
	S(Y\mid X,T,F)=-\int_\Omega \sum_{i=1}^n dP_{X,T,F} \  p_i
	\log p_i\, 
\label{CondEntropyDef}
\end{equation}
where $p_i$ is the  probability of $Y$ conditioned
to the values of $X$, $T$ and $F$, $dP_{X,T,F}$ is the joint measure of
the variable $(X,T,F)$, and $\Omega$ is a Borel set in the $(X,T,F)$
space. The agent is supposed to generate symbols  $y_{i}, \ i=1,\ldots,n$ along with the
corresponding probabilities. 
Now, let us make two fundamental assumptions:
\begin{itemize}
\item The conditional probability $p_i(x,t,f)$, where $f$ is a realization of
random variable $F$,  is given by the $i$-th feature field $\feat_i(x,t)$.
Notice that one can also distinguish between the feature map $\feat_i(x,t)$
and the symbols to be used in the codebook. In that case, 
we need an additional map $\feat(x,t) \to \Psi(\Phi(x,t))$, that could
be properly expressed by a feedforward neural network
that is charged of computing the probability $p_i(x,t,f)$.

\item Random variables $X,T,F$ follows the ergodic-like assumption, 
so as 
we can perform the replacement:
$$\int_\Omega dP_{X,T,F}\longrightarrow\int_{\cal D}d\mu,
$$
where ${\cal D} = {\cal X} \times {\cal T}$ is the set of pairs $(x,t)$, with
$x \in {\cal X}$ and $t \in {\cal T}$.
\end{itemize}
A reasonable measure is given by
$d\mu = f(x,t)\, dxdt$; basically, this comes from the
visual environment on which the agent is supposed to operate.
Furthermore, we will assume that we are given 
the trajectory of the focus of attention $a(t) \in {\cal X}$ and that
$f(x,t)$ is factorized according to 
\begin{equation}
	f(x,t)=g(x-a(t))h(t),
\label{ergodic-eq}
\end{equation} 
This ergodic-like translation
of the probabilistic measure suggests that the density is higher
where the eye is focussing attention, that is in the neighborhood
of $a(t)$; this can be achieved by means of a function $g(x-a(t))$
peaked on the focus of attention. Finally, the factor $h(t)$
can be thought of as a weight  so as
to penalize more and more the errors of the agents as  
time goes by. As it will be shown in the following, this 
factor plays a crucial role in the establishment of dissipation
which is related to the enforcement of a temporal direction. 
It is quite obvious that the measure 
$d\mu$ only makes sense provided
that the function $h$ does not change significantly during 
statistically significant portions of visual environments.  

Notice that in truly active environments humans and robots can select even the environment, which may result in a remarkable variability of the probability distributions. 
For instance, living like Eskimos leads to acquire visual environments that are remarkable different from Newyork\`{e}se. Regardless of the huge visual environmental gap, however, humans seem to adapt very well their visual system when moving from New York to snow territories and vice versa. This suggests that when learning in natural environments  focus of attention strategies, that are associated with the computation $a(t)$, seem to be remarkably 
important in the acquisition of visual skills.

The research on focussing of attention trajectories $a(t)$ 
is rooted on solid studies at the crossroad of neuroscience and computer
vision, and it has been recently given a formulation 
~\cite{DBLP:conf/nips/ZancaG17, zanca2019gravitational}
that is very much aligned with
the theoretical framework of this paper. 

Whenever these two assumptions hold, 
we can rewrite the conditional entropy defined by
Eq.~(\ref{CondEntropyDef}) as 
\begin{equation}
	S(Y\mid X,T,F)=-\int_{\cal D}d\mu(x,t) \ \sum_{i=1}^n\feat_i(x,t)\log
	\feat_i(x,t).
\label{CondEntrbyC}
\end{equation}
Similarly for the entropy of the variable $Y$ we can write
\begin{equation}
	S(Y)=-\sum_{i=1}^n \Pr(Y=y_i)\log\Pr(Y=y_i).
\label{BalEntropybyPr}
\end{equation}
Now, if we use the law of total probability to express $\Pr(Y=y_i)$ 
in terms of the conditional probability $p_i$ and  use
the above assumptions we get
\begin{equation}
	\Pr(Y=y_i)=\int_\Omega dP_{X,T,F} \ p_i=\int_{\cal D} d\mu(x,t) \, \feat_i(x,t).
\label{BiGPfromp}
\end{equation}
Then
\begin{equation}
	S(Y)=-\sum_{i=1}^n \Big(\int_{\cal D} d\mu(x,t) \, \feat_i(x,t)\Big)
	\log\Big(\int_{\cal D} d\mu(x,t) \, \feat_i(x,t)\Big).
\label{BalancingEntropybyC}
\end{equation}
Finally the mutual information becomes
\begin{equation}
	I(Y;X,T,F)=\sum_{i=1}^n\Big(\int_{\cal D} d\mu \, \feat_i\log
	\feat_i - \int_{\cal D} d\mu \,\feat_i \cdot \log\int_{\cal D} d\mu \,\feat_i\Big).
\label{MIbyC}
\end{equation}
Of course, $\forall x,t: \ \feat_i(x,t)$ is subject to the probabilistic constraints
\begin{align}
\vcenter{\halign{$#$\hfil&\qquad#\hfil\cr
	\sum_i \feat_i(x,t)=1&(normalization)\cr
	0\le\feat_i(x,t)\le1&(positivity)\cr}}
\label{ProbNormalization}
\end{align}
In the case there is an additional neural map $\Psi$ to determine 
the probability, the normalization is moved to the range of the
map itself, which allows the typical presence of more distributed 
representations on $\Phi$.

\medskip\noindent  
\textbf{MaxEnt principle.\enspace} 
An agent driven by the MMI principle carries out an unsupervised
learning process aimed at discovering the symbols defined by random
variable $Y$. Interestingly, when the constraints are given a soft-enforcement, 
the MMI principle has a nice connection with the Max-Ent principle~\cite{Jaynes1957}:
The maximization of the mutual information is somewhat related 
to the maximization of the entropy while softly-enforcing 
the constraint that the conditional entropy is null. 
In particular, in MMI both the entropy terms get the same  value of the weight, 
but one can think of different implementations of the MaxEnt principle 
that very much depend on the  choice of the weights of the two entropy terms.
As an extreme case, one can also remove the conditional entropy term
and consider motion invariance only.  
The satisfaction of the conditional entropy constraint needs to be paired
with the maximization of the entropy, which protects us from the development of
trivial solutions (see~\cite{MachineLearningMG2018} pp. 99--103 for further details).
Of course, the probabilistic normalization  constraints stated by Eq.~\ref{ProbNormalization} comes along
with the information-based formulation.

\rev{
Concerning the MMI principle, it is worth mentioning that it can be regarded as a 
special case of the MaxEnt principle when the constraints correspond with the
soft-enforcement of the conditional entropy, where the weight of its associated penalty
is the same as that of the entropy (see e.g.~\cite{DBLP:journals/tnn/MelacciG12}). 
Notice that while the maximization of the mutual information nicely addresses the
need of maximizing the information transfer from the source to the selected alphabet of
symbols, it does not guarantee temporal consistency of this attachment. Basically, 
the optimization of the index is also guaranteed by using the same symbol for different
visual cues. Motion consistency faces this issue for any pixel, even if it is fixed.}

\rev{
\medskip\noindent  
\textbf{Parsimony principle.\enspace} 
While the computational mechanism that drives the discovery of the symbols described in this paper is inspired by MaxEnt, a  
well-posed learning process requires that 
the map which originates the symbols be subjected to some
kind of parsimony assumption.  
	Amongst the philosophical implications, it also favors the development of a unique solution.
	The development of filters that are consistent with the above principles requires 
	the construction of an on-line learning scheme, 
	where the role of time becomes of primary importance. 
	The main reason for such a formulation is the need of imposing the development of 
	motion invariance features. Given the filters $\varphi$, 
	there are two parsimony terms, one 
	${\cal P}(\varphi)$, that penalizes abrupt spatial
	changes, and another one, ${\cal K}(\varphi)$  that penalizes 
	quick temporal transitions.}

The conditional entropy constraint only involves the
value taken by $\feat_{i}$ which depends on  $\varphi_{ij}(x,t)$, but there is no 
structural enforcement on the function $\varphi_{ij}$; its spatiotemporal changes are ignored. Ordinary regularization issues suggest to discover functions $\varphi_{ij}$ such that
$\langle P_{x,t} \varphi_{ij},P_{x,t} \varphi_{ij}\rangle$ is small, where $P_{x,t}$ is
a spatiotemporal differential operator. A simplified, yet effective choice is that of 
separating the spatial from the temporal regularization and consider 
\begin{equation} 
  \frac{\lambda_{P}}{2}
  \int_{\cal D} dtdx\, h(t) (P_{x}
  \varphi_{ij}(x,t))^{2}
  +\frac{\lambda_{K}}{2} \int_{\cal D} dtdx\, h(t)
(P_{t} \varphi_{ij}(x,t))^{2},
\label{reg-term}
\end{equation}
is ``small'', where $P_{x}, P_{t}$ are spatial and temporal differential operators, and  $\lambda_{P},  \lambda_{K}$ are non-negative reals\footnote{A simple introduction to 
differential operator that is appropriate in this context is given in~\cite{MachineLearningMG2018},
pp.~512--516.}.
Notice that the ergodic-like translation of
$d\mu$, in this case, only involves the temporal factor
$h(t)$.

\rev{
	While information-based indices optimize the 
	information transfer from the input source $\textrm{C}$ to the symbols,
	the major cognitive issues of invariances are not covered. 
	The same object, which is presented at different scales and under 
	different rotations does require different representations, which transfers 
	all the difficulty of learning to see to the subsequent problems interwound with
	language interpretation. 
	Hence, it turns out that the most important requirement that 
	the visual field $\Phi$ must fulfill is that of exhibiting the typical 
	cognitive invariances that humans 
	and animals experiment in their visual environment. 
	We claim that there is only one  fundamental invariance, 
	namely that of producing the same representation for moving pixels. 
	This incorporates 
	classic scale and rotation invariances in a natural way, which is what is experimented 
	in newborns. Objects comes at different scale and with different rotations simply 
	because children experiment their movement and manipulation. As we track moving pixels, 
	we enforce consistent labeling, which is clearly far more general than 
	enforcing scale and rotation  invariance. 
	We claim that the enforcement of motion constraint is the key for the construction of a 
	truly natural invariance.} 
As already pointed out, the visual features that in the ventral mainstream are involved in the ``what'' function need to be motion invariant.
Just like an ideal fluid is adiabatic\dash---meaning that the entropy
of any particle fluid remains constant as that the particles move about
in space\dash---in a video, once we have assigned the correct symbol to a
pixel, it must be conserved as the pixel moves on the retina. If we focus attention
on a the pixel $x$ at time $t$, which moves according to the trajectory $x(t)$
then this is formally stated by $\feat_{i}(x(t),t) = c$, being $c$ a constant. 
This ``adiabatic'' condition is thus expressed by the condition
$d\feat_{i}/dt=0$, which yields 
\begin{equation} 
 	\partial_t \feat_i+v_j\partial_j \feat_i=0,
\label{LocalMIEq}
\end{equation}
where $v\colon D\to\bbR^2$ is the velocity field that we assume that is given,
and $\partial_k$ is the partial derivative with respect to $x_k$.
Notice that in case $\phi_{ij}(x,t)=\delta(x)$ then the previous
invariance on the feature becomes the brightness invariance
condition 
\begin{equation}
	\partial_t \textrm{C}_j+v_{\alpha}\partial_{\alpha} \textrm{C}_j=0,
\label{OptFlowInv}
\end{equation}
that is typically used to estimate the optical flow~\cite{HornAI1981}. 
Here, the unknown is in fact the velocity field, whereas 
in the feature motion invariance condition~\ref{LocalMIEq}
the unknown are the filters. This can promptly be seen when
replacing $\feat_i$ as stated by Eq.~(\ref{Kernel-contex-def}) we get
\begin{align*}
	\partial_t\varphi_{ij} \textrm{C}_j+\varphi_{ij}
	\partial_t  \textrm{C}_j +  v_k \partial_k \varphi_{ij}  \textrm{C}_j
	= \partial_t\varphi_{ij} \textrm{C}_j+\varphi_{ij}
	\partial_t  \textrm{C}_j +  v_k \varphi_{ij}  \partial_k  \textrm{C}_j=0.
\end{align*}
Clearly, this is equivalent to
\begin{align*}
	\int_{\cal X}dy\,\big(\partial_t\varphi_{ij} \textrm{C}_j+\varphi_{ij}
	\partial_t  \textrm{C}_j +  \varphi_{ij} v_k \partial_k   \textrm{C}_j\big)=0,
\end{align*}
which holds for any $i=1,\ldots,n$ and $(x,t) \in {\cal D}$.
Notice that this constraint is linear in the field $\varphi$.
This can be interpreted by stating that learning under
motion invariance, for any $(x,t)$, consists of determining elements of the
kernel of  function 
\begin{equation}
	\mathscr{M}_{(x,t)}(\varphi_{ij}):= \int_{\cal X}dy\,\big(\partial_t\varphi_{ij} 
	\textrm{C}_j+\varphi_{ij} \partial_t  \textrm{C}_j 
	+   \varphi_{ij} v_k \partial_k 
 	\textrm{C}_j\big).
\label{M-fun-inv}
\end{equation}
As we can promptly see  $\mathscr{M}_{(x,t)}(\cdot)$ is defined by the knowledge
of  the video signal $\textrm{C}$ and the by availability of the optical flow $v$.
Depending on the color field $\textrm{C}$ 
it quite easy to realize that $\mathscr{M}_{(x,t)}(\varphi_{ij})$ might be
the null space, since while the possible visual configurations increase exponentially with the growth of the measure of ${\cal X}$ ,
the information associated with $\phi_{ij}$ only grows 
linearly the distance to the focus point.
Hence condition~\eqref{M-fun-inv} can be better satisfied
in case of video with smooth spatiotemporal transitions.
This is what happens for newborns, who experiment similar smooth transitions
in early stage of development~\cite{BraddickVR-2011}.
Moreover, sparseness of  $\varphi_{ij}$ 
also favors the satisfaction of~\ref{M-fun-inv}. In particular, 
as will be better discussed in the remainder of the paper, 
the satisfaction of motion invariance is favored by the receptive-field
assumption. 
It is worth mentioning that the above constraints can be enforced at
least in two different ways:
\begin{enumerate}
\item[i.] As stated above, we can impose constraint~\eqref{M-fun-inv}
	for all points $(x,t) \in {\cal D}$.
	In doing so, one enforces motion invariance in any point of 
	the retina.
	
\item[ii.] We can impose constraint~\eqref{M-fun-inv} only on the
	$(a(t),t) \in {\cal D}$, namely on the focus of attention trajectory.
\end{enumerate}
In this paper we will follow the first approach. 

\rev{
Overall, the process of learning is regarded as the minimization of the {\em cognitive action}
\begin{equation}
	{\cal A}(\varphi)= - {\cal I}(\varphi) + \lambda_{M} {\cal M}(\varphi)
	+\lambda_{P} {\cal P}(\varphi)
	+ \lambda_{K} {\cal K}(\varphi),
\label{CognitiveActionEq}
\end{equation}
where $\lambda_{M},\lambda_{P},\lambda_{K}$ are positive multipliers.
Since the above action functional ${\cal A}(\varphi)$ depends on the choice
of the multipliers $ \lambda_{M},  \lambda_{P},  \lambda_{K}$, it is quite clear that
there is a wide range of different behavior that depend on the relative weight
that is given to the terms that compose the action. As it will be shown in the following, 
the minimization of ${\cal A}(\varphi)$ can be given an efficient computational
scheme only if we give up to optimize the information transfer in one single step
and rely on a piping scheme that clearly reminds deep network architectures.}



%% file: continuum.tex
In the previous section we have discussed principles that
drive the discovery of the filters
$\varphi_{ij}$ based on the MaxEnt principle and 
regularization. We provide a soft-interpretation
of the constraints, so as the adoption of the principle corresponds with the minimization of a functional that, following~\cite{DBLP:journals/tcs/BettiG16}, 
it  referred to as the ``cognitive action'':
\begin{equation}
\begin{split}
{\cal A}_0(\varphi)=& \int_{\cal D} d\mu \,\feat_i(\varphi) \cdot\log\int_{\cal D}
d\mu \, \feat_i(\varphi)- \lambda_{C} \int_{\cal D} d\mu
\,\feat_i(\varphi)\log \feat_i(\varphi)\\ &+\lambda_{1} \int_{\cal D} \,d\mu
\, \Bigl(\sum_{i=1}^n\feat_i(\varphi)-1\Bigr)^2 -\lambda_{0} \int_{\cal D}
d\mu \, \feat_i(\varphi)\cdot[\feat_i(\varphi)<0] \cr &+ \frac{\lambda_{P}}{2}
\int_{\cal D} dtdx\, h(t) (P_{x} \varphi_{ij}(x,t))^{2} +
\frac{\lambda_{K}}{2} \int_{\cal D} dtdx\, h(t) (P_{t}
\varphi_{ij}(x,t))^{2} \\ &+\lambda_{M} \int_{\cal D} d\mu \,
\bigl(\partial_t \feat_i(\varphi)+v_j\partial_j \feat_i(\varphi)
\bigr)^2,\\
\end{split}
\label{IntegroDiffEL}
\end{equation}
where the notation $\feat_i(\varphi)$ is used to stress the fact that
$\feat_i$ depends functionally on the filters $\varphi$.  Here, 
if $\lambda_{C}=1$,  the first line is the negative of the mutual information 
and the constants $\lambda_{C}, \lambda_{1}, \lambda_{0}, \lambda_{P},\lambda_{K}$, and
$\lambda_{M}$ are positive multipliers. 
In the above formula, and in
what follows, we will use consistently Einstein summation convention.
This cognitive action can be given two different interpretations.
First, one could think of the regularization terms and on the
motion terms as penalty constraints, so as learning is interpreted
in the classic framework of the MaxEnt principle. 
Second, we can (preferably) think of 
enriching the entropy with the regularization terms in the objective
functions and regard motion term as the only actual constraint.
Furthermore, notice that the mutual information (the first line)
is rather involved, and it becomes too cumbersome to be used with a principle of least action.
However, if we give up to attach the information-based terms their
interpretation in terms of bits, we can rewrite the entropies that define the
mutual information as
\[S(Y\mid X,T,F)\to -\int_D d\mu \, \feat_i^2 \quad \hbox{and}\quad S(Y)\to
-\Bigl(\int_D  d\mu \, \feat_i\Bigr)^2.\]
Interestingly, this replacement does retain all the basic properties on the stationary points
of the mutual information and, at the same time,  it simplifies dramatically the
overall action, which  becomes
\begin{align}
\begin{split}
\label{CognitiveActionEq}
{\cal A}(\varphi)=&
{1\over 2}\Bigl(\int_{\cal D} d\mu \, \feat_i(\varphi) \Bigr)^2
- {\lambda_{C} \over 2}\int_{\cal D}  d\mu \, \feat_i^2(\varphi) \cr
&+{\lambda_{1} \over2}\int_{\cal D} d\mu \, \Bigl(\sum_{i=1}^n\feat_i(\varphi)-1\Bigr)^2-\lambda_{0}
\int_{\cal D} d\mu \, \feat_i(\varphi)\cdot[\feat_i(\varphi)<0]\cr
&+ \frac{\lambda_{P}}{2} \int_{\cal D} dtdx\, h(t)  (P_{x} \varphi_{ij})^{2}
+\frac{\lambda_{K}}{2} \int_{\cal D} dtdx\, h(t) (P_{t} \varphi_{ij})^{2} \cr
&+{\lambda_{M}\over2} \int_{\cal D} d\mu \, \bigl(\partial_t \feat_i(\varphi)+v_j\partial_j \feat_i(\varphi)
\bigr)^2.\cr\end{split}\\
&\feat_i(x,t)= \sigma(\varphi_{kj} \ast \textrm{C}_j)(x,t)=
\sigma \bigg(\int_{\cal X} \varphi_{kj}(x-y,t)\textrm{C}_j(y,t)\,dy\bigg)
\end{align}
In the following analysis we will consider the case in which
$\sigma(\cdot)$ is the identity function, but the extension to the
general case is straightforward. In order to be sure to preserve 
the commutativity of convolution\dash---a property that in general holds when 
the integrals are extended to the entire plane\dash---we have to make assumptions 
on the retina and on
the domain on which the filters are defined. 
First of all assume that ${\cal D}={\cal X}_R\times[0\dts T]$, with
${\cal X}_R=[-R\dts R]\times [-R\dts R]$, $R>0$; we will assume that
$\textrm{C}_i$ has spatial support in ${\cal X}_R$ and it is identically null outside,
while $\varphi_{ij}$ will be taken with spatial support in ${\cal  X}_r$ with
$0<r\le R$ and zero outside ${\cal X}_r$. Under these assumption 
we can guarantee that
the convolution $\varphi_{ij} \ast \textrm{C}_j$ is commutative 
in ${\cal X}_R$. In particular, for all $x\in {\cal X}_R$ we have
\begin{equation}
\begin{split}
\feat_i(x,t)&=(\varphi_{ij} \ast \textrm{C}_j)(x,t)=\int_{{\cal X}_R}\varphi_{ij}(x-y,t) \textrm{C}_j(y,t)\, dy\\
&=\int_{{\cal X}_R}\varphi_{ij}(y,t) \textrm{C}_j(x-y,t)\, dy
=\int_{{\cal X}_r}\varphi_{ij}(y,t) \textrm{C}_j(x-y,t)\, dy \\
&=
(\textrm{C}_j  \ast \varphi_{ij})(x,t).
\end{split}
\label{conv-comm}
\end{equation}
Before studying the stationarity of ${\cal A}(\cdot)$ we 
can conveniently elaborate its functional structure so 
as to get a more direct expression in terms of $\varphi_{ij}$.
In particular, in order to provide an explicit expression of
the motion term we need to introduce a number of 
coefficients that can be computed
whenever we are given the video signal and the optical flow. 
Let us define
\begin{align}
\begin{split}
\textrm{W}_{ml}(\xi,\zeta,\tau)&=\int dz\, f(z,\tau) 
\textrm{C}_m(z-\xi,\tau) \textrm{C}_l(z-\zeta,\tau) \\
\textrm{Y}_{ml}(\xi,\zeta,\tau)&= \int dz\, f(z,\tau)[\partial_\tau 
\textrm{C}_m(z-\xi,\tau)+
v_\alpha \partial_\alpha \textrm{C}_m(z-\xi,\tau)] \textrm{C}_l(z-\zeta,\tau) \\ 
\textrm{H}_{ml}(\xi,\zeta,\tau)&=
\int dz\, f(z,\tau)[\partial_\tau \textrm{C}_m(z-\xi,\tau)+
v_\alpha \partial_\alpha \textrm{C}_m(z-\xi,\tau)] \cdot\\
& \cdot [\partial_\tau \textrm{C}_l(z-\zeta,\tau)
+
v_\beta \partial_\beta \textrm{C}_l(z-\zeta,\tau)].
\end{split}
\end{align} 
In case of still images we can promptly see that only 
$\textrm{W}_{ml}(\xi,\zeta,\tau) \neq 0$. Its value turns out
to be a sort of autocorrelation of the color field, which 
operates over the different channels $m,l$, as well as 
at spatial level between the values at $\xi$ and $\zeta$.
The coefficients $\textrm{Y}_{ml}(\xi,\zeta,\tau)$, 
$\textrm{H}_{ml}(\xi,\zeta,\tau)$ are affected by motion
but have a related autocorrelation meaning. 
Once, we introduce these coefficients, 
the following property can be stated.
\begin{proposition}
Motion term turns out to be a quadratic function of  $\varphi$
and $\partial_{\tau} \varphi$, that is
\footnote{According to a strong interpretation of Einstein notion, we also dropped $\int$ to simplify the notation.}
\begin{align*}
\begin{split}
\omega(\varphi)&=
\frac{1}{2} d\tau d\xi d\zeta\,
\big(
\partial_\tau \varphi_{km}(\xi,\tau) \textrm{W}_{ml}(\xi,\zeta,\tau)
\partial_\tau \varphi_{kl}(\zeta,\tau)
+2  \varphi_{km}(\xi,\tau)
\textrm{Y}_{ml}(\xi,\zeta,\tau)\partial_\tau \varphi_{kl}
(\zeta,\tau)\\
&+  \varphi_{km}(\xi,\tau) \textrm{H}_{ml}(\xi,\zeta,\tau)\varphi_{kl}(\zeta,\tau)
\big).
\end{split}
\end{align*}
\label{SecondOrderMotionEq}
\end{proposition}
The proof arises from plugging expression of the features into
the motion term.
The statement of the Euler-Lagrange equations also benefits from 
defining
\begin{align}
\begin{split}
\Xi_{jk}(x,\xi,t)&=-\textrm{W}_{jk}(\xi,x,t)\\
\Pi_{jk}(x,\xi,t)&=\textrm{Y}_{jk}(x,\xi,t)-\textrm{Y}_{kj}(\xi,x,t)
-\partial_t \textrm{W}_{kj}(\xi,x,t)\\
\Upsilon_{jk}(x,\xi,t)&=\textrm{H}_{kj}(\xi,x,t)-\partial_t \textrm{Y}_{kj}(\xi,x,t).
\end{split}
\end{align}
In addition, based on $\Xi_{jk}(x,\xi,t), \Pi_{jk}(x,\xi,t)$, and
$\Upsilon_{jk}(x,\xi,t)$, we also  introduce
\begin{align}
\begin{split}
c_j(x,t)&:=\int dz\, f(z,t)\textrm{C}_j(z-x,t),\\
\textrm{T}_{jk,im}(x,\xi,t)&:=\lambda_C\Xi_{jk}(x,\xi,t)\delta_{im}+
\sum_{l=1}^n\Xi_{jk}(x,\xi,t)\delta_{\ell m}\\
&+\lambda_M(\Xi_{jk}(x,\xi,t)\del_t^2
+\Pi_{jk}(x,\xi,t)\del_t+\Upsilon_{jk}(x,\xi,t))\delta_{im},\\
\Delta_{jk,im}(x,\xi,t)&:=\textrm{T}_{jk,im}(x,\xi,t)+c_j(x,t)c_k(\xi,t)
\delta_{im},\\
\rho_{ij}(x,t)&:=-\lambda_1 c_j(x,t)-\lambda_0\int dz\, f(z,t)\textrm{C}_j(z-x,t)
[\feat_i(z,t)<0].
\end{split}
\label{defined-quantities}
\end{align}
In what follows we will regard $\rho$ as a function that is independent of
the variables\footnote{Actually
  $\rho$ depends on $\varphi$ through the step function $\(\feat\le0)$,
  so that the precise statement would be that $\rho$
  is independent of $\varphi$ in the regions with definite sign of
  the feature $\varphi$. This can be avoided if we impose the 
  perfect satisfaction of the normalization
  conditions or if we assume a softmax normalization of the features.} 
$\varphi$
We are now ready to express the stationary condition of the 
action~\eqref{CognitiveActionEq}.
\begin{theorem}
  The stationarity conditions of~(\ref{CognitiveActionEq}) leads to
  the following Euler-Lagrange equations in the filters $\varphi_{ij}$
  \begin{equation}\begin{split}
    &\lambda_K
    P_t^\star(h(t)P_t\varphi_{ij}(x,t))+\lambda_P h(t)P_x^\star P_x\varphi_{ij}(x,t)
    +\int d\xi\, \textrm{T}_{jk,im}(x,\xi,t,\del_t)\varphi_{mk}(\xi,t)\\
    &\qquad+\int d\xi d\tau\, c_j(x,t)c_k(\xi,\tau) \varphi_{ik}(\xi,\tau)
    +\rho_{ij}(x,t)=0,\\
  \end{split}\label{ELth1}\end{equation}
where $T$  and $\rho$ are defined in Eq.~(\ref{defined-quantities}).
\label{E-L-E-th}
\end{theorem}

\begin{proof}
The Euler-Lagrange equation of the action arises from $\delta{\cal A}(\varphi)/
\delta \varphi_{ij}(x,t)=0$. So we need to take the variational derivative
of all the terms of  action in Eq.~(\ref{CognitiveActionEq}).
In the following calculation, we will assume  that $d\mu(x,t)=f(x,t)\, dx\,dt$.
The first term yields
\begin{equation}\begin{split}
\int_D \feat_k\, d\mu&\cdot {\delta\over\delta
\varphi_{ij}(x,t)}\int dz\,d\tau\,dy\,f(z,\tau)\varphi_{kj}(y,\tau)
\textrm{C}_j(z-y,\tau)\cr
&=\int dz\, f(z,t) \textrm{C}_j(z-x,t)\cdot\int dz\,d\tau\, d\xi\, f(z,\tau)
\varphi_{ik}(\xi,\tau)\textrm{C}_k(z-\xi,\tau);\cr\end{split}
\label{first-term-variation-non-local}
\end{equation}
while the second term gives
\begin{equation}\begin{split}
&{\delta\over\delta \varphi_{ij}(x,t)}{1\over 2}\int_{\cal D} \feat_k^2(z,\tau)
 f(z,\tau)\, dz\,d\tau=
\int_X dz\, f(z,t)\feat_i(z,t)\textrm{C}_j(z-x,t)\cr
=&\int d\xi\,\Big(\int dz\,f(z,t)\textrm{C}_j(z-x,t)\textrm{C}_k(z-\xi,t) \Big)
\varphi_{ik}(\xi,t).\cr\end{split}\end{equation}
The variation of the third term similarly yields
\begin{equation}\begin{split}
    \sum_{m=1}^{n}\int d\xi\,\Big(\int dz\, f(z,t) \textrm{C}_j(z-x,t) 
    \textrm{C}_k(z-\xi,t)\Big)
    &\varphi_{mk}(\xi,t)\\
    &-\int dz\, f(z,t) \textrm{C}_j(z-x,t).\\
  \end{split}\end{equation}
The variation of the terms that implements positivity is a bit more tricky:
\[\mskip -6 mu\begin{split}
    {\delta\over\delta \varphi_{ij}(x,t)}\int_{\cal D} \feat_k\cdot[\feat_k<0]\,d\mu&=\int
    {\delta\feat_k(z,\tau) \over\delta \varphi_{ij}(x,t)}\cdot[\feat_k(z,\tau)<0]f(z,\tau)\,dz\,d\tau\cr
    &+\int \feat_k(z,\tau)\cdot{\delta[\feat_k(z,\tau)<0]\over
      \delta \varphi_{ij}(x,t)}f(z,\tau)\,dz\,d\tau. \cr\end{split}\]
However, the second term is zero since
\[\begin{split}
&\int \feat_k(z,\tau)\delta[\feat_k(z,\tau)<0]f(z,\tau)\,dz\,d\tau=\int dz\,d\tau\,d\xi\varphi_{km}(\xi,\tau)\textrm{C}(z-\xi,\tau)\cr
&\cdot\Bigl(\big[\int d\xi\, \varphi_{km}(\xi,\tau)\textrm{C}_m(z-\xi,\tau)+\epsilon
\int d\xi\, \delta\varphi_{km}(\xi,\tau)\textrm{C}_m(z-\xi,\tau)<0\big]\cr
&\qquad-\big[\int d\xi\, \varphi_{km}(\xi,\tau)\textrm{C}_m(z-\xi,\tau)<0\big]\Bigr).\cr\end{split}\]
The difference of the two Iverson's brakets is always zero unless the
epsilon-term makes the argument of the first braket have an opposite sign with
respect to the second. Since $\epsilon$ is arbitrary small, this
can only happen if $\int d\xi\, \varphi_{km}(\xi,\tau)\textrm{C}_m(z-\xi,\tau)=0$.
Thus in either cases the whole term vanishes. Hence, we get
\begin{align}
	{\delta\over\delta \varphi_{ij}(x,t)}\int_{\cal D} \feat_k \cdot[\feat_k<0]\,d\mu
=\int dz\, f(z,t)\textrm{C}_j(z-x,t)[\feat_i(z,t)<0].
\label{ProbConsELT}
\end{align}
Finally, the variation of the last term is a bit more involved and yields 
(see Appendix~\ref{motion-invariance-var-append}):
\begin{align}
  \int d\xi\bigl(\Xi_{jk}(x,\xi,t)\del_t^2+\Pi_{jk}(x,\xi,t)\del_t+
  \Upsilon_{jk}(x,\xi,t)
	\bigl)\varphi_{ik}(\xi,t).
\label{MotionELT}
\end{align}
In these calculations we have used intensively the commutative property of the
convolution as stated in Eq.~(\ref{conv-comm}),
which allows us to avoid expressions with an higher degree
of space non-locality.
Then the Euler-Lagrange equations reads:
\begin{equation}\begin{split}
    &\lambda_K
    P_t^\star(h(t)P_t\varphi_{ij}(x,t))+\lambda_P h(t)P_x^\star P_x\varphi_{ij}(x,t)\\
    &+c_j(x,t)\cdot\Bigl(\int d\tau\, d\xi\, c_k(\xi,\tau)
\varphi_{ik}(\xi,\tau)-\lambda_1\Bigr)+\lambda_C\int d\xi\, \Xi_{jk}(x,\xi,t)\varphi_{ik}(\xi,t)\cr
&\qquad -\lambda_1\sum_{m=1}^n\int d\xi\, \Xi_{jk}(x,\xi,t)\varphi_{mk}(\xi,t)
-\lambda_0\int dz\, f(z,t)\textrm{C}_j(z-x,t)[\feat_i(z,t)<0]\cr
&\qquad\qquad+\lambda_M\int d\xi\bigl(\Xi_{jk}(x,\xi,t)\del_t^2
+\Pi_{jk}(x,\xi,t)\del_t+\Upsilon_{jk}(x,\xi,t)
\bigl)\varphi_{ik}(\xi,t)=0,\cr\end{split}\label{EL1}\end{equation}
which can be reduced to Eq.~(\ref{ELth1}).
\end{proof}
\medskip
\noindent
\textbf{Boundary conditions.\enspace} 
In order to be solved, E-L equations Eq.~\ref{ELth1} require the 
definition of the boundary conditions on ${\cal D}$. 
Clearly the mutual information term does not add any
boundary conditions to the E-L equations and, in
Appendix~\ref{motion-invariance-var-append}, we discuss why
also the motion term does not add any conditions on the boundaries.
As we will see in details in the following section, however,
boundary conditions appear that are due to the temporal regularization term. Interestingly, it will be shown that 
the actual solution is made possible by the statistical
regularity of video signals.

\medskip
\noindent
\textbf{Non-locality and ill-position.\enspace} 
This theorem  shows that the EL-equations are non-local
integro-differential equations. Notice that Eq.~(\ref{ELth1}) is non-local
in both spatial (third and fourth terms) and time (fourth term).
This  result suggests that an agent designed
on the basis of Eq.~\eqref{E-L-E-th} would be doomed to fail, since 
its solution is inherent intractable in terms of 
computational complexity.
Basically, the lack of locality, makes Eq.~\eqref{ELth1} unsuitable to 
model the emergence of visual features in nature.
In what follows we will show how to overcome this critical complexity issues
by modifying the position of the problem of visual feature so as to make it 
well-posed.

\medskip
\noindent
\textbf{Temporal locality.\enspace} 
From Eq.~(\ref{ELth1})
we immediately see that the last term is non-local in time; this means that the
equations are non-causal. This is basically due to the need of
knowing the probability of the hidden symbols to determine
the entropy. Formally, the probability of the symbols does 
require to know all the video over the life interval $[0\dts T]$, 
which breaks temporal locality.
This problem can be faced in different ways:
\begin{enumerate}
\item [$i.$] Enforce time locality by computing the entropy 
by splitting the averaging on frames and time as 
follows:
\begin{align}
S(Y)\leadsto
  \int_0^T dt \rev{h(t)}\,\biggl(\int_{\cal X} dx \ g(x-a(t)) \feat_i(x,t) f(x,t)\, \biggr)^2.
\label{time-loc-fix-1}
\end{align}
Clearly this way of splitting the measure $d\mu$ only
approximates the actual entropy of the source.
When averaging at frame level one might get a biased
view on the probability of the symbols that, however, is
somewhat balanced by the temporal average over all the
time horizon.
\item [$ii.$] Let us define the following estimation of the probability 
of symbol $i$ at $t$:
\[
s_i(t)=\int_0^t\, d\tau\int_{\cal X}dx\, \feat_i(x,\tau) f(x,\tau)
= \int_0^t d\tau \, h(\tau) \int_{\cal X}dx\, \feat_i(x,\tau) g(x-a(\tau)).
\]
and express the entropy on the basis of this estimation instead
of the actual value of the probability of symbol $i$ given by
$\int_{\cal D} d\mu \Phi_{i}$. In this way the entropy
term $S(Y)$ in the Lagrangian can be replaced with
\begin{equation}
	\tilde{S}(Y)={1\over T}\int_0^Ts_i^2(t)\, dt+\alpha\int_0^Tdt\, \biggl(s_i(t)-
\int_0^t\, d\tau\int_Xdx\, \feat_i(x,\tau) f(x,\tau) \biggr)^2
\label{average-s-entr}
\end{equation}
where the second term, with an appropriate non-negative $\alpha$ 
is required to enforce the constraint on the
value gained by $s_{i}(t)$. 

\item [$iii.$] Let us consider the above causal entropy $\tilde{S}_{Y}$ 
given by Eq.~\ref{average-s-entr} and enforce a differential form of
the the constraint on $s_{i}(t)$. In doing so,  the entropy
term in the Lagrangian can be replaced with
\begin{equation}
\tilde{\tilde S}(Y)
={1\over T}\int_0^Ts_i^2(t)\, dt+\alpha\int_0^Tdt\, \biggl(\dot s_i(t)-
\int_{\cal X}dx\, \feat_i(x,t) f(x,t) \biggr)^2.
\label{average-s-entr-diff}
\end{equation}
Clearly, in doing so, unlike the formulation based on 
the cognitive action~\ref{CognitiveActionEq},
the corresponding E-L equations that we derive are local in time.
However, we need to involve the auxiliary variable $s_{i}$
in addition to the other Lagrangian coordinates.
\end{enumerate}
Interestingly, $\tilde{S}_{Y}$ offers a consistent asymptotic approximation
of $S(Y)$.  In particular, the following results
connects the two terms.
\begin{proposition}
	 If 
	 $
	 \lim_{t \rightarrow \infty} s_{i}(t) =p_{i}(T):=
         \int_{\cal X}\int_0^T\feat_i(x,t) f(x,t)\, dxdt
	$,
	then
\[\lim_{T\to+\infty}\Bigl|p_i^2(T)-\frac{1}{T}\int_0^Ts^2_i(t)\, dt\Bigr|=0.\]
\end{proposition}
\begin{proof}
	From the hypothesis 
	$\forall \epsilon>0$ there exists  $T_{\epsilon}$
	such that $\forall t> T_{\epsilon}: \ \ |p_{i}-s_{i}(t)| \leq \epsilon$
	\begin{align*}
	\alpha_{S}(T) &
	= \frac{1}{T} \bigg| T p_{i}^{2} -
	\int_{0}^{T} dt s_{i}^{2}\bigg|
	= \frac{1}{T} \bigg| \int_{0}^{T} dt \ p_{i}^{2} -
	\int_{0}^{T} dt \ s_{i}^{2}\bigg|\\
	&=\frac{1}{T} \bigg| \int_{0}^{T}dt \
	(p_{i}+s_{i})(p_{i}-s_{i}) \bigg|
	\leq \frac{1}{T} \int_{0}^{T}dt \
	(p_{i}+s_{i}) | p_{i}-s_{i} | \\
	& \leq 
	\frac{1}{T} \int_{0}^{T_{\epsilon}}dt \
	(p_{i}+s_{i}) | p_{i}-s_{i} |+
	\frac{1}{T}\int_{T_{\epsilon}}^{T}dt \ (2p_{i}+\epsilon)
	\epsilon\\
	&\leq 2\frac{T_{\epsilon}}{T} \epsilon+
	\frac{T-T_{\epsilon}}{T} (2+\epsilon)\epsilon
	<  \bigg(2\frac{T_{\epsilon}}{T} +(2+\epsilon)\bigg)\epsilon.
	\end{align*}
	Now, for any $\delta>0$ the condition $\alpha_{S}(T) < \delta$
	yields
	$
		2\frac{T_{\epsilon}}{T} +(2+\epsilon)\epsilon< \delta
	$
	which is satisfied when choosing
	\[
		\epsilon < \sqrt{\bigg(1
		+\frac{T_{\epsilon}}{T}\bigg)^{2}+\delta}
		-\bigg(1+\frac{T_{\epsilon}}{T}\bigg) 
	\]
	and $T>T_{\epsilon}$.
\end{proof}
We are now ready to see how how the Euler-Lagrange equations
are transformed once time-locality is handled of learning.
In particular, in the following, we consider the case $i$, but
extension to $ii$ and $iii$ are straightforward.
\begin{theorem}
  The functional ${\cal A}(\varphi)$ under the replacement described in
  Eq.~(\ref{time-loc-fix-1}) admits time-local E-L equations, i.e
  Eq.~(\ref{ELth1}) becomes
  \begin{equation}\begin{split}
    &\lambda_K
    P_t^\star(h(t)P_t\varphi_{ij}(x,t))+\lambda_P h(t)P_x^\star P_x\varphi_{ij}(x,t)+\int d\xi\, \Delta_{jk, im}(x,\xi,t,\del_t)\varphi_{mk}(\xi,t)\\
    &
    +\rho_{ij}(x,t)=0,\\
  \end{split}\label{ELth2}\end{equation}
\end{theorem}

\begin{proof}
It is sufficient to replace the variation of the energy term, which is 
now dramatically simplified
\[\frac{\delta}{\delta\varphi_{ij}(x,t)}
\int_0^T d\tau\,\biggl(\int_X \feat_i(z,\tau) f(z,\tau)\, dz\biggr)^2
=c_j(x,t)\cdot\int d\xi\, c_k(\xi,t)
\varphi_{ik}(\xi,t).\]
Finally, the theorem arises when considering the
definitions~\eqref{defined-quantities}.
\end{proof}
It is easy to see that temporal locality can also be gained in the case in which
the entropy is defined according to Eq.~\eqref{average-s-entr-diff}.

\medskip
\noindent
\textbf{Space locality.\enspace}
We will now show how to gain space locality, which is still missing in 
Eq.~\eqref{ELth2}. The intuition is that the lack of space locality is 
inherently connected with the definition of convolutional features, 
whenever one makes no delimitation on the context required to
compute the features. As already pointed when addressing 
motion invariance, while the possible visual configurations increase exponentially with the growth of the measure of ${\cal X}$ ,
the information associated with $\varphi_{ij}$ only grows 
linearly the distance to the focus point.
We will make use of a generalized notion
of {\em receptive field} that, as it will be proven in the following,
allows us to gain spatial locality.

To be more precise assume the following factorization for the filters
\begin{equation}
	\varphi_{ij}(x,t)=G(x)\phi_{ij}(x,t), 
\label{RF-filter-eq}
\end{equation}	
where $G\colon {\cal X}\to \bbR$ is a smooth
function, of typical bell-shape structure.
Notice that this corresponds with expressing the
computation of the features by
\begin{equation}
	\feat_i(x,t) = \int_{\cal X} dy \  
	G(y)\phi_{ij}(y,t) \textrm{C}(x-y,t).
\end{equation}
In so doing, the contribution of the color field at distance
$x-y$ is weighed on the basis of the receptive field structure
induced by bell-shaped function $G$. 
Then the non-local term in Eq.~(\ref{ELth2}) reads
$G(\xi)\Delta_{jk, im}(x,\xi,t,\del_t)\phi_{mk}(\xi,t)$.

\begin{theorem}
Let $G\colon {\cal X}\to \bbR$ be the Green function of an self-adjoint operator
$L$ and let $G(\partial {\cal X})=0$, where $\partial {\cal X}$ denotes
the boundary of ${\cal X}$. Then
Eq.~(\ref{ELth2}) is equivalent to the following (local) system of
differential equations:
  \begin{equation}
    \begin{cases}
      \lambda_K
      P_t^\star(h(t)P_t G(x)\phi_{ij}(x,t))+\lambda_P h(t)P_x^\star P_x G(x)
      \phi_{ij}(x,t)+\Lambda_{ij}(x,0,t)+\rho_{ij}(x,t)=0;\\
      L\Lambda_{ij}(x,\xi,t)=\Delta_{jk, im}(x,\xi,t,\del_t)\phi_{mk}(\xi,t).
    \end{cases}
    \label{ELEsys}
    \end{equation}
\label{RedLoc}
\end{theorem}
\begin{proof}
  Let $\Lambda_{ij}(x,\xi,t)$ be a solution of the differential equation
  \[L\Lambda_{ij}(x,\xi,t)=\Delta_{jk, im}(x,\xi,t,\del_t)\phi_{mk}(\xi,t),\]
  where $L$ is a self-adjoint operator.
  Then the non-local term $d\xi\, G(\xi)\Delta_{jk, im}(x,\xi,t,\del_t)\phi_{mk}(\xi,t)$ becomes
  \begin{equation}
  \int d\xi\, G(\xi)\Delta_{jk, im}(x,\xi,t,\del_t)\phi_{mk}(\xi,t)=
    \int d\xi\, G(\xi) L\Lambda_{ij}(x,\xi,t).
 \label{DiracRedZero}
 \end{equation}
  Now, since $L$ is self-adjoint, we have
  $L^*G=LG=\delta$ and, consequently, we get 
  \begin{equation}
  	\int d\xi\, G(\xi)\Delta_{jk, im}(x,\xi,t,\del_t)\phi_{mk}(\xi,t)=
    	\Lambda_{ij}(x,0,t),
  \end{equation}
which is a local expression in space. Finally, 
Eq.~(\ref{ELth2}) turns out to be  equivalent to  Eq.~(\ref{ELEsys}).
 \end{proof}
These differential equations, along with their boundary conditions,
can be thought of as  information-based laws that dictate the spatiotemporal 
behavior of the visual filters. 
Notice that space locality has been gained at the price of 
enriching the space by the adjoint variable $\Lambda_{ij}$.
It contributes to face and break chicken-egg dilemma 
on whether we first need to define the context for computing the
related visual feature or if the feature does in fact define
also the context from which it is generated. The transformation
of Eq.~(\ref{ELth2}) (integro-differential equations) into 
Eq.~(\ref{ELEsys}) (differential equations) is paid by the 
introducing of the cyclic computational structure of 
Eq.~(\ref{ELEsys}) that, however, is affordable from a 
computational point of view. It is worth mentioning that
from an epistemological point of view, Eq.~(\ref{ELEsys})  
comes from variational principles that very much remind us
the scheme used in physics; for this reason we use the
term information-based laws of visual features. Clearly, 
we can always read these differential equations as a 
computational model of learning visual features.

The following theorem gives insights on the possibility
of finding $G$ and $L$
that  satisfy the properties required by Theorem~\ref{RedLoc} with arbitrary
precision.
  
\begin{theorem}
    Let $G_\sigma(x)$ be a gaussian with variance $\sigma$ and zero mean;
   let $L_\sigma^m:=\sum_{n=0}^m
    (-1)^n(\sigma^{2n}/ 2^n n!) {\nabla^{2n}}$, then $G_\sigma$ and
    $L_\sigma^m$ satisfy the hypothesis of Theorem~\ref{RedLoc}
    if $\sigma$ is chosen small enough. More precisely we have that
    \[
      \lim_{\sigma\to0}\int \bigr(L^m_\sigma G_\sigma(x)\bigl)
      \varphi(x)\, dx=\varphi(0),
\qquad \forall \varphi\in C^\infty_0(\bbR).\]
\label{rec-field-th}
\end{theorem}
\begin{proof}
See Appendix~\ref{Gauss-appendix}
\end{proof}
This result expressed by this  theorem makes the 
reduction of Eq.~\eqref{DiracRedZero} possible in case we
adopt {\em receptive fields}. 
Let $\rho_\sigma(x):= L_\sigma^m G_\sigma(x)$ be. In Appendix~\ref{Gauss-appendix}
we can see that, for a given $m$ we have that
 $\rho_\sigma(x)$ approaches the $\delta$ distribution as 
$\sigma \rightarrow 0$. Basically, we meet the assumption of 
Theorem~\ref{RedLoc} for finite $m$, which is a crucial 
computational issue concerning the adjoint equation
$L\Lambda_{ij}(x,\xi,t)=\Delta_{jq, ip}(x,\xi,t,\del_t)\phi_{pq}(\xi,t)$.
As stated by the theorem, this holds for ``small'' 
$\sigma$, that can be regarded as a receptive field assumption.

 It is interesting to notice that the property claimed in the 
 theorem works also if $G$ is not itself a Green's function 
 but in case it is a linear combination of
 Green's functions evaluated at different points, that is
 \begin{equation}
 	G(x)=\sum_{i=1}^N\alpha_iJ(x-x_i),
\label{KernRep}
\end{equation} 
so as Eq.~\eqref{RF-filter-eq} is in fact quite general in terms of function representation.
 However, it is evident that  as $N$ increases also the number of terms in
 Eq.~(\ref{ELEsys}) does the same, so that it might indicate that the
 resolution of such equations becomes harder.
 
 \begin{figure}
		\centering
		\includegraphics{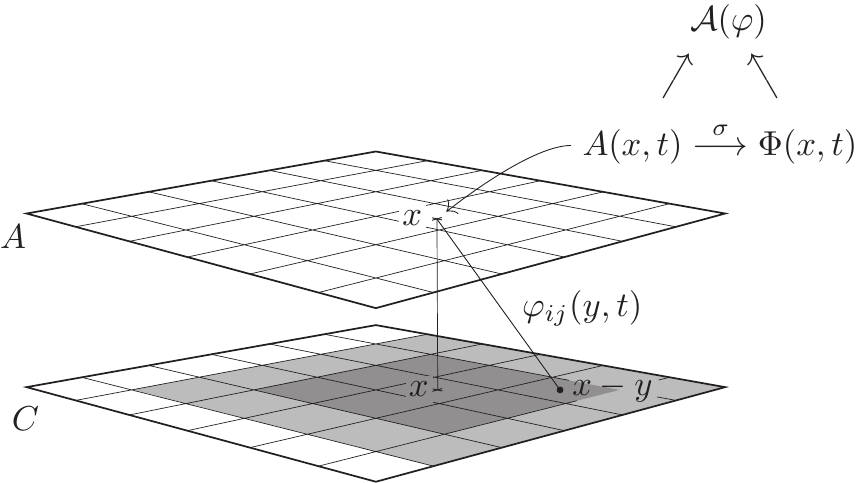}
		\caption{Without explicit constraints in the action that enforces 
			      probabilistic normalization the outcomes of convolution
			      must be remapped with a nonlinear function $\sigma$ (for example a
			      softmax function) in order to be used to built the information based part of 
			      the functional index.}
	\label{ConvFig}
\end{figure}

\medskip
\noindent
\textbf{Softmax formulation and focus of attention.\enspace}\\
Instead of imposing probabilistic normalization implicitly, we
can express the constraints by classic soft-max as follows:
\[A_i(x,t):=\int_{\cal X}\varphi_{ij}(y,t)\textrm{C}_j(x-y,t)\, dy,\qquad
\feat_i(x,t)=\sigma_i(A_1(x,t),\dots, A_n(x,t)),\]
where $\sigma_i(x_1,\dots,x_n):=e^{x_i}/\sum_{k=1}^n e^{x_k}$. 
With this redefinition, the  the information theory based terms of the action
are  automatically well-defined, while the motion
invariance term can still be imposed on the convolutional activations
$A_i(x,t)$.
This formulation therefore it is based on the following action
\begin{align}
\begin{split}
{\cal A}(\varphi)=&
{1\over 2}\Bigl(\int_D d\mu \, \feat_i \Bigr)^2
- {\lambda_{C} \over 2}\int_D  d\mu \, \feat_i^2 \cr
&+ \frac{\lambda_{P}}{2} \int_D dtdx\, h(t)  (P_{x} \varphi_{ij}(x,t))^{2}
+\frac{\lambda_{K}}{2} \int_D dtdx\, h(t) (P_{t} \varphi_{ij}(x,t))^{2} \cr
&+{\lambda_{M}\over2} \int_D d\mu \, \bigl(\partial_t A_i(\varphi)+
  v_j\partial_j A_i(\varphi)
\bigr)^2.\cr\end{split}
\label{CognitiveActionEq-softmax}
\end{align}
that gives rise to EL-equations very related to Eq.~\eqref{ELEsys}. 

%
%


%% file: discrete.tex
So far, we have a field  theory on ${\cal D}$. We can
reformulate it in a discretized retina ${\cal X}^\sharp$, so as
for each point $x$, the filter is defined by the
variable $\varphi_{ijx}(t)$. 
We need to see
how this fields can be re-written on a discretized retina
${\cal X}^\sharp=\{(i,j)\mid 0\le i<\ell, \quad 0\le j<\ell\}$. 
As already noticed, while the filters are characterized by
$\varphi_{ij x}(t)$, the color field will be replaced with
$\textrm{C}_{ix}(t)$.
Notice that, because of the  factorization
$f(x,t)=h(t)g(x-a(t))$, the term $g_x$ in the discretized
formulation is also a function of time, which will turn out to 
contribute to the time dependence that affects the coefficients 
of the differential equation that governs
the evolution of the filters. However, since $g_x$ plays the role of a
probability distribution over the retina, for every $t$, we have 
$\sum_{x\in X^\sharp}g_x=1$. As 
a consequence this yields $\int_{0}^{T} h(t)=1$. Now, 
for each pixel $x$ in the discrete retina, let us define
\[\begin{aligned}
&\gamma^x:=(\textrm{C}_{1(x_1-1)(x_2-1)},\textrm{C}_{1(x_1-1)(x_2-2)},\dots
\textrm{C}_{m(x_1-\ell)(x_2-\ell)})\in \bbR^{m\ell^2},\\
&\chi^i:=(\varphi_{i111},\varphi_{i112},\dots, \varphi_{im\ell\ell})\in \bbR^{m\ell^2};
\end{aligned}\]
similarly let 
$\zeta^x\in \R^{m \ell^2}$ the vector that for each $x$ contains the components
of $v(x,t)\cdot \nabla_x \textrm{C}_j(x-\xi,t)$  with respect to the indexes $\xi$ and  $j$  (for further 
details see Appendix~\ref{vectorization-app}). 
Let $\otimes$ be the Kronecker product.  
We will show that the 
problem and its dynamics can be described in terms of the following matrices
$\textrm{O}^\natural:=(\textrm{O}'\otimes \textrm{I}_{m\ell^2})$,
$\textrm{N}^\natural:=(\textrm{N}'\otimes \textrm{I}_{m\ell^2})$, $\textrm{M}^\natural:=(\textrm{M}'\otimes \textrm{I}_{m\ell^2})$,
where $\textrm{O}_{\alpha\beta}:=\bigl(g_x(\dot \gamma^x_\alpha\dot\gamma^x_\beta
+\zeta^x_\alpha\zeta^x_\beta +2\dot\gamma^x_\alpha\zeta^x_\beta)\bigr)$,
$\textrm{N}_{\alpha\beta}:=\bigl(g_x(\dot \gamma^x_\alpha\gamma^x_\beta
+\zeta^x_\alpha\gamma^x_\beta)\bigr)$ and
$\textrm{M}_{\alpha\beta}:=\dot\chi^i_\alpha\bigl(g_x\gamma^x_\alpha\gamma^x_\beta
\bigr) \delta_{ij}\dot\chi^j_\beta$. These matrices are the discrete 
counterpart of the functions $\textrm{W}$, $\textrm{Y}$, $\textrm{H}$ defined in the previous section.
Let $q$ be the vectorization of tensor $\varphi$ (for a precise
definition see Appendix\ref{vectorization-app}) and let us define
\[
U(q,C)={1\over2}\bigl(
g_x \sigma_i(A_x)\bigr)^2-{\lambda_C\over2}
g_x \Bigl(\sigma_i\bigl(A_x\bigr)
\Bigr)^2.
\]
Then the following result holds:

\begin{proposition}
On the discrete retina the functional 
\[{1\over 2}\Bigl(\int_D d\mu \, \feat_i \Bigr)^2
- {\lambda_{C} \over 2}\int_D  d\mu \, \feat_i^2
+{\lambda_{M}\over2} \int_D d\mu \, \bigl(\partial_t A_i(\varphi)+
  v_j\partial_j A_i(\varphi)
\bigr)^2,\]
which is the Cognitive Action in 
  Eq.~(\ref{CognitiveActionEq-softmax}) without the regularization terms,
becomes
\begin{equation}
{\cal V}(q)= \int_0^T dt \ \tw(t) 
  U\bigl(q,C\bigr)\, dt+ \lambda_M{\cal M}(q) 
\label{cognitive-action-disc-non-reg}\end{equation}
where
\[
{\cal M}(q):=\int_0^T dt\rev{h(t)}\, \left(\frac{1}{2} \dot q \textrm{M}^\natural(t)\dot q
    +q \textrm{N}^\natural(t)\dot q
    +\frac{1}{2} q(t)\textrm{O}^\natural(t)q(t)\right).
\]
\end {proposition}

\begin{proof}
See Appendix~\ref{vectorization-app}
\end{proof}
We will now show that if we pair the
functional~(\ref{cognitive-action-disc-non-reg}) with the
regularization term
\[{\cal R}(q):=\int_0^T \tw(t) dt \ \left(\frac{\alpha}{2} |\ddot q(t)|^2
  +\frac{\beta}{2}|\dot q(t)|^2+\frac{1}{2}|\gamma_1\dot q(t)
  +\gamma_2\ddot q(t)|^2 +\frac{k}{2}|q|^2\right),\]
then the resulting cognitive action
\begin{equation}
\Gamma(q):= {\cal V}(q)+{\cal R}(q)
\label{whole-disc-functional}
\end{equation}
admits a minimum.

In order to understand the peculiar structure of the chosen regularization term
notice that if we pose $\mu=\alpha+\gamma_2^2$,
$\nu=\beta+\gamma_1^2$, $\gamma=\gamma_1\cdot\gamma_2$
 then Eq.~(\ref{whole-disc-functional}) can be rewritten as
\begin{equation}\Gamma(q)=
  \int_0^T \tw\Bigl(\frac{\mu}{2} |\ddot q|^2
  +\frac{\nu}{2}|\dot q|^2+\gamma\dot q\cdot\ddot q
  +\frac{k}{2}|q|^2
  +U\bigl(q,C\bigr)\Bigr)\, dt+\lambda_M {\cal M}(q).\label{cognitive-action-2}\end{equation}
The interpretation of learning by means of functional~(\ref{cognitive-action-2})
is especially interesting since, unlike the case of the classic action in mechanics, 
it admits a minimum under appropriate conditions. 

The following theorem, that is a straightforward
extension of  a results appeared in~\cite{2018arXiv180809162B},
offers an important result on the well-posedness of learning.
\begin{theorem}
	If the following coercivity conditions\footnote{These conditions are indeed equivalent to $\alpha>0$,
        $\beta>0$ and $k>0$.}
	\begin{equation}
        \mu>\gamma_{2}^{2},\quad \nu>\gamma_{1}^{2},\quad k>0
          \label{coerc-cond}
	\end{equation}  
	hold true then functional $\Gamma$,  defined by Eq.~\ref{cognitive-action-2},
	admits a minimum on the set
        \[\mathscr{K}=\{\,q\in H^2((0,T), \bbR^n)\mid q(0)=q^0, \dot q(0)=q^1\,\}.\]
\label{GlobalMinimaRes}
\end{theorem}
\begin{proof}
  The proof is the same as the one in~\cite{2018arXiv180809162B}
  once one observes that
  ${\cal  M}(q)\ge0$  and that it contains at most first derivatives of $q$.
\end{proof}

\medskip\noindent
{\bf Euler-Lagrange Equations.\enspace} For the porpuse of taking the
variation of the functional $\Gamma$ it is convenient to rearrange
it so to have all the terms with at least one derivative all grouped together:
$\Gamma(q)=\Gamma_1(q)+\Gamma_2(q)$ with
\[\Gamma_1(q)=\int_0^T\left(\hat U(q,C)
    +\frac{1}{2} q(t)(\hat O^\natural(t)+\hat k)q(t)
  \right)\, dt,\]
and
\[\Gamma_2(q):=\int_0^Tdt\, \Bigl(\frac{\hat\mu}{2}|\ddot q|^2+
\frac{\hat\nu}{2}|\dot q|^2+\hat\gamma\dot q\cdot \ddot
q+\frac{\lambda_M}{2}\dot q\cdot \hat M^\natural \dot q +\lambda_M q\cdot
\hat N^\natural \dot q\Bigr).\]
We have also introduced the following notation: for any expression $A$ we
let $\hat A(t):=\tw(t) A$. In what follows we will also assume 
\begin{equation}
	\tw(t)=\frac{\theta}{e^{\theta T}-1}e^{\theta}t
\label{DissExpNor}
\end{equation}
with $\theta>0$. In general, $\tw(t)$ needs to be monotone
increasing, so as to yield dissipation and define the time
direction. 
With this factorization it is immediate to see that the variation of
$\Gamma_1$, other than being immediate,
does not give any extra boundary condition. So let us focus on the variation
of $\Gamma_2(q)$.

Let us consider the variation $v$ and define 
$\psi(s)=\Gamma_2(q+sv)$, where $s \in \R$.
In the analysis below, we will repeatedly use the fact
that $v(0)=\dot v(0)=0$. This corresponds with the assignment of the initial values  
$q(0)$ and $\dot{q}(0)$. Since we want to provide a causal computational framework
for $q(t)$, this is in fact the first step towards this direction. 
The stationarity condition for the
functional $\Gamma_2$ is $\psi'(0)=0$, \footnote{Here and in the rest of the paper, we sometimes simplify the notation by removing the explicit dependence on time.}
\[\psi'(0)=\int_0^T dt\, \bigl\{(\hat\mu\ddot q+\hat \gamma\dot q)\cdot\ddot v
  +[(\hat\nu+\lambda_M\hat M^\natural)\dot q+\hat\gamma\ddot q+\lambda_M
  (\hat N^\natural)'q]\cdot\dot v+\lambda_M\hat N^\natural \dot q\cdot v\bigr.\}
\]
With a few integration by parts we get
\begin{align}
  \begin{split}
\psi'(0)=&\big[(\hat\mu\ddot q+\hat \gamma\dot q)\dot v+
\big((\hat\nu+\lambda_M\hat M^\natural)\dot q+\hat\gamma\ddot
q+\lambda_M (\hat N^\natural)'q -(\hat\mu\ddot q+\hat \gamma\dot q)\dot{ }\big)v\big]_{t=T}
\\
&+\int_0^T \big\{(\hat\mu\ddot q+\hat \gamma\dot q)\,\ddot{ }\,
- \big((\hat\nu+\lambda_M\hat M^\natural)\dot q+\hat\gamma\ddot
q+\lambda_M (\hat N^\natural)'q\big)\dot{ }\, +\lambda_M\hat N^\natural \dot q\big\}\cdot v
\end{split}
\label{ByPart-CAL-eq}
\end{align}
As it often happens in variational calculus we proceed as follows:

\begin{enumerate}
\item Consider only the variations such that $v(T)=\dot v(T)=0$. In this case
  $\psi'(0)=0$ yields the following differential equations
\begin{equation}\begin{split}
  \hat\mu q^{(4)}+2\dot{\hat\mu} q^{(3)}&+(\ddot{\hat\mu}+\dot{\hat \gamma}-\hat \nu
  -\lambda_M\hat M^\natural)\ddot q\\
  &+(\ddot {\hat \gamma}-\dot{\hat\nu}-\lambda_M
  (\dot{\hat M}^\natural+(\hat N^\natural)'-\hat N^\natural))\dot q-\lambda_M
  (\dot{\hat N}^\natural)'q=0.\end{split}
\label{appendix-e-l-eq}
\end{equation}
\item Because of Eq.~(\ref{ByPart-CAL-eq}), $\psi'(0)=0$ reduces to
  $\big[(\hat\mu\ddot q+\hat \gamma\dot q)\dot v+
\big(\hat\nu\dot q+\hat\gamma\ddot
q -(\hat\mu\ddot q+\hat \gamma\dot q)\dot{ }\big)v\big]_{t=T}=0$. Moreover,
since $v(T)$ and $\dot v(T)$ can be chosen independent one of each other, then
the vanishing of the first variation also implies that
\begin{equation}
\begin{split}
  &\hat\mu\ddot q(T)+\hat\gamma\dot q(T)=0;\\
  &-\hat\mu q^{(3)}(T)-\dot{\hat\mu}\ddot q(T)+(
  \hat\nu-\dot{\hat\gamma}+\lambda_M\hat M^\natural) \dot q(T)
  +\lambda_M(\hat N^\sharp)'q(T)=0.
 \end{split}\label{neumann-cond}
\end{equation}
\end{enumerate}
We summarize the previous analysis in the statement of
the following theorem:

\begin{theorem}
  The Euler-Lagrange equation relative to the functional $\Gamma(q)$ defined
  on $\mathscr{K}$ are
  \begin{equation}\begin{split}
  \hat\mu(t) q^{(4)}(t)+2\dot{\hat\mu}(t) q^{(3)}(t)+Z_2(t)\ddot q(t)
  +Z_1(t)\dot q(t)+Z_0(t)q(t) +\nabla_q \hat U(q,C)=0.\end{split}
\label{Full-EL}
\end{equation}
where
\begin{gather}
    Z_2=\ddot{\hat \mu}+\dot{\hat \gamma}-\hat \nu
  -\lambda_M\hat M^\natural,\qquad
Z_1=\ddot {\hat \gamma}-\dot{\hat\nu}-\lambda_M
(\dot{\hat M}^\natural+(\hat N^\natural)'-\hat N^\natural),\\
Z_0=\hat k+\lambda_M\hat O^\natural-\lambda_M
  (\dot{\hat N}^\natural)',\end{gather}
together with the boundary conditions in Eq.~(\ref{neumann-cond}).
\end{theorem}
It is worth mentioning that the above theorem holds also if we  redefine $\Gamma(q)$ by arbitrary functions $\hat\mu(t)$,
$\hat\nu(t)$, $\hat\gamma(t)$, and $\hat k(t)$. 
This is one of the key observations that will allow us to devise a mechanism through which
we will be able to have Eq.~(\ref{neumann-cond}) automatically satisfied during the learning.

\medskip\noindent
{\bf Boundary conditions.\enspace} 
The solution of the forth-order differential equation on the
filter parameters requires the satisfaction of the boundary 
conditions~(\ref{neumann-cond}). The underlying idea that drives
the learning process is that one is expected to 
solve the problem of determining the filters in a causal way,
which corresponds with imposing 
Cauchy's initial condition. 
However, the solution of Eq.~(\ref{Full-EL}) under Cauchy's initial
condition will not, in general,
satisfy conditions~(\ref{neumann-cond}) at the end of learning.
Hence, we get into a dilemma that involves the choice of the
initial conditions, since the values 
$q(T),\dot{q}(T),\ddot{q}(T),q^{(3)}(T)$ do depend on the 
video signal in $(0\dts T]$, that is on the ``future.''
We can break the dilemma when pairing a couple of important
remarks: First, a special case in which conditions~(\ref{neumann-cond})
are satisfied is whenever we have still images at $T$, so as 
$N^\sharp=0$, 
\begin{equation} 
	\dot{q}(T)=\ddot{q}(T)=q^{(3)}(T)=0.
\label{ZeroBorderCond}
\end{equation}
Second, without limitations of generality, the color field 
$\textrm{C}(x,t)$ in ${\cal D}$  will always contain brief portions of
null signal. Moreover, its eventual manipulation with the purpose of injecting
brief portions of null signal does not change its information structure,
so as one can reasonably regard the visual environment with such
a manipulation equivalent with respect to the one from which it
is generated. The intuition is that such a ``reset'' of the video 
results in $N^\sharp=0$ and, moreover, the null signal also affects
the differential equation of learning~\ref{Full-EL} by resetting the
dynamics, so as $\dot{q}(T)=\ddot{q}(T)=q^{(3)}(T)=0$ is also
very well approximated. Hence, no matter what are the initial conditions,
it turns out the we can satisfy conditions~(\ref{neumann-cond}) in 
small portions of the video. 

%
%
Now, we will translate this intuition into a formal statements.
Let us consider a sequence
of times $0<t_0<t_1<t_2<\cdots<t_{2N}<T$ that defines the two sets
$A=\bigcup_{i=0}^{N}A_i$ with $A_i=(t_{2i-1}\dts t_{2i})$, $t_{-1}=0$
and $B=\bigcup_{i=0}^{N} B_i$ with $B_i=(t_{2i}\dts t_{2i+1})$, $t_{2N+1}=T$.
Suppose furthermore that we modify the video signal in the following way
$\textrm{C}(x,t)\to \textrm{C}(x,t)\(t\in A)$, so that it is identically null on $B$.
As already pointed out, in doing so, we do not change the
problem of discovering visual features, since we just dilute the
information that is contained in $C$.  On the other hand, 
whenever $\textrm{C}=0$, this results into
a remarkable simplification of the system dynamics in $B$: 
the potential $U$ and all the terms coming from the motion invariance
term (the ones proportional to $\lambda_M$) are identically zero. 
Moreover, since the EL equations still holds true for time-variant
coefficients $\hat\mu(t)$, $\hat\nu(t)$, $\hat\gamma(t)$, and $\hat k(t)$,
we can always decouple the dynamics so that whenever $t\in B$
Eq.~(\ref{Full-EL}) becomes (see~\cite{2018arXiv180809162B})
\begin{equation}
  \bar\mu q^{(4)}+2\bar\theta\bar\mu q^{(3)}+(\bar\theta^2\bar\mu
+\bar\theta\bar\gamma-\bar\nu)\ddot q
+(\bar\theta^2\bar\gamma-\bar\theta\bar\nu)\dot q+ \bar kq = 0,
\, \qquad t\in B
\label{dyn-in-B}
\end{equation}
where $\bar \theta$, $\bar\mu$, $\bar \nu$ and $\bar\lambda$ are
arbitrary constants  different from
$\theta$, $\mu$, $\nu$ and $\lambda$.
In particular the following Theorem
guarantees us that $\bar \theta$, $\bar\mu$, $\bar \nu$ and $\bar\lambda$
can be chosen  in such a way that the boundary conditions 
in Eq.~(\ref{neumann-cond}) are satisfied at the end of each 
$B$ interval.
\begin{theorem}
  We can always choose the system parameters of Eq.~(\ref{dyn-in-B}) 
  in such a way that  $|q^{(k)}(t_{2i+1})|=0$, $k=1,2,3$, 
  up to an arbitrary precision for  $i=0,1,\dots, N$ 
  regardless of the initial Cauchy  conditions,
  which is in fact a special way of satisfying boundary conditions~(\ref{neumann-cond}).
   \label{vanishing-derivatives}
\end{theorem}
\begin{proof}
See \cite{2018arXiv180809162B} for the proof.
\end{proof}
The intuition behind this result is that the dynamical system defined
by~(\ref{dyn-in-B}) becomes asymptotically stable under an appropriate
choice of the parameters, which corresponds with driving the 
dynamics to a reset state arbitrarily fast.

Another important property of the dynamics in the $B_i$ is that it
we can arrange things in such a way that it 
does not alter the solution found in the previous $A_j$. 
More precisely, let $(0,\lambda_2,\lambda_3,\lambda_4)$
be the roots of the characteristic polynomial associated with
Eq.~(\ref{dyn-in-B}) and let $V_{3}=V(\lambda_{2},\lambda_{3},\lambda_{4})$
be the Vandermonde matrix associated with the $\lambda_i$
eigenvalues. The the following theorem holds.
\begin{theorem}
	Let $\Lambda=(V(\lambda_2/\rho,\lambda_3/\rho,\lambda_4/\rho))^{-1}$ be.
  	For every even $i=0,\dots, 2N$ consider the defined sets
  	$A_i=(t_{i-1}\dts t_i)$, $B_i=(t_i\dts t_{i+1})$.
  	It is always possible to choose the coefficients in Eq.~(\ref{dyn-in-B})
  	such that $\forall \epsilon>0$, if we choose 
	\[
		\rho > [(9C/\epsilon)  \cdot\max_k |q^{(k)}(t_i)|]^{1/2} >1
	\]
         we have $|q(t_{i+1})-q(t_i)|<\epsilon$, 
         where  $|\Lambda_{kj}|\le C$ for all $k$ and $j=1,2,3$.
\label{MemoryDNS}
\end{theorem}
\begin{proof}
See \cite{2018arXiv180809162B} for the proof.
\end{proof}

\medskip\noindent
{\bf System dynamics.\enspace} Here we will mainly focus on the ``free dynamics''
$C\equiv 0$. This particular case is particularly important since
it is possible to analyze this case in details, and it gives us 
insights on the solutions depending on the choice of the parameters.
Let $\chi(x)=x^4+bx^3+cx^2+dx+e$ be the characteristic polynomial of the
EL equation~(\ref{Full-EL}) with $U\equiv0$ (which is just the same as
Eq.~(\ref{dyn-in-B}) only with the unbarred variables); here
we assume $\mu\ne 0$ and use the notation $b=2\theta$,
$c=(\theta^2\mu+\theta\gamma-\nu)/\mu$, $d=(\theta^2\gamma-\theta\nu)/\mu$,
and $e=k/\mu$.

    If we replace $x=z-b/4$ with $\chi(x)$ then
    we obtain the reduced quartic equation
    $\zeta(z):=\chi(z-b/4)=z^4+q z^2+r z+s=0,$
where
$q=c-3 b^2/8, \quad r=b^3/8-bc/2+d,\quad s=
b^2c/16-3/256 b^4-bd/4+e$.
    
Then one can prove (see~\cite{2018arXiv180809162B}) that the following proposition holds:
\begin{proposition}
    If we choose $\theta, \mu, \nu, \gamma_1,\gamma_2, k$ such that
    $\theta>0$ and:
    \begin{align}
    \begin{split}
    	&\mu>\gamma_2^2,\quad \nu>\gamma_1^2,\quad \nu<\theta
      	\gamma_1\gamma_2,\quad
        0<k\le\frac{(\nu-\theta\gamma_1\gamma_2)^2}{4\mu} \\
    	&\gamma_1<0,\quad\gamma_2<\frac{\gamma_1}{\theta}\quad \hbox{or}\quad
    \gamma_1>0,\quad\gamma_2>\frac{\gamma_1}{\theta}.
    \end{split}
    \end{align} 
    then the following conditions are jointly verified:
    \begin{enumerate}
    \item $\Gamma$ admits a minimum in $\mathscr{K}$;
    \item the homogeneous equation associated with Eq.~(\ref{dyn-in-B}) has the following two 
    	properties:
    	\begin{enumerate}
     	\item [$i.$] it is asymptotically stable;
    	\item [$ii.$]it yields aperiodic dynamics  (the roots of the characteristic polynomial are real).
	\end{enumerate}
      \end{enumerate}
      \label{prop:coef}
\end{proposition}


%% file: lig.tex
We are now in condition to partially address  the questions
raised in Section~\ref{inq-vis} in the light of the proposed theory.
Some questions are quite general and can be addressed by
arguments based on the literature. Others are more specific,
and can be answered by relying on the results derived in this paper from
the principle of least cognitive.

\medskip
\noindent
\textit{Q1,Q2: How can humans conquer visual skills 
without requiring ``intensive supervision''?}
\textit{How can animals gradually conquer visual skills in 
a truly temporal-based visual environment?} \\
These questions are stimulating many 
debates especially on the evolution of computer vision.
At the light of the results that arise from the principle of least cognitive action,
we can offer a novel view that emerges from the proposed theory.   
The given information-based equations of learning shows that any visual agents
can conquer visual skills without requiring $(Q1)$ ``intensive supervision.'' 
In particular, the Euler-Lagrange differential equations that dictate the agent life 
only process visual streams without any supervision, so as they represent a 
fully-unsupervised method of feature generation$(Q2)$. 
The development of methods that learn from visual streams without supervision
opens the doors towards
a radically different approach to large visual (labelled) repositories, since visual sources
are virtually infinite. The minimization of the 
motion invariance term $\omega(\phi)$ (see Proposition~\ref{SecondOrderMotionEq})
over the life interval enforces visual 
consistency, which turns out to be a sort of virtual supervision offered by nature for
free. The agent interaction with the environment can, later on, at different stage of development,  benefit from a number of different forms of supervision that can refine the features developed according to the proposed scheme. Unlike most approaches from machine learning, in this paper
the role of time is of crucial importance. 

\medskip
\noindent
\textit{$Q3$: Can animals see in a world of shuffled frames?} \\
Shuffling video frames  would likely drive into 
a ``cul de sac'' for animals and humans.
\rev{The way in which data is presented to humans has also been proved
to influence the quality and the rate of learning
\cite{krueger2009flexible, skinner1958reinforcement, huettel2002perceiving}
to the extent that we can regard visual information to be the teaching plan
offered by nature for the development of visual skills.}
The proposed theory \rev{exploits such coherent information selection to}
imposes motion invariance by enforcing the minimization 
of $\omega(\phi)$, that is null only in case the adiabatic condition 
$d\textrm{C}_{i}/dt=0$ holds true. The approximation of this 
condition along with information and parsimony-based principles  
is  the outcome of the visual laws given by Eq.~\eqref{ELEsys}.

\medskip
\noindent
\textit{Q4: How can humans attach semantic labels at pixel level?} \\
The given laws of feature development (Eq.~\eqref{ELEsys})
inherently operate at pixel level, so as to generate visual features capable of
supporting the recognition process by semantic-based labels attached to single pixels. 
The convolutional filters are perfectly
suited to support the generation of pixel-based features. However, it is worth
mentioning that the typical pooling mechanisms that is paired with deep
convolutional network, as well as their classic overall architecture,
typically dismiss this strict association with pixels, since they are mostly charged
of returning object recognition and related high-level tasks. 
A different path has been more recently followed with the purpose of performing
semantic label~\cite{DBLP:journals/cviu/GoriLMM16}, which is much more aligned with the research reported in this paper.
There are in fact tight links between convolutional networks for segmentation and the
model proposed in this paper. However, motion invariance is supposed to be the main
difference, and it is expected to facilitate the tasks of attaching semantic labels at pixel level.

\medskip
\noindent
\textit{Q5: What could drive the functional difference between the
ventral and dorsal mainstream in the visual cortex?} \\
The discussion on the role of motion invariance also indicates the reason 
why we need distinct feature models
depending on the visual purpose. In particular, motion invariance
turns out to be useful for recognition tasks, whereas such a 
property must not hold for neurons involved in motion control. 
In the light of the theory proposed in this paper, we conjecture that
Eq.~\eqref{ELEsys} might be adequate to simulate
the behavior of neurons in both the mainstreams, the difference being that
the ``what'' neurons do require the term arising from the minimization of
motion term $\omega(\varphi)$, whereas the ``where'' neuron must
not possess such a dynamic invariant property. As a consequence, the
``where'' neurons satisfy Eq.~\eqref{ELEsys}, where the terms deriving from
motion term are removed. Clearly, one might have a gradual transition
between these two different functional requirements. Hence, the 
theory suggests that the dorsal and the ventral mainstream 
are basically processing motion information a in remarkable different way.

\medskip
\noindent
\textit{Q6: Why do we need a hierarchical architecture with receptive fields?} \\
The spatial non-locality of the Euler-Lagrange equations
discussed in the previous section (Eq.~\eqref{EL1})
suggests that efficient solutions cannot be discovered 
without making assumptions on the convolutional filters.
Interestingly, the introduction of adjoint variables makes
is possible to remove spatial locality. 
In particularly, the interpretation of Theorem~\ref{rec-field-th}, leads us to conclude that
only peaked bell-shaped filters following Eq.~\eqref{RF-filter-eq}
can give rise to solutions of Eq.~\eqref{ELEsys}.
Clearly, this suggests the adoption of deep architectures that can
conquer computational capabilities thanks to their compositional structure.
An alternative, that is likely to be less effective, is that of thinking of  a
shallow architecture based on~\eqref{KernRep}.

Hence, the spatial locality property of Eq.~\eqref{ELEsys} needs to be
paired with special filter selections that, to some extent, 
correspond with the assumption of processing by receptive fields
~\cite{Hubel62},\cite{Moody88},\cite{Stokbro90},\cite{hubel:monkey}.
This is one of the most significant outcome of the theory, which
prescribes the presence of receptive fields on the basis the 
derived information-based laws, and gives guidelines also on the 
relationship between the dimension of the receptive field and the
order of the associated differential equation. A related analysis
can be found in~\cite{DBLP:journals/corr/LuoLUZ17}
Notice that neurons with response peaked on their receptive fields 
only react to pixels close to the focus of attention, while spatial regularization
contribute to provide smooth response, a process that is related to
pooling. 

\medskip
\noindent
\textit{Q7: Why do animals focus attention?}\\
In the light of the theory, the reasons for modeling the focus of attention come
directly from the information-based framework along with the
ergodic assumptions. The measure $d\mu = h(t)g(x-a(t))$
suggests that information-based indexes, like the mutual information,
need to be estimated by focusing attention on those areas where
there is more information. As pointed out just after the definition
of $\mathscr{M}_{(x,t)}(\cdot)$ (see Eq.~\eqref{M-fun-inv}), in this paper
we assume that the trajectory of the focus of attention $a(t)$ is
given along with the optical flow, and that 
motion invariance is enforced for any point of the retina. Hence
the role of the focus of attention trajectory is limited to offer an
appropriate estimation of the information-based indexes, whereas
it does not play a direct role in the enforcement of motion invariance,
which is the subject of future investigation.
 
\medskip
\noindent
\textit{Q8: Why do foveal animals perform eye movements?}\\
As already pointed out, the role of focus of attention is crucial to 
provide a good measure of visual informativeness. Hence, the appropriate
positioning on most informative areas requires the movement 
along $a(t)$, which corresponds with eye movements in animals.
The adoption of focus of attention trajectories can give further insights on 
learning visual features. Peaked filters can be updated only in regions that are close
to the area of focus, which dramatically simplifies the learning process, since
this leads to remove the inconsistencies deriving from the presence of multiple
weights updating, that is inherently connected with translation-invariance
of the convolutional computation. 
In a sense, focus of attention is a mechanism that separates 
learning  and inference; the weight updating takes place  where we focus
attention, while inference clearly takes place all over the retina. 


\medskip
\noindent  
\textit{Q9: Why does it take 8-12 months for newborns to achieve adult visual acuity?} \\
As already noticed, while the transformation $\textrm{C} \rightarrow 
\textrm{C}_{b}=\rho \textrm{C}$, that reduces the level of the signal,
does not destroy information in the continuum setting of computation, 
as the video is 	quantized, no matter how many bits we choose,
for arbitrarily small values of $\rho$ the signal $\textrm{C}_{b}$ can be significantly
blurred. In the continuum setting of computation, a similar effect is due to the 
inherent presence of noise. 
Interestingly, the blurring of the video turns out to facilitate the satisfaction of the 
boundary conditions~\eqref{neumann-cond}, since the dynamics is driven in such a 
way that the video signal and the derivatives of the weights are close to zero.  
More importantly, the quantization errors in the numerical solution of Euler-Lagrange
differential equations, as well as the noise effect in the continuum counterpart, are 
dramatically reduced by the described video blurring. A related process
seems to be behind the newborns behavior during their evolutive process of 
vision acquisition. It is worth mentioning that the scale transformation
$\textrm{C} \rightarrow \textrm{C}_{b}=\rho \textrm{C}$ is not the only way of
supporting a system dynamics that is nominally compatible with the boundary conditions. 
If the signal $\textrm{C}$ undergoes a spatiotemporal low-pass filtering computation
we end up in similar conclusions.
Notice that video blurring, which induces a 
deformation into the Lagrangian,
favors the development of a technically sound solution, and it is
at the same time fully compatible with the problem of learning in
visual environments. The reason is that humans and machines 
must do their best to extract visual cues also in presence of 
fog or in dark environments. Hence, a learning process focussed 
on information extraction in case of blurred video turns out to be 
useful for the kind of visual skills that are ordinarily required.
A strong property that is acquired by visual agents carrying out 
a learning trajectory that is {\em nominally compatible} 
with the boundary conditions is that they are {\em causal agents}.
This is a straightforward consequence of the fact that the 
conditions~(\ref{ZeroBorderCond}) are nominally verified at any
time of the agent life.   The causality of the agent is of crucial 
importance for the learning process, since causality  
has important consequences on statistical consistency. 
Moreover, as already stated, the described blurring process plays
a crucial role to make the Euler-Lagrange differential equations
of learning very-well conditioned.  This allows us to set up
a numerical framework with strong precision and, at the same time, 
with limited computational burden. 
 An accurate analysis on the dynamics of $\rho$
clearly indicates that the choice of the damping factor $\eta$ plays a crucial role. 
One very much would like to synchronize the end of blurring with 
the end of learning, that must be somewhat connected with the 
statistical structure of the given visual environment. 


\medskip
\noindent
\textit{Q10: Causality and Non Rapid Eye Movements (NREM) sleep phases} \\
Interestingly, while the described blurring process facilitates the
satisfaction of the boundary conditions and makes the Euler-Lagrange
equations well-conditioned, in humans, the mechanisms behind the saccadic
movements and the day-night human alternation rhythm
might play an important role in the conquering of visual skills.
In both cases, there are temporal portions in which the video signal
is null which, as pointed out in the previous section, facilitates the
satisfaction of the boundary conditions. Basically, in humans, 
blurring, saccadic movements, and  day-night human alternation rhythm
contributes to enforce learning trajectories that are glued to the boundary 
conditions naturally. 
Notice that a ``null video'' seems to correspond with NREM phases of sleep,
in which also eye movements and connection with visual memory are nearly absent. 
The proposed theory suggests that during this phase of sleep, the reset of the video
strongly facilitates the satisfaction of the boundary conditions 
Interestingly, the Rapid Eye Movements (REM) phase is, on the opposite, similar to ordinary visual processing, the only difference being that the construction of visual features during 
the dream is based on the visual internal memory representations. 

Finally, notice that the overall  dynamics very much depends on $h$ (e.g. $e^{\theta t}$). 
Its consequence is that for an effective 
generalization process to take place, one needs to establish a 
very slow dissipative process to guarantee that the weighting term $h$ does not 
penalize too much back in time.  This is of crucial importance in order to
give the agent the capability of storing enough information for its decision.



%% file: exp.tex


We implemented a solver for the differential equation of Eq. (\ref{Full-EL}) that is based on the Euler method. After having reduced the equation to the first order, the variables that are updated
at each time instant are $q$, $\dot q$, $\ddot q$, and $q^{(3)}$. The code and data we exploited to run the following experiments can be downloaded at \rev{\url{http://www.dii.unisi.it/~melacci/calnn/nn_code_data_params.zip}}, together with the full list of model parameters.

\rev{First (Section \ref{exp1}), we evaluated the quality of the visual features extracted by the proposed approach in different real-world videos. This evaluation involves an in-depth experimental analysis that is aimed at concretely studying the role of the kinetic terms and the role of motion coherence, also in the case of multi-layer convolutional architectures. Then (Section \ref{exp2}), we considered a significantly larger set of real-world videos of car driving scenes, comparing the features learned by the proposed model with those of Sparse Convolutional Autoencoders trained in an online manner using gradient-based updates. In particular, we considered a Semantic Labeling task and measured the quality of the predictions of the semantic class to which each pixel of the scene belongs.}

\subsection{\rev{Visual Features}}
\label{exp1}

We randomly selected two real-world video sequences from the Hollywood Dataset HOHA2 \cite{marszalek09}, that we will refer to as ``skater'' and ``car'', and a clip from the movie ``The Matrix'' (\textcopyright  Warner Bros. Pictures). The frame rate of all the videos is $\approx$ 25 fps, so we set the step-size of the Euler method to $1/25$, and each frame was rescaled to $240\times110$ and, for simplicity, it was converted to grayscale. Videos have different lengths, ranging from $\approx 10$ to $\approx 40$ seconds, and they were looped until $45,000$ frames were generated, thus covering a significantly longer time span. 
We randomly initialized the variable $q$ for $t=0$, while the derivatives $\dot q$, $\ddot q$, and $q^{(3)}$ were set to $0$. We used the softmax function to force a probabilistic activation of the features, as suggested in Section~\ref{Cognitive-action-section}, Eq. (\ref{CognitiveActionEq-softmax}), and we computed the optical flow $v$ using an implementation from the OpenCV library. Convolutional filters cover squared areas of the input frame, and we set $g_x$, for each $x$, to be the inverse of the frame area, i.e., we assume that we have a uniform distribution over the retina.
All the results that we report are averaged over 10 different runs of the algorithms.

The video is presented gradually to the agent so as to favour the acquisition of small chunks of information. We start from a completely null signal (all pixel intensities are zero), and we slowly increase the level of detail and the pixel intensities, in function of $\tau(t) \in [0,1]$, where $\tau(t) = 0$ leads to null signal and $\tau(t) = 1$ to full details. In detail, 
$$C(x,t) = \tau(t) \left[ G_{(1- \tau(t))  \delta} \ast \overline{C}(x,t)\right],$$
where $\ast$ is the spatial convolution operator, $\overline{C}(x,t)$ is the original source video signal, $G_{\sigma}$ is a Gaussian filter of variance $\sigma$, and $\delta > 0$ is a customizable scaling factor, that we set to the size of the squared discrete Gaussian filter mask. It is easy to see that for $\tau(t)=1$ we get $C(x,t) = \overline{C}(x,t)$.
We start with $\tau(0)=0$, and then $\tau(t),\ t > 0,$ is progressively increased as time passes with the following rule, 
\begin{equation}
\tau(t+1) = \tau(t) + \eta(1-\tau(t)),
\label{eq:blurplan}
\end{equation} 
where we set $\eta = 0.0005$. 
We refer to the quantity $1-\tau$ as the ``blurring factor'', being it proportional to the variance of the Gaussian blur.

In order to be able to (approximately) satisfy the conditions in Eq.~(\ref{neumann-cond}) we need to keep the derivatives small, so we implement a
``reset plan'' according to which the video signal undergoes a reset
whenever the derivatives become too large. 
Formally, if $\|\dot q(t')\|^2\geq\epsilon_{1}$, or $\|\ddot q(t')\|^2\geq\epsilon_{2}$, or $\|q^{(3)}(t')\|^2\geq\epsilon_{3}$ then we set to $0$ all the derivatives, and we also force $\tau(t')$ to $0$, leading to null video signal, as described above.
We used $\epsilon_j=300 \cdot n$, for all $j$.

\begin{figure}
\centering
\includegraphics[width=0.45\textwidth]{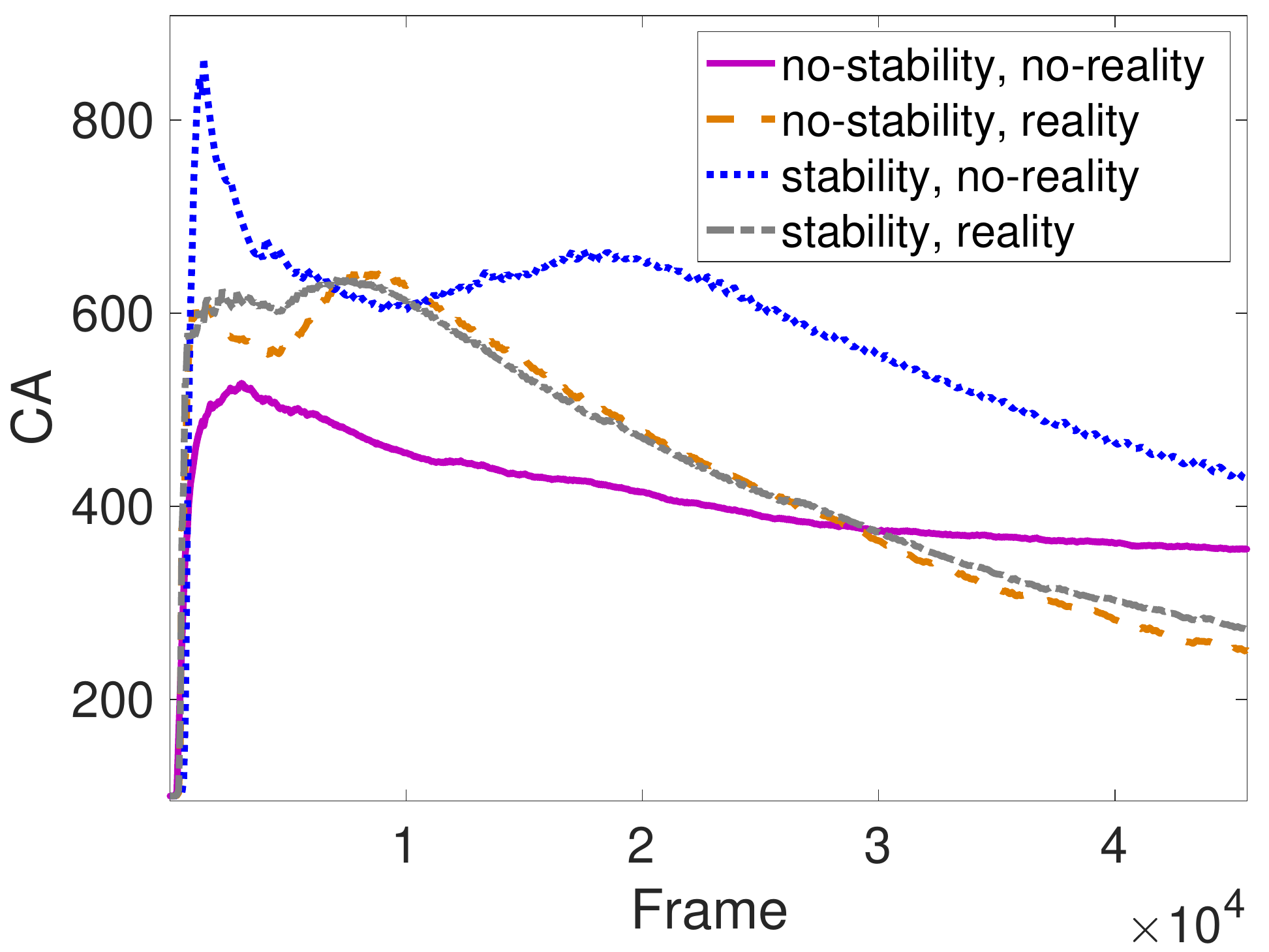}
\includegraphics[width=0.45\textwidth]{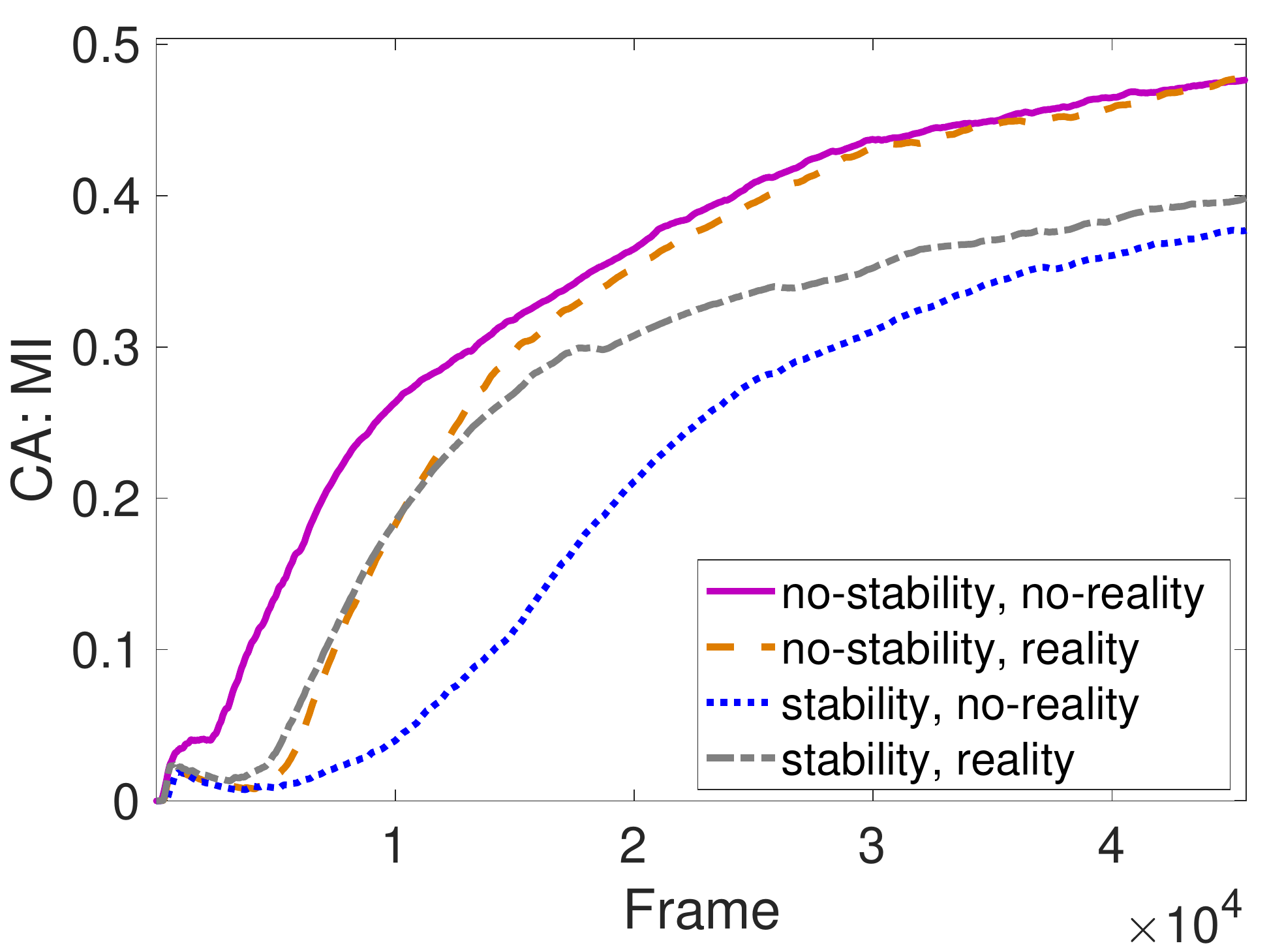}\\
\includegraphics[width=0.45\textwidth]{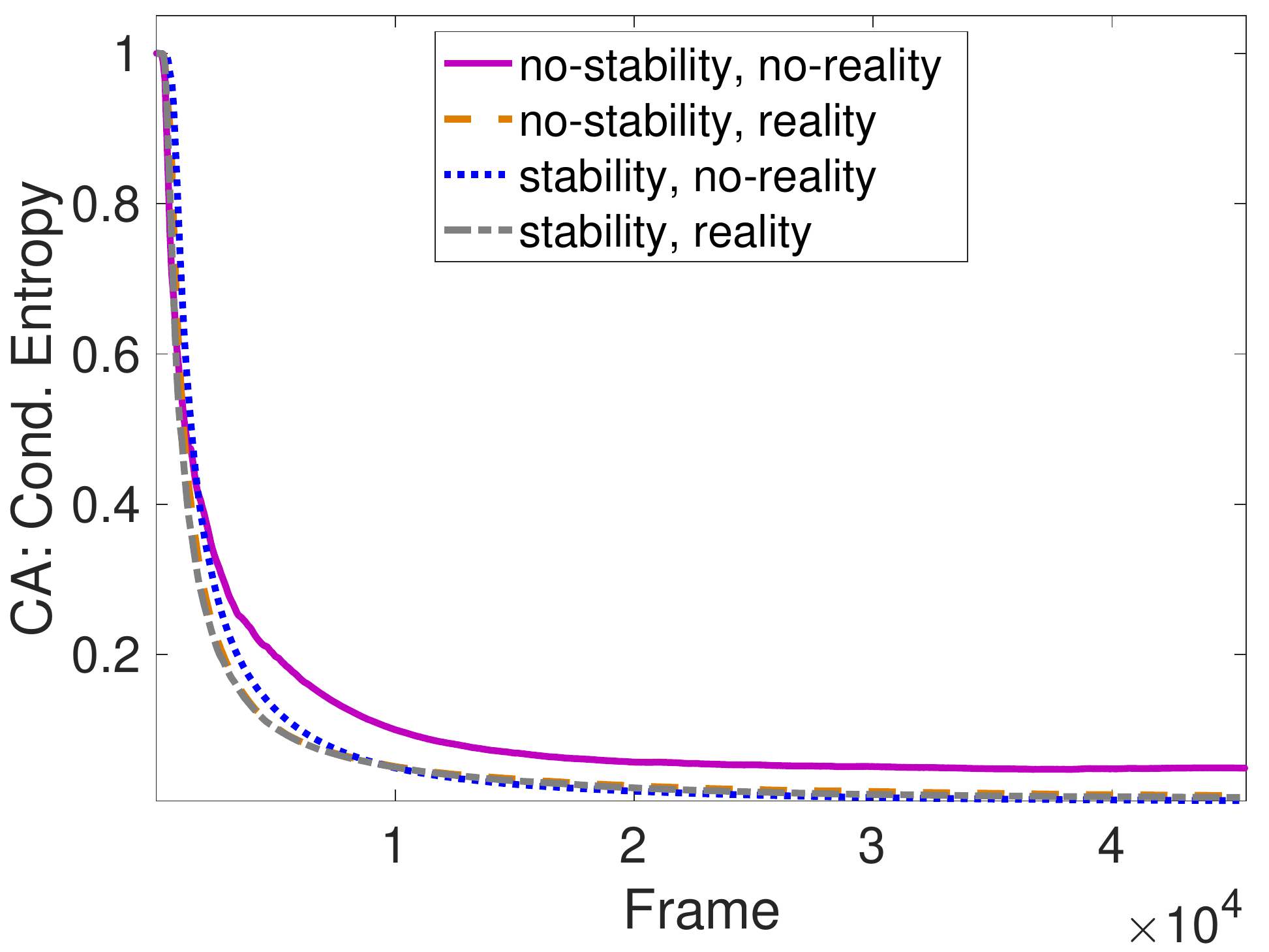}
\includegraphics[width=0.45\textwidth]{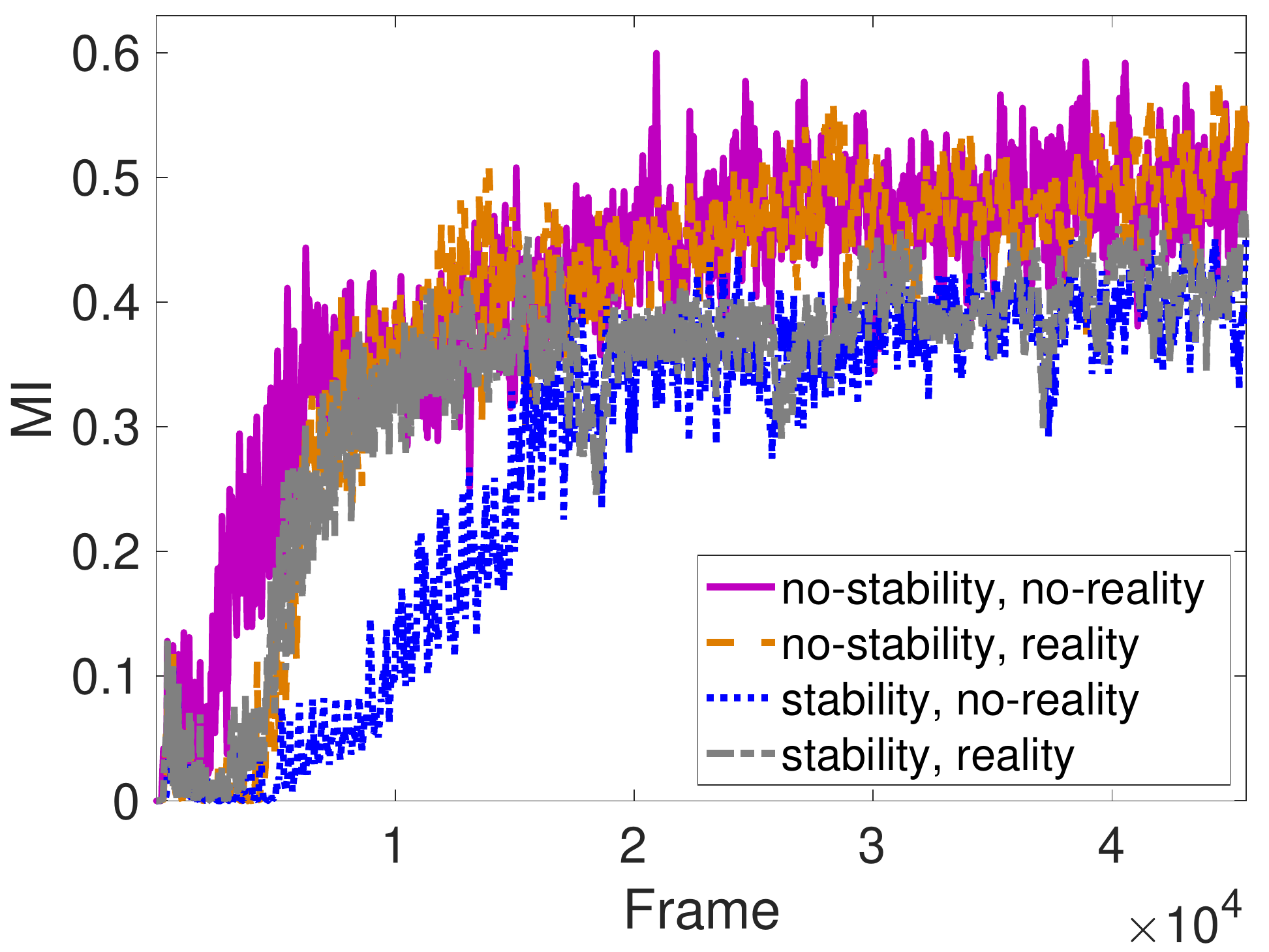}\\
\includegraphics[width=0.45\textwidth]{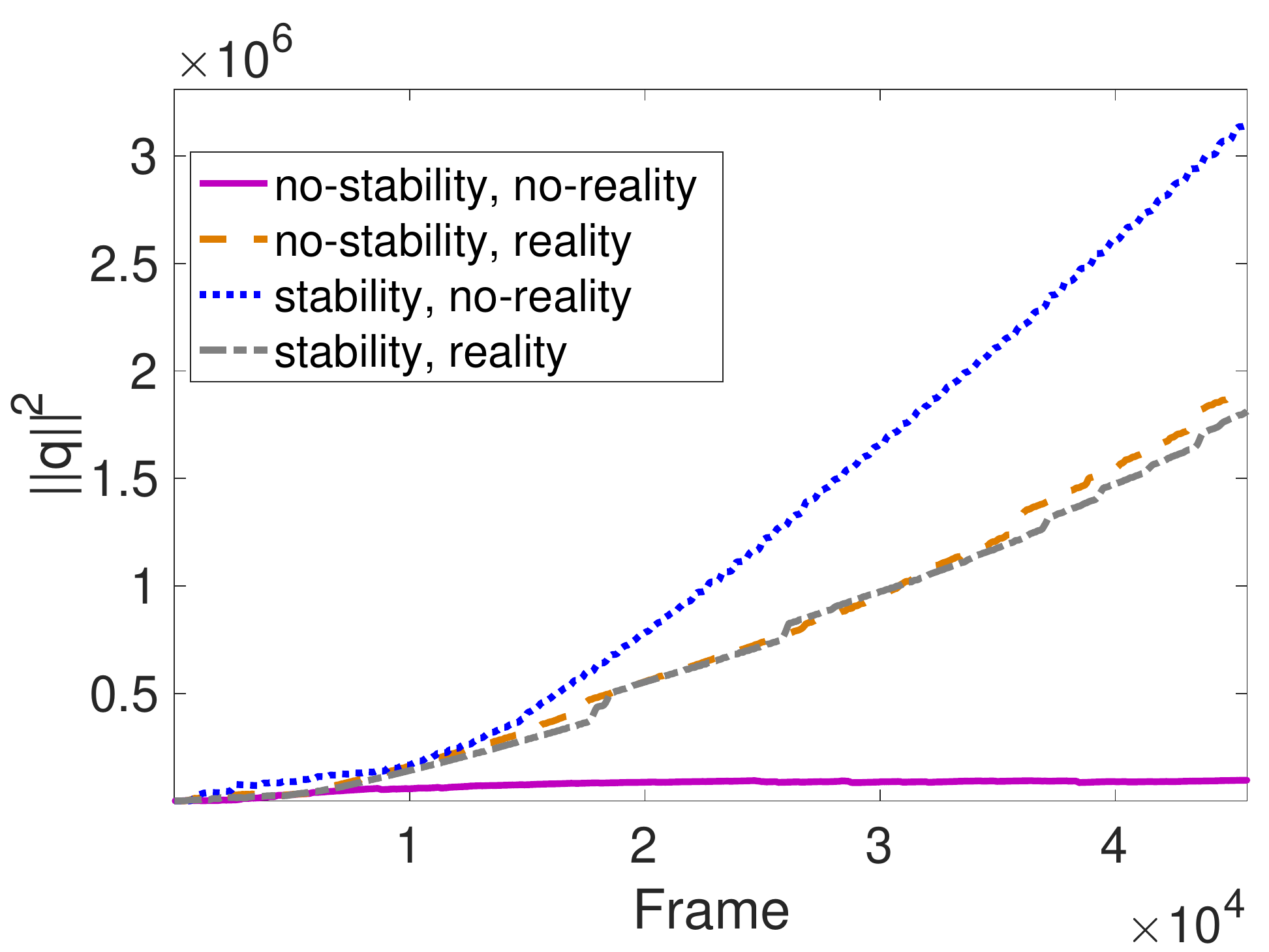}
\includegraphics[width=0.45\textwidth]{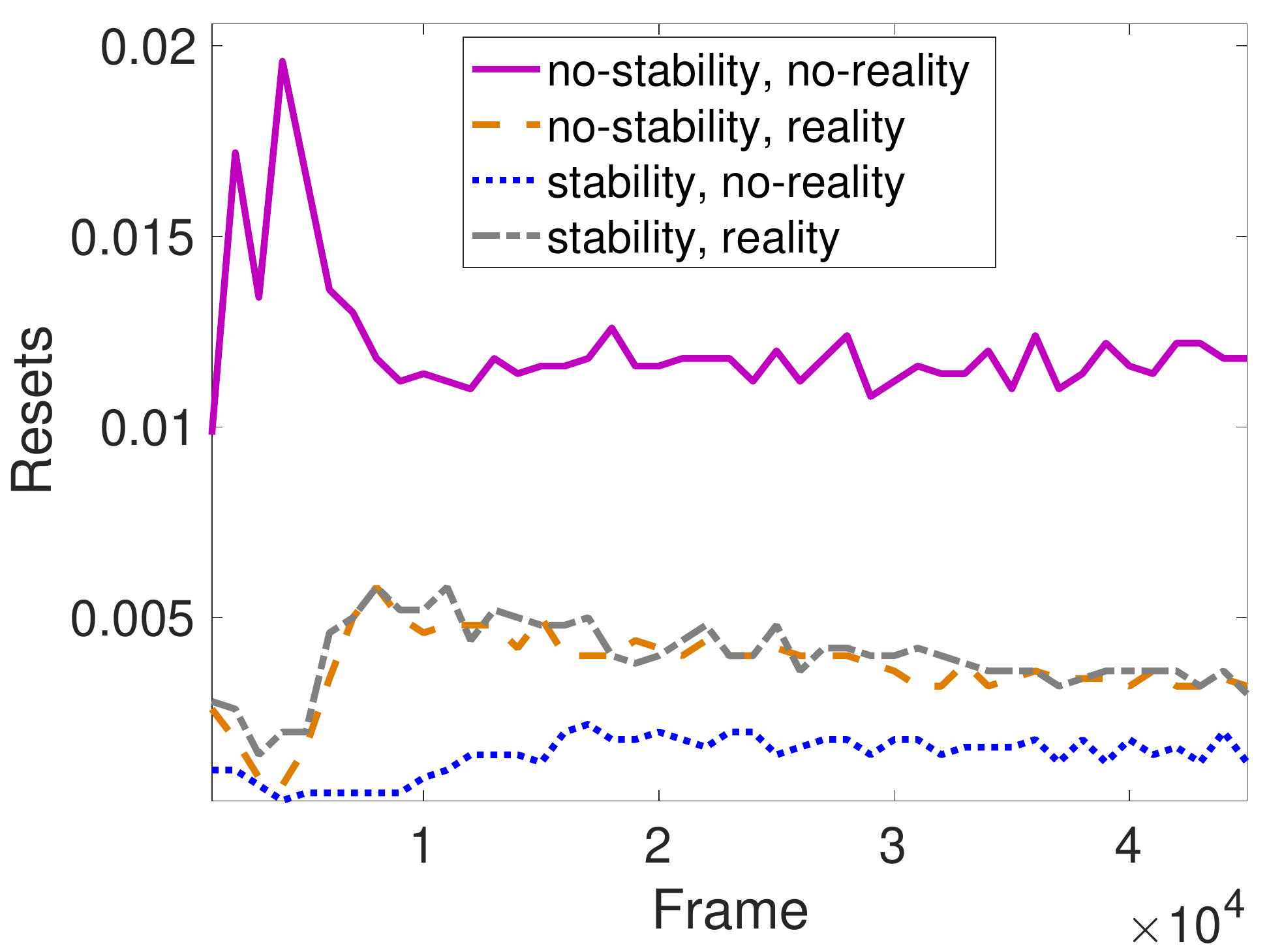}\\
\caption{Comparing 4 configurations of the parameters, characterized by different properties in terms of stability and reality of the roots of the characteristic polynomial. The input video is reproduced (in loop) for 45k frames (x-axis). From left-to-right, top-to-bottom we report the Cognitive Action (CA), the portion of the cognitive action that is about the Mutual Information (MI) (that we maximize), the portion that is about the Conditional Entropy, the MI per-frame, the norm of $q(t)$, and the fraction of ``reset'' operations performed every 1000 frames.}
\label{smallk} 
\end{figure}

Our experiments are designed (\textit{i}) to evaluate the dynamics of the cognitive action in function of different temporal regularities imposed to the model weights (parsimony), and then (\textit{ii}) to evaluate the effects of motion, that introduces a spatio-temporal regularization on single and multi-layer architectures.
When evaluating the temporal regularities, the cognitive action is composed by the information-based and parsimony terms only, and we experiment four instances of the set of parameters $\{\mu, \nu, \gamma, k\}$ of Eq. (\ref{cognitive-action-2}), leading to different dynamics. Each instance is characterized by the roots of the characteristic polynomial that lead to \textit{stable} or \textit{not-stable} configurations, and with only \textit{real} or also \textit{imaginary} parts, keeping the roots close to zero, and fulfilling the conditions of Proposition~\ref{prop:coef} when stability and reality are needed. These configurations are all based on values of $k \in [10^{-19}, 10^{-3}]$, while $\theta=10^{-4}$.
We performed experiments on the ``skater'' video clip, setting $n=5$  features, 
and filters of size $5\times5$. Results are reported in Fig.~\ref{smallk}. 
The plots indicate that there is an initial oscillation that is due to the effects of the blurring factor, that vanish 
after about 10k frames. The Mutual Information (MI) (${\cal I}$) portion of the cognitive action correctly increases over time,
and it is pushed toward larger values in the two extreme cases of ``no-stability, reality'' and ``no-stability, no-reality''. The latter shows more evident oscillations in the frame-by-frame MI value, due to the roots with imaginary part. In all the configurations the norm of $q$ increases over time (with different speeds), due to the small values of $k$, while the frequency of reset operations is larger in the ``no-stability, no-reality'' case, as expected. 

%
We evaluated the quality of the developed features by freezing the final $q$ of Fig.~\ref{smallk} and computing the MI index over a single repetition of the whole video clip, reporting the results in Tab.~\ref{nffsb} (a). This is the procedure we will follow in the rest of the paper when reporting numerical results in all the tables. We notice that, while in Fig.~\ref{smallk} we compute the MI on a frame-by-frame basis, here we compute it over the whole frames of the video at once, thus in a batch-mode setting. The result confirms that the two extreme configurations ``no-stability, reality'' and ``no-stability, no-reality'' show better results, on average. These performances are obtained thanks to the effect of the reset mechanism, that allows even such unstable configurations to develop good solutions. When the reset operations are disabled, we easily incurred into numerical errors due to strong oscillations while, for example, the ``stability" cases were less affected by this phenomenon.

\begin{figure}
\centering
\includegraphics[width=0.45\textwidth]{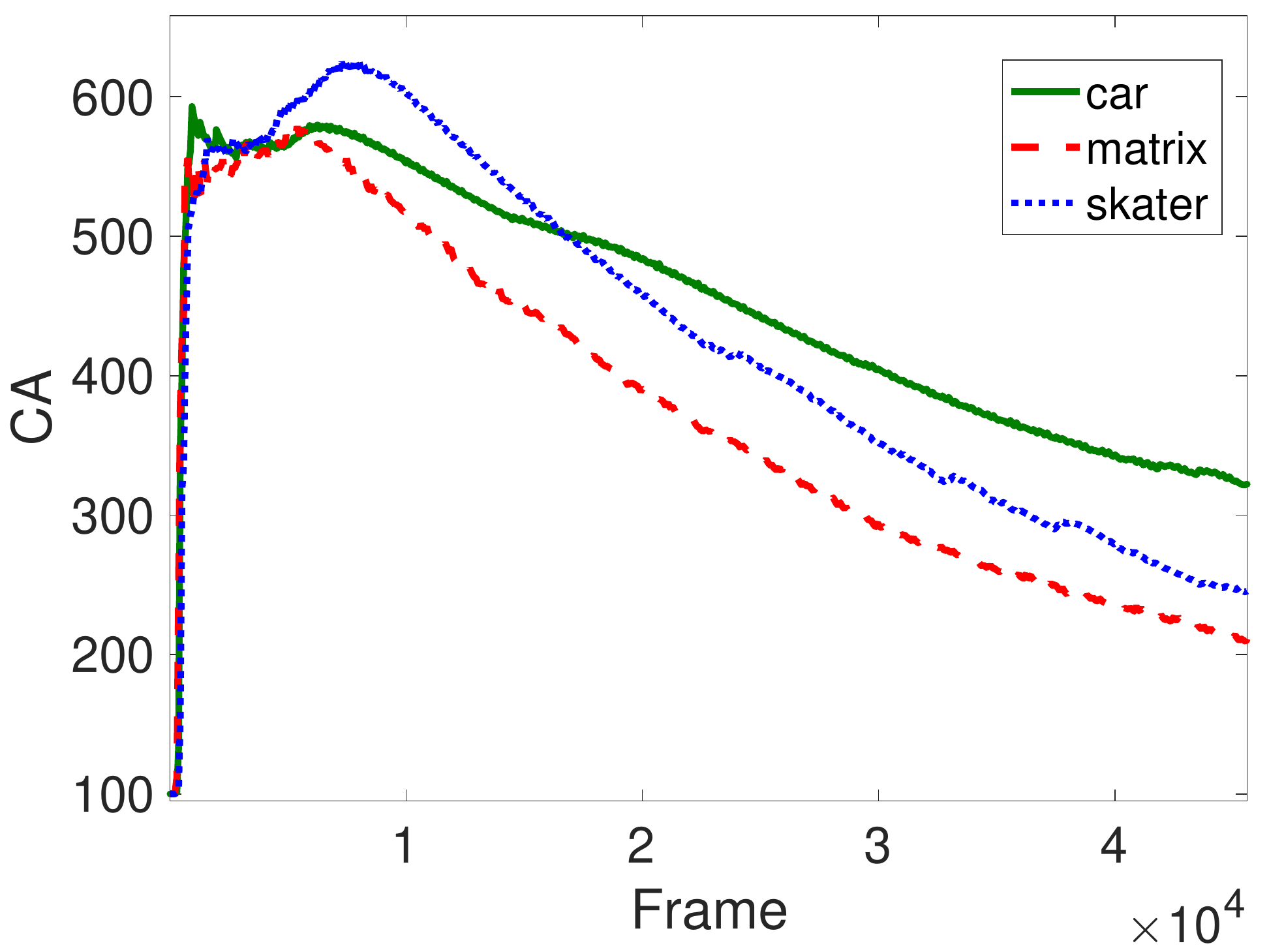}
\includegraphics[width=0.45\textwidth]{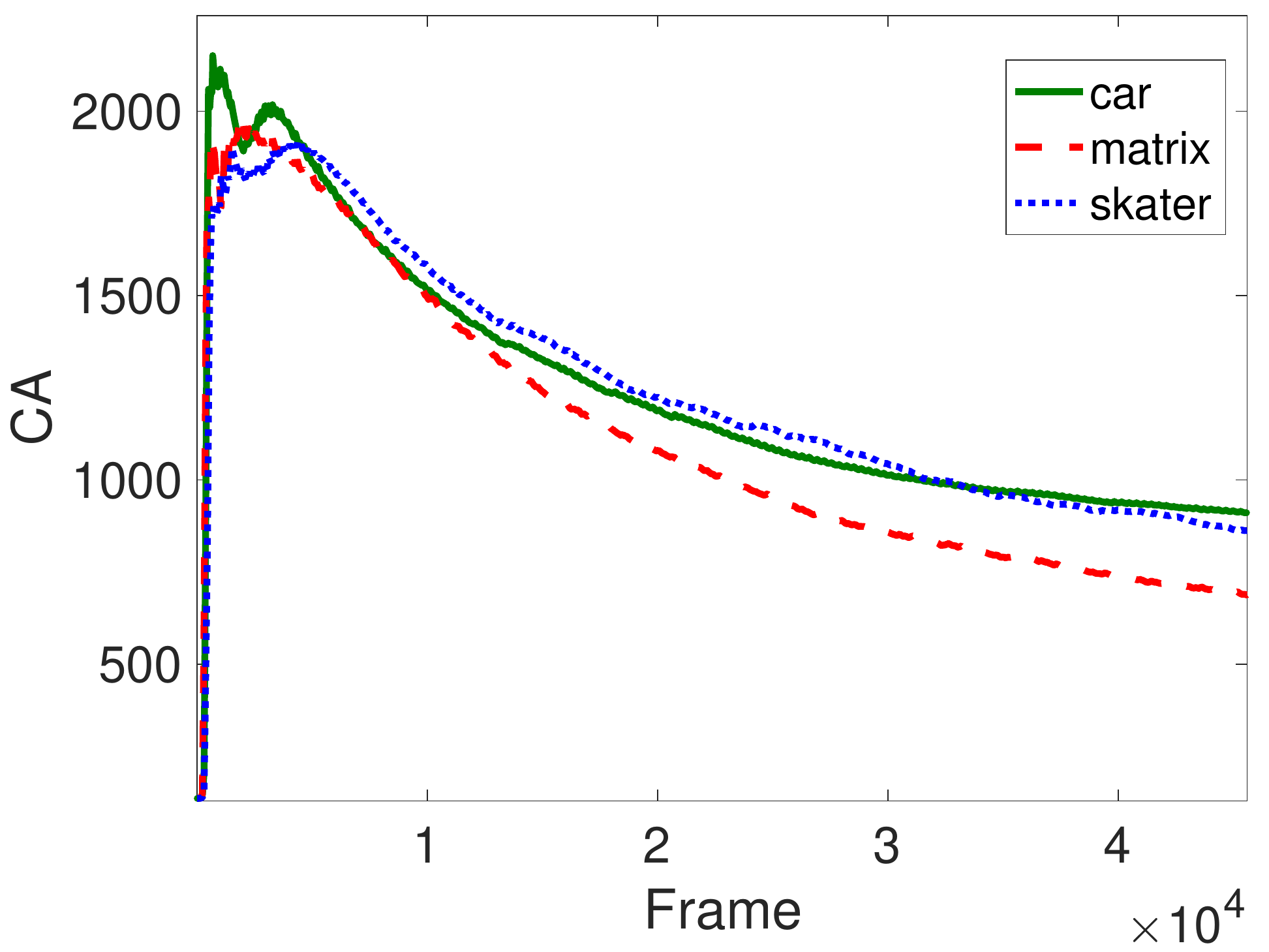}\\
\includegraphics[width=0.45\textwidth]{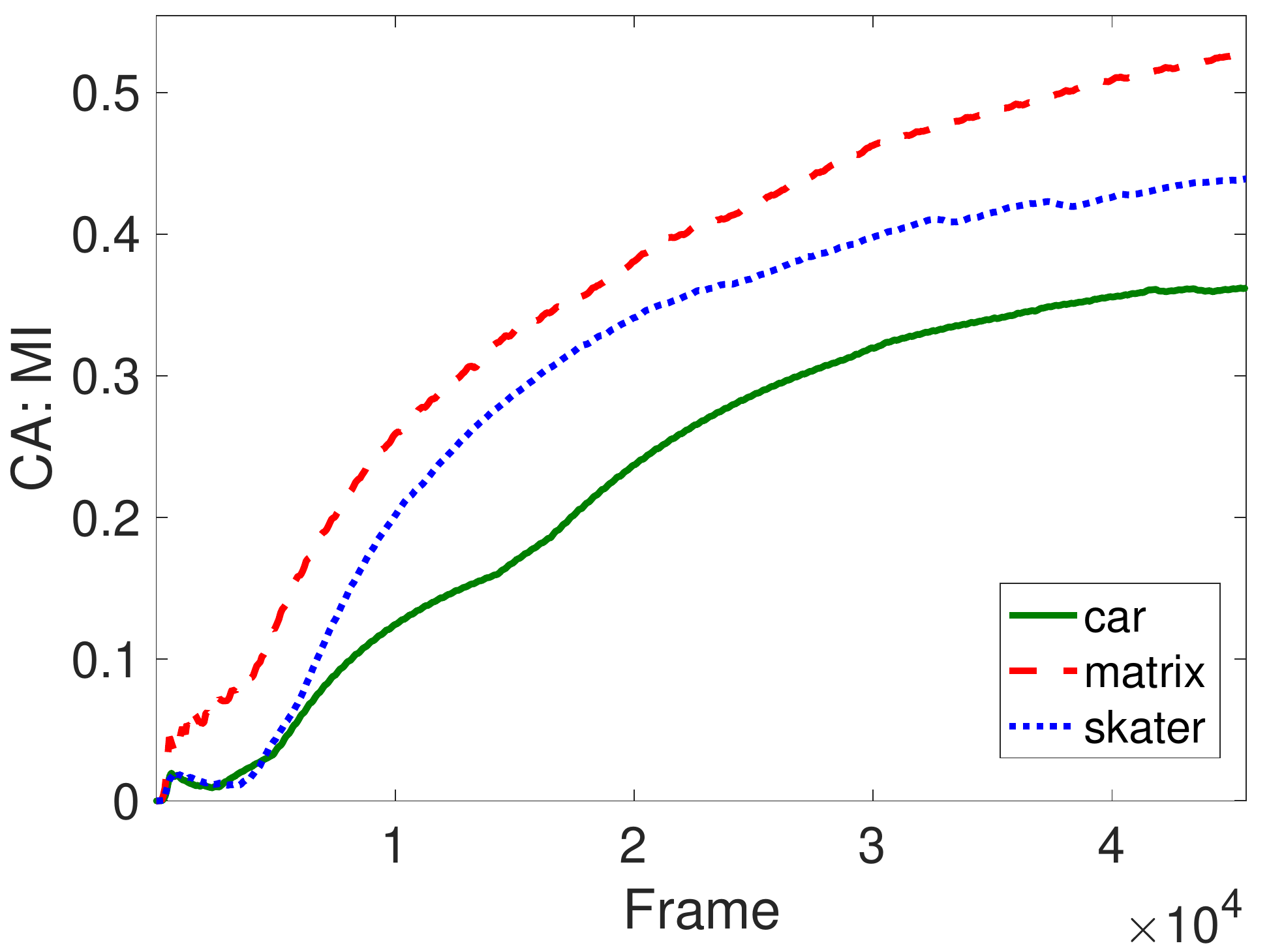}
\includegraphics[width=0.45\textwidth]{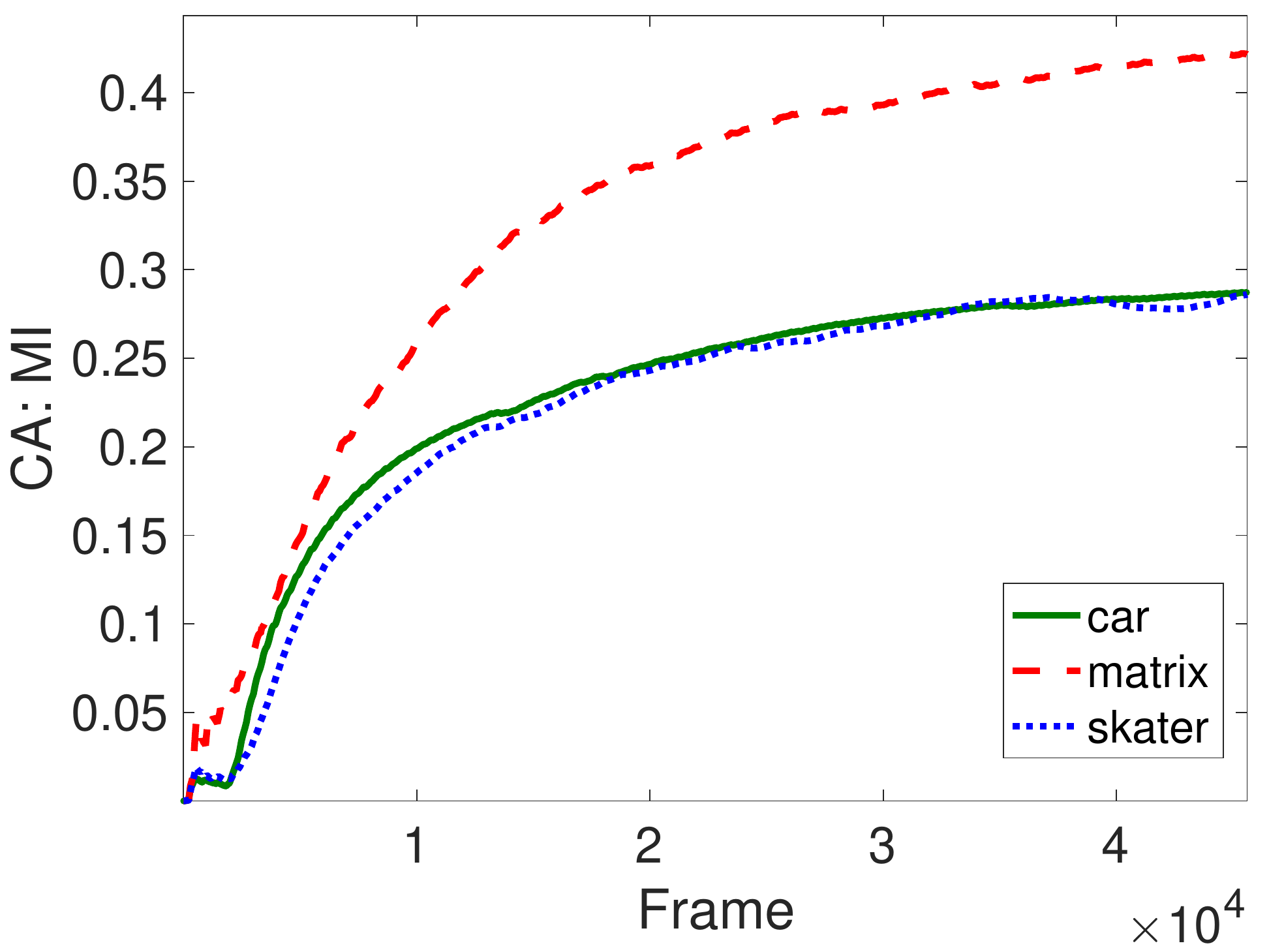}\\
\includegraphics[width=0.45\textwidth]{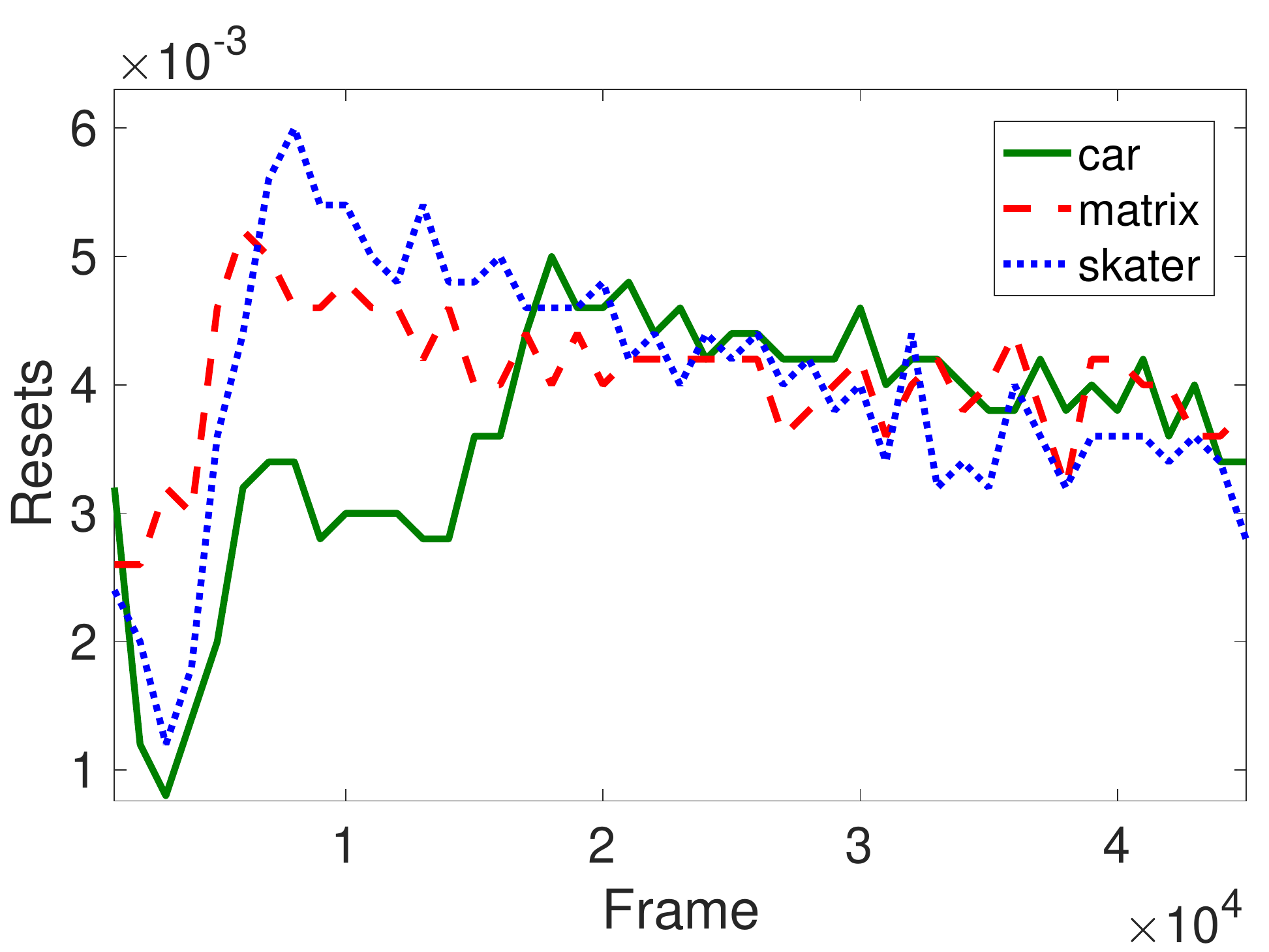}
\includegraphics[width=0.45\textwidth]{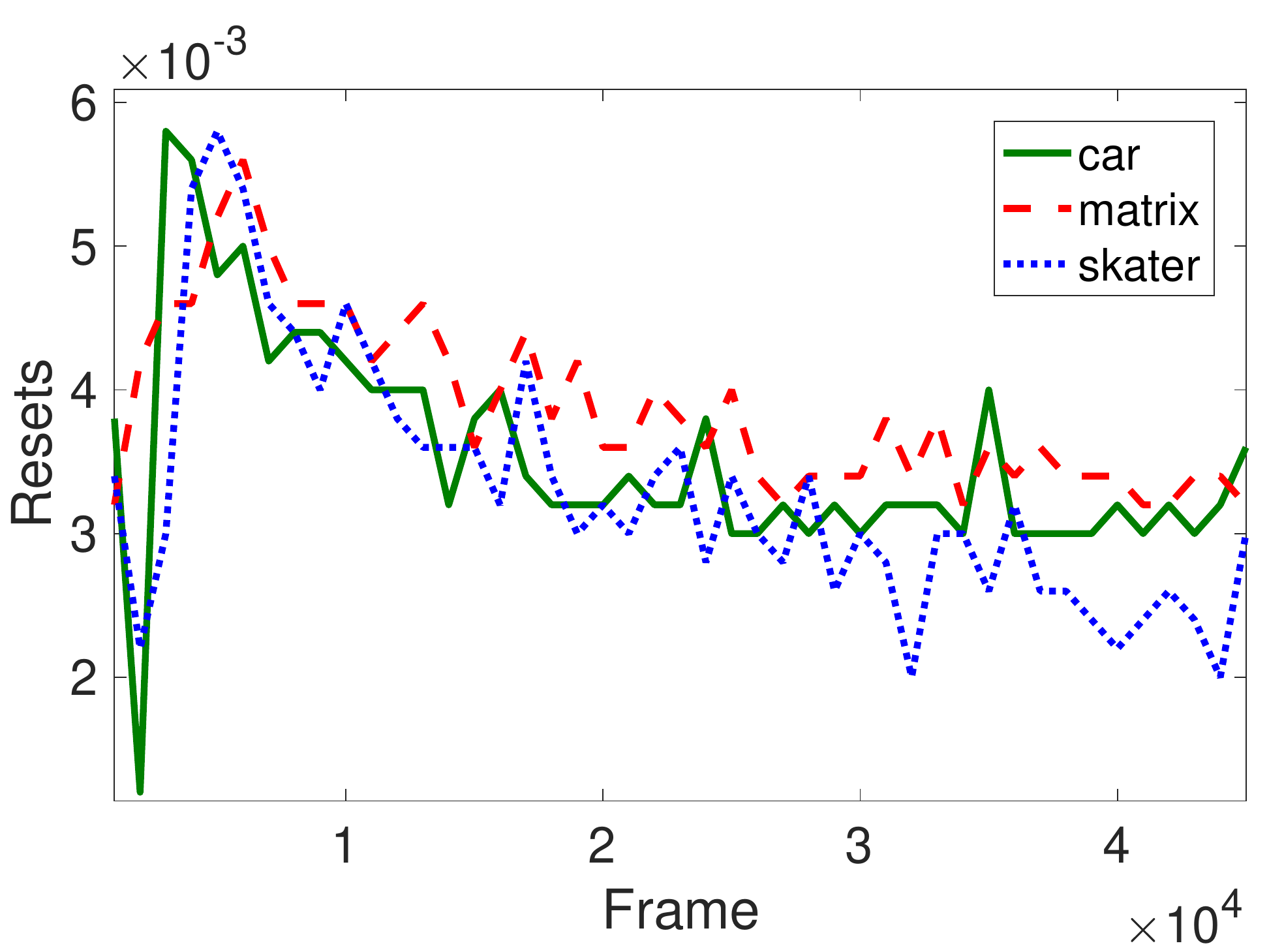}\\
\caption{Different number of features and filter sizes (1st column: $n=5, size=5\times5$; 2nd column: $n=11, size=11\times11$) in 3 videos. See Fig. \ref{smallk} for a description of the plots.}
\label{vids}
\end{figure}

We also compared the dynamics of the system on multiple video clips and using different filter sizes ($5\times5$ and $11\times11$) and number of features ($n=5$ and $n=11$) in Fig~\ref{vids}. We selected the ``stability, reality'' configuration of Fig.~\ref{smallk}, that fulfils the conditions of Proposition~\ref{prop:coef}. Changing the video clip does not change the considerations we did so far, while increasing the filter size and number of features can lead to smaller MI index values, mostly due to the need of a better balancing the two entropy terms to cope with the larger number of features. The MI of Tab.~\ref{nffsb} (b) confirms this point. Interestingly, the best results are obtained in the longer video clip (``The Matrix'') that requires less repetitions of the video, being closer to the real online setting. 

\begin{figure}
\centering
\includegraphics[width=0.45\textwidth]{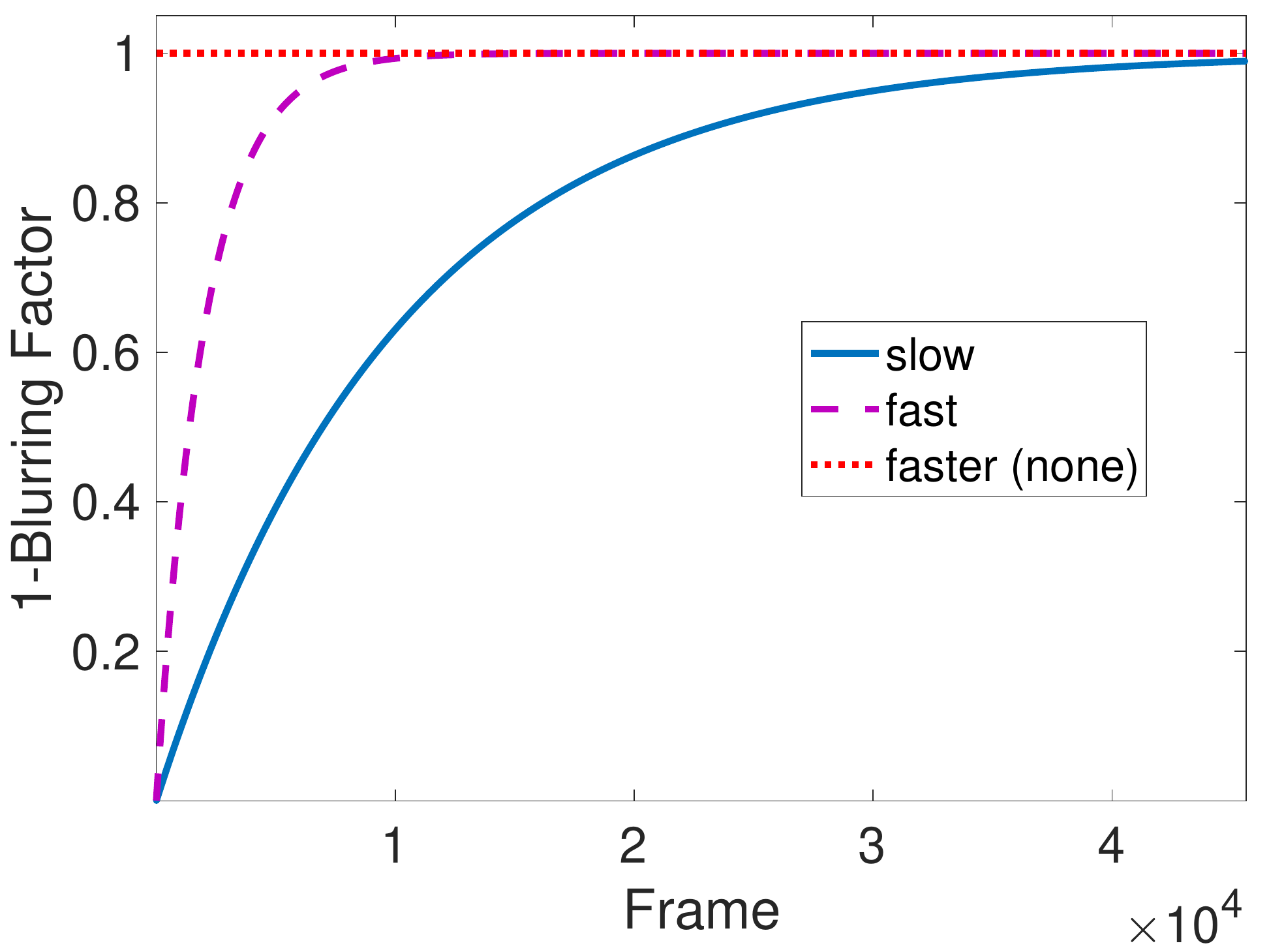}
\includegraphics[width=0.45\textwidth]{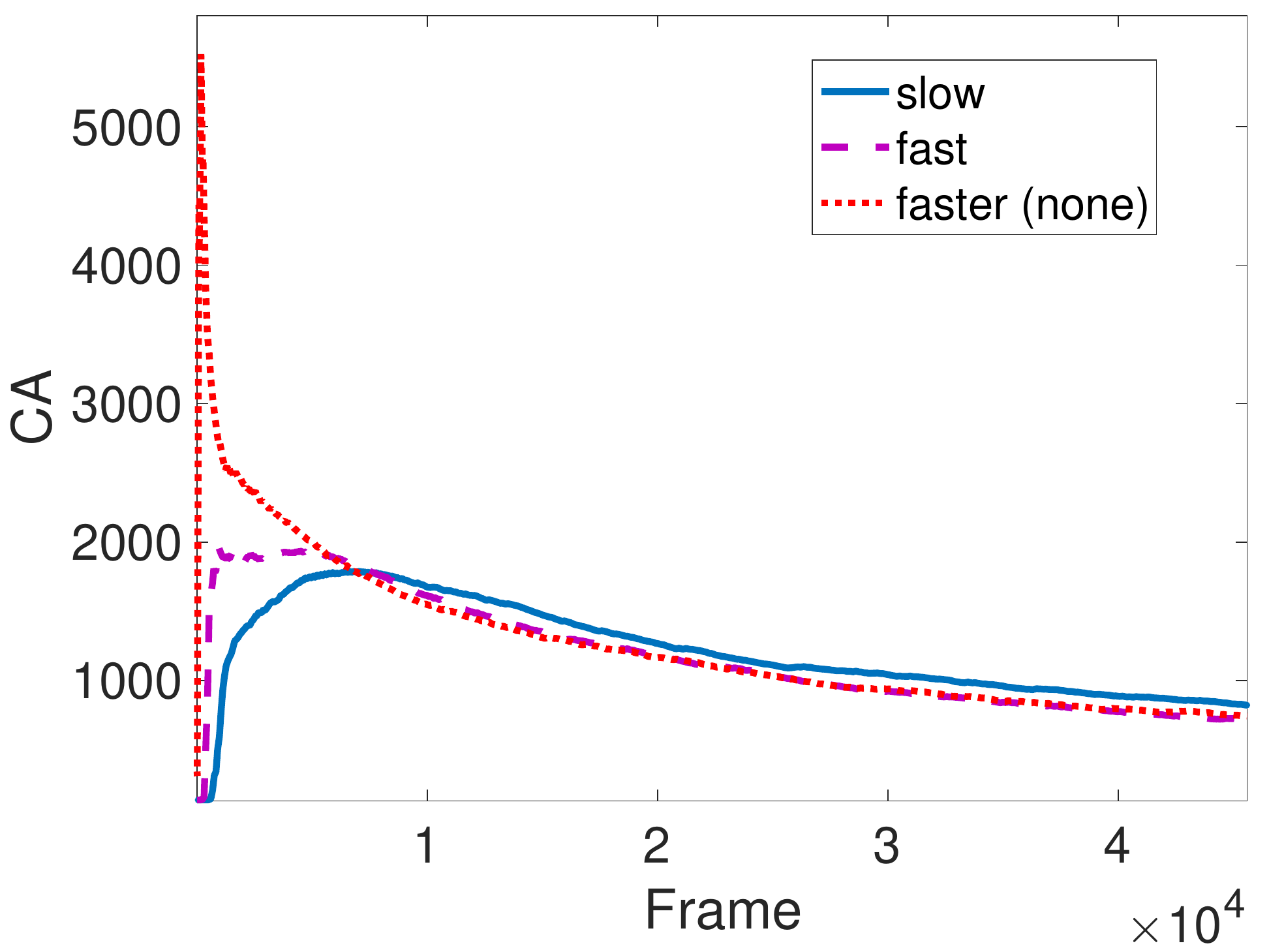}\\
\includegraphics[width=0.45\textwidth]{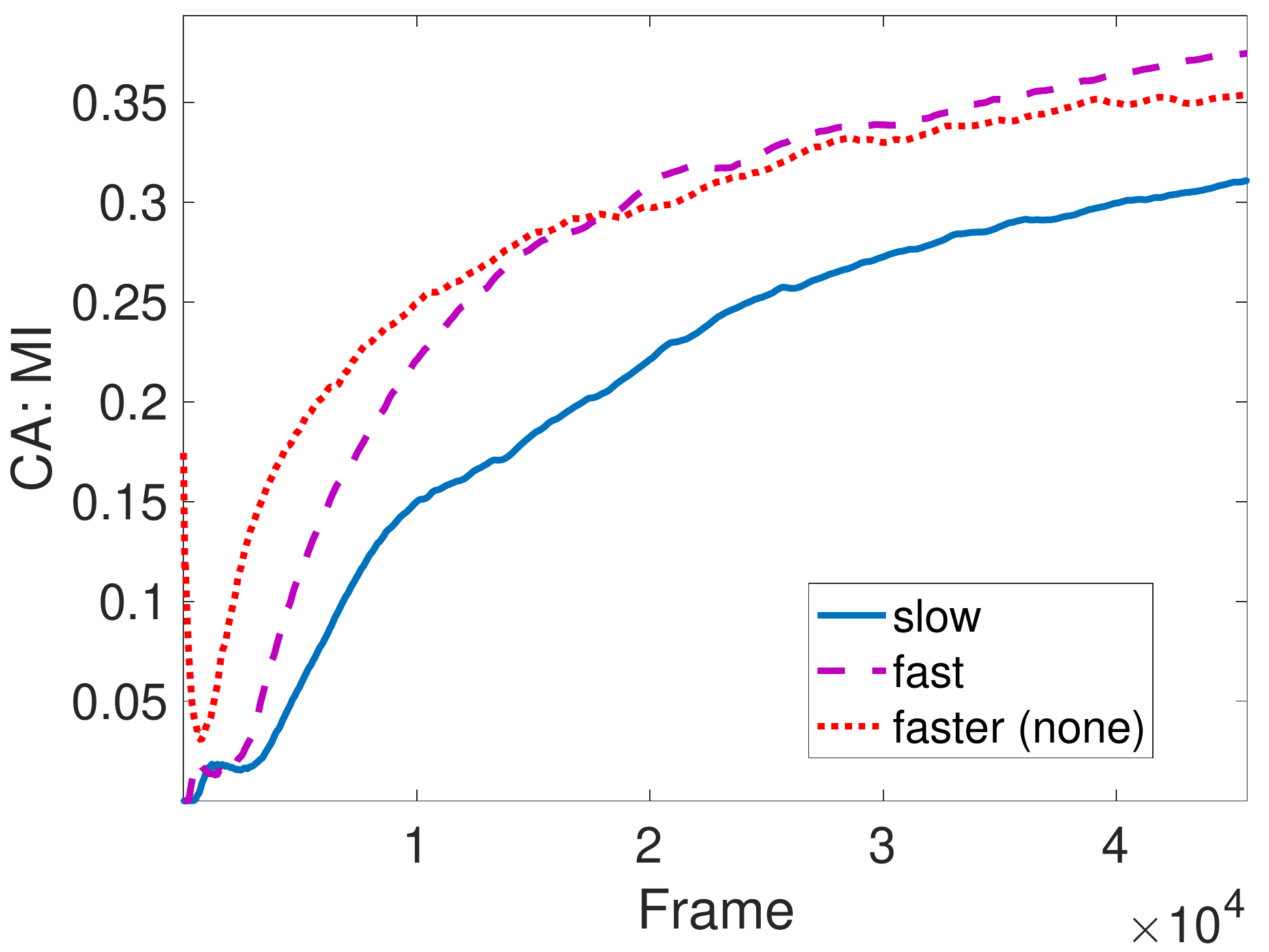}
\caption{Three different blurring plans: slow, fast, faster (i.e., no bluring). We consider $n=11$ and filters of size $11\times11$.}
\label{blurs}
\end{figure}

Fig.~\ref{blurs} and Tab. \ref{nffsb} (c) show the results we obtain when using different blurring plans (``skater'' clip), that is, different values of $\eta$ in Eq. (\ref{eq:blurplan}), that lead to the blurring factors reported in the first graph of Fig.~\ref{blurs}. These results suggest that a gradual introduction of the video signal helps the system to find better solutions than in the case in which no-plans are used, but also that a too-slow plan is not beneficial. The cognitive action has a big bump when no-plans are used, while this effect is more controlled and reduced in the case of both the slow and fast plans.


In order to study the effect of motion in multi-layer architectures (up to 3 layers), we still kept the most stable configuration (``stability, reality'', $5\times5$ filters, 5 features), and introduced the motion-related term in the cognitive action. Our multi-layer architecture is composed of a stack of computational models developed accordingly to (\ref{CognitiveActionEq}). A new layer $\ell$ is activated whenever layer $\ell-1$ has processed a large number of frames ($\approx 45k$), and the parameters of layer $\ell-1$ are not updated anymore. 
We initially considered the case in which all the layers $\ell=1,\ldots,3$ share the same value $\lambda_M$ that weighs the motion-based term. Tab. \ref{multilayer-1} shows the MI we get for different weighting schemes. Introducing motion helps in almost all the cases (for appropriate $\lambda_M$ - the smallest values of $\lambda_M$ are a good choice on average), and, as expected, a too strong enforcement of the motion-related term leads to degenerate solutions with small MI. We repeated these experiments also in a different setting. In detail, after having evaluated layer $\ell$ for all the values of $\lambda_M$, we selected the model with the largest MI and started evaluating layer $\ell+1$ on top of it. Tab. \ref{multilayer-2} reports the outcome of this experience. We clearly see that motion plays an important role in increasing the average MI. In the case of ``car'', we also obtained two (unexpected) positive results when strongly weighing $\lambda_M$. They are due to very frequent reset operations, that avoid the system to alter the filters when the too-strongly-enforced motion-based term yields very large derivatives. This is an interesting behaviour that, however, was not common in the other cases we reported. 

\begin{table}
\scriptsize
  \caption{MI on (a) the ``skater'' video, given the models of Fig. \ref{smallk} ($S$=stability, $R$=reality, $\bar X$=not X); (b) different videos, number of features, filter sizes (SR); (c) different blurring plans (SR).}
  \label{nffsb}
  \centering
  \hskip 4mm
  \begin{tabular}{rcrccrc}
    \toprule
    \multicolumn{2}{c}{(a)}&\multicolumn{3}{c}{(b)}&\multicolumn{2}{c}{(c)}\\
    \noalign{\smallskip}
    Config& \scriptsize(Skater)&Video& \scriptsize ($n=5$, $5\times 5$)& \scriptsize($n=11$, $11\times 11$)&
    Blurring& \scriptsize($n=10$, $5\times 5$)\\
    \cmidrule(r){1-2}\cmidrule(r){3-5}\cmidrule(r){6-7}
    $\bar S\bar R$ & $0.54\pm0.07$ & Car    &$0.38\pm 0.03$& $0.272\pm 0.003$ &Slow           & $0.35\pm 0.08$\\
    $\bar S R$ & $0.54\pm 0.08$ & Matrix &$0.60\pm 0.03$& $0.45\pm 0.02$   &Fast           & $0.39\pm 0.05$\\
    $S \bar R$ & $0.44\pm 0.11$ & Skater &$0.45\pm 0.13$& $0.35\pm 0.05$   &None & $0.34\pm 0.08$\\
    $SR$ & $0.45\pm 0.13$ \\
    \bottomrule
  \end{tabular}
\end{table}


\begin{table}
\scriptsize  
  \caption{MI in different videos, up to 3 layers ($\ell=1,2,3$), and for multiple weighting factors $\lambda_M$ of the motion-based term. All layers share the same $\lambda_M$.}
  \label{multilayer-1}
  \centering
    \hskip 2mm
  \newcolumntype{C}{>{\normalsize}c}
  \begin{tabu}{p{1mm}cccccccc}
    \toprule
    \rowfont{\small}
    &$ $& $\lambda_M=0$ &$10^{-8}$&$10^{-6}$
    &$10^{-4}$&$10^{-2}$&$1$&$10^2$\\
    \midrule\multirow{3}{*}{\rotatebox[origin=c]{90}{~Skater}}
    &$\ell=1$& $.61{\tiny \pm  .11}$ & $.54\pm .11$ &$.52\pm .07$&$.53\pm .08$
              &$\mathbf{.69}\pm .07$&$.53\pm 0$&$.01\pm 0$\\
    &$\ell=2$&$.53\pm .12$&$\mathbf{.62}\pm .15$&$.60\pm .11$&$.43\pm .06$&$.48\pm .06$
              &$.1 \pm .1$& $.03\pm .01$\\
    &$\ell=3$& $.56\pm .17$&$.58\pm .20$&$\mathbf{.62}\pm.10$& $.18\pm .16$
    &$.16\pm .17$& $.04\pm .02$&$.03\pm .02$\\
    \noalign{\smallskip}
    \midrule
    \multirow{3}{*}{\rotatebox[origin=c]{90}{~Car}}
    &$\ell=1$& $.49\pm.05$&$.44\pm.02$&$.46\pm.04$&$.47\pm.04$&$\mathbf{.66}\pm.10$&$.60\pm.02$
            &$.01\pm0$\\
    &$\ell=2$&$.25\pm .26$&$.54\pm.10$&$\mathbf{.65}\pm.08$&$.46\pm.03$&$.63\pm.11$&$.18\pm.32$
            &$.03\pm.01$\\
    &$\ell=3$&$.26\pm.34$&$.45\pm.22$&$\mathbf{.51}\pm.11$&$.38\pm.20$&$.24\pm.20$&$.09\pm.12$
    &$.04\pm.02$ \\
    \noalign{\smallskip}
    \midrule
    \multirow{3}{*}{\rotatebox[origin=c]{90}{~Matrix}}
    &$\ell=1$&$.66\pm.01$&$.66\pm.02$&$\mathbf{.67}\pm.01$&$.63\pm.05$&$.59\pm.03$&$.44\pm0$
             &$.23\pm.02$\\
    &$\ell=2$&$.55\pm.13$&$.56\pm.14$&$.43\pm0$&$.45\pm.04$&$\mathbf{.62}\pm.02$&$.35\pm.19$
             &$.13\pm.08$\\
    &$\ell=3$&$\mathbf{.64}\pm.03$&$.54\pm.11$&$.35\pm.07$&$.40\pm.01$&$.21\pm.07$&$.06\pm.03$
             &$.04\pm.02$\\
    \bottomrule
  \end{tabu}
\end{table}

\begin{table}
 \scriptsize  
  \caption{Same structure of Tab. \ref{multilayer-1}. Here the model with the best $\lambda_M$ is selected and used as basis to activate a new layer (layer $\ell=1$ is the same as Tab. \ref{multilayer-1}).}
  \label{multilayer-2}
  \centering
   \hskip 2mm
  \newcolumntype{C}{>{\normalsize}c}
  \begin{tabu}{p{1mm}cccccccc}
    \toprule
    \rowfont{\small}
    &$ $& $\lambda_M=0$ &$10^{-8}$&$10^{-6}$
    &$10^{-4}$&$10^{-2}$&$1$&$10^2$\\
    \midrule\multirow{2}{*}{\rotatebox[origin=c]{90}{~Skater}}
    &$\ell=2$&$.38\pm .34$&$\mathbf{.53}\pm .12$&$.50\pm .1$&$.47\pm .1$&$.41\pm .02$
              &$.33 \pm .17$& $.21\pm .2$\\
    &$\ell=3$& $.55\pm .12$&$\mathbf{.62}\pm .11$&$.55\pm.13$& $.42\pm .01$
    &$.36\pm .09$& $.2\pm .18$&$.39\pm .22$\\
    \noalign{\smallskip}
     \midrule\multirow{2}{*}{\rotatebox[origin=c]{90}{~Car}}
    &$\ell=2$& $.48\pm.1$&$.59\pm.17$&$.59\pm.18$&$.55\pm.12$&$.41\pm.01$&$.01\pm0$
            &$\mathbf{.64}\pm.01$\\
    &$\ell=3$&$.67\pm.01$&$.60\pm.12$&$\mathbf{.73}\pm.09$&$.36\pm.05$&$.33\pm.11$&$.27\pm.14$
    &$\mathbf{.73}\pm.01$ \\
    \noalign{\smallskip}
     \midrule\multirow{2}{*}{\rotatebox[origin=c]{90}{~Matrix}}
    &$\ell=2$&$.55\pm.13$&$.56\pm.14$&$.43\pm0$&$.45\pm.04$&$\mathbf{.62}\pm.02$&$.35\pm.19$
             &$.13\pm.08$\\
    &$\ell=3$&$.55\pm.12$&$.53\pm.12$&$\mathbf{.82}\pm.14$&$.35\pm.05$&$.35\pm.31$&$.02\pm.01$
             &$.01\pm0$\\
    \bottomrule
  \end{tabu}
\end{table}

\subsection{\rev{Semantic Labeling}}
\label{exp2}

\begin{figure}[ht]
\centering
\includegraphics[width=0.45\textwidth]{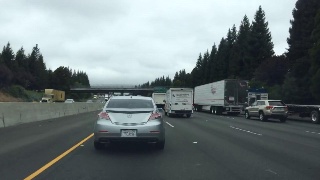}
\includegraphics[width=0.45\textwidth]{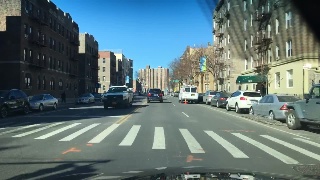}
\rev{\caption{Sample frames from the BDD100K dataset.}}
\label{bdd}
\end{figure}

\rev{
The task of Semantic Labeling consists in predicting the semantic class to which each pixel of the input frame belongs \cite{DBLP:journals/cviu/GoriLMM16}. This task is very challenging and it is usually faced in a fully-supervised context, in which the labeling of each pixel on a batch of images is provided to the system and a predictor is learned from such supervised signal \cite{yu2018bdd100k,yu2017dilated,ronneberger2015u}. 
While the nature of this task departs from the study reported in this paper -- in which we focus on the unsupervised development of pixel-level features -- we casted the Semantic Labeling task of the BDD100K data in a setting in which we can evaluate the behaviour of the learned features. BDD100K is a large-scale database of video sequences of driving, recorded using a camera mounted inside the driven car \cite{yu2018bdd100k}. Such data have been exploited in different competitions, also in the context of Semantic Labeling\footnote{\rev{See \url{https://bdd-data.berkeley.edu/} and links therein.}}. The HD video sequences are about different times in the day and weather conditions, and we provide a small sample in Fig. \ref{bdd}. Some frame were randomly taken from the whole video collection ($9,000$ frames), labeled with pixel-wise semantic information, and divided into training ($7,000$ frames), validation ($1,000$ frames), test set ($1,000$ frames). Labels belong to $19$ classes, such as sky, building, lamp, sign, etc. (see \cite{yu2018bdd100k} for all the details). Fully-supervised Dilated Residual Networks \cite{yu2017dilated} were evaluated on this data, showing how some of the target classes are very hard to recognize (see Table 4 in \cite{yu2018bdd100k}). 
} 

\rev{
We randomly selected 90 videos belonging to BDD100K, composed of $105,207$ frames, and that correspond to $\approx 1$ hour of driving scenes, 30 frames-per-second. These videos have no supervision attached. We rescaled each frame to $320 \times 240$ (RGB), keeping the original aspect ratio and frame rate, and we fed all the videos to the multi-layer architecture proposed in this paper, focussing on the same target configuration of Section \ref{exp1} (``stability, reality''). 
We considered three different neural architectures. The first two ones, named \textit{ca-1L} and \textit{ca-3L}, are a 1-layer and a 3-layer network, respectively, based on the proposed cognitive action with motion coherence. The weight $\lambda_M$ of the motion-related term was set to $10^{-6}$ in \textit{ca-1L} and to $10^{-6}, 10^{-4}, 10^{-2}$ in the 3-layers of \textit{ca-3L} (from lower to upper layers), in order to emphasize motion invariance as long as higher-level features are developed. We followed the same criterion of the experiments of Tab. \ref{multilayer-1} and Tab. \ref{multilayer-2}, setting $n=5$ and using filters of size $5\times5$. In \textit{ca-3L}, a new layer is activated when the one below has processed $30,000$ frames.
The third configuration, named \textit{ca-7L}, is composed of 7-layers, with weights $\lambda_M$ set to $0, 10^{-10}, 10^{-9}, 10^{-8}, 10^{-7}, 10^{-6}, 10^{-5}$, from the lower to the upper layer, respectively. In this case, we set $n=7$ and filters of size $7\times7$. This architecture is different from the ones experimented in Section \ref{exp1}. We selected more layers and larger receptive inputs in order to virtually cover a larger portion of the input frames around each pixel when extracting features. The last layer in \textit{ca-7L} covers $43 \times 43$ pixels of the input frame ($7\times 7$ pixels in the first layer, while each following layers extends each dimension by filter-size-minus-one pixels ($6$)), that is 13\% of the frame width. A new layer is activated whenever the one below has processed $15,000$ frames. 
}

\rev{
The unsupervised learning stage develops convolutional feature extractors, and each pixel is represented by the concatenation of the feature vectors that are yielded by each convolutional layer. Then, such pixel representation is provided to a Multi-Layer Perceptron (MLP) with 1 hidden layer ($100$ hidden neurons with tanh activation, softmax activation in the output layer), that was trained on the BDD100K Semantic Labeling training set. 
Since we are not using any supervisions in the feature extraction stage, the convolutional filters do not learn discriminative information from the supervision signal, differently from the models tested in \cite{yu2018bdd100k}.
For this reason, we focussed on the $5$ classes (road, building, vegetation, sky, car) that, accordingly to \cite{yu2018bdd100k}, lead to large recognition accuracies in the fully-supervised case of Dilated Residual Networks, avoiding those classes that are hardly classified by full-supervised convolutional models. We exploited the same measure of accuracy (Mean Intersection over Union (Mean IoU)) of related work \cite{yu2017dilated,yu2018bdd100k}. Our implementation is based on the code that implements Dilated Residual Networks for Semantic Labeling\footnote{\rev{It can be dowloaded from \url{https://github.com/fyu/drn}.}}, on which we plugged our feature extractor and the MLP. We considered both the cases in which only the features extracted by the convolutional layers are fed to the MLP classifier, and when they are augmented by the RGB information of the pixel. In the latter case, models are named \textit{ca-rgb-1L}, \textit{ca-rgb-3L}, \textit{ca-rgb-7L}, respectively.
}

\rev{
We compared the cognitive action-based models with stacked Sparse Convolutional Autoencoders \cite{luo2017convolutional,masci2011stacked}, indicated with \textit{autoenc}. Autoencoders learn to rebuild the receptive input (the area covered by each convolutional filter) after having projected it toward a lower dimensional sparse representation. We set the representation to have size $n$, as in the case of the proposed model, and we used the same filter size and number of layers of the \textit{ca-$*$L} models. Sparsity is enforced by means of $L_1$-norm regularization, weighted by a scalar coefficient that, for each layer, was tuned in $\{10^{-4}, 10^{-6}, 10^{-7}, 10^{-8} \}$. Autoencoders are trained with stochastic gradient, updating the model after each processed frame. Learning rate was set to $0.033$, that is $1$ over frame-rate, as the step-size of the Euler method in the case of the cognitive action-based models. Also in the case of autoencoders we considered the augmentation of the feature representation with RGB information -- we follow the same naming convention of the other models. Finally, we compared with the setting in which only the RGB information is fed to the MLP (\textit{baseline-rgb}).
}

\rev{
Tab. \ref{sl} reports the results of our experiments, averaged over 5 runs. All the compared models were initialized with the same filter values. Standard deviations were very small (below $1$), and we do not report them for better readability. The upper portion of the table (above the horizontal line) shows that all the models that exploit convolutional features overcome the baseline result. At a first glance, this might look obvious, since the convolutional models develop a representation of each pixel that includes information about the area around the pixel itself. However, the learned features have no information at all on the considered task (being them developed in an unsupervised manner), and some of them could also be strongly unrelated with what is needed to correctly classify data belonging to the considered classes, making the result more appealing.
Tab. \ref{sl} confirms that adding more layers improve the quality of the representations, and it shows that cognitive action-based models with motion coherence lead to better results than sparse autoencoders, when both are paired with RGB information. This result is mostly due to the positive effects of the kinetic and motion-related terms, that introduce a form of regularization over time. Interestingly, the class ``car'' is better recognized in our model than in the case of autoencoders, being it a class of explicitly  moving objects (of course also the camera motion implicitly introduces motion information). 
Focussing on the bottom portion of Tab. \ref{sl} (below the horizontal line), the convolutional models yield results below the baseline. This indicates that those features that are learned by an unsupervised process do not significantly encode color information about the target pixel, but they are more likely oriented toward shape-related description of the area around such pixel. Autoencoders show, on average, a better performance than the models proposed in this paper. The class ``car''  seems better classified by autoencoders, that is in contrast with our previous comment. However, we noticed that such class is frequently predicted, leading to very small IoU scores that are harder to compare. Adding the RGB information makes the pixel representations less ambiguous and more tightly activated on the car-related areas.
}

\begin{table}[ht]
\scriptsize
  \rev{\caption{Semantic Labeling, BDD100K dataset, Intersection over Union (IoU) score over $5$ classes, comparing different models: those that are based on cognitive action with motion coherence (``cal'') and those that exploit sparse convolutional autoencoders trained with stochastic gradient (``autoenc''), both of them with different numbers of layers (``$\ell$L'' suffix, being $\ell$ the number of layers). We report the features per layer (``$n$'') and the size of the receptive fields in brackets. We distinguish between models that augment the pixel-wise representations explicitly including the RGB information (``rgb'' string included in their name) and those that only exploit features from the unsupervised learning stage (the horizontal line separates these two classes of models). The model named ``baseline'-rgb' only uses RGB information.}
  \vskip 2mm
  \label{sl}
\begin{center}
  \begin{tabu}{l|ccccc|c}
    \toprule
    \rowfont{\scriptsize}
{Model} &  \textit{Road} & \textit{Building} & \textit{Vegetation} & \textit{Sky} & \textit{Car} & {Mean IoU}\\ 
\midrule 
cal-rgb-7L ($n=7$, $7\times7$, $\ell=7$)  & 43.92 & 23.05 & 29.83 & 79.71 & 15.82 & 38.47\\  
cal-rgb-3L ($n=5$, $5\times5$, $\ell=3$) & 43.17 & 23.02 & 29.35 & 79.36 & 14.62 & 37.90\\  
cal-rgb-1L ($n=5$, $5\times5$, $\ell=1$)& 42.33 & 21.59 & 28.98 & 78.40 & 14.14 & 37.09\\  
autoenc-rgb-7L ($n=7$, $7\times7$, $\ell=7$)& 42.00 & 20.36 & 29.25 & 79.33 & 12.03 & 36.59\\
autoenc-rgb-3L ($n=5$, $5\times5$, $\ell=3$)& 40.74 & 20.43 & 28.47 & 77.68 & 13.94 & 36.25\\ 
autoenc-rgb-1L ($n=5$, $5\times5$, $\ell=1$) & 39.26 & 20.87 & 28.06 & 77.66 & 12.20 & 35.61\\ 
baseline-rgb & 38.91 & 19.65 & 27.98 & 76.69 & 10.99 & 34.85\\
\midrule
autoenc-7L ($n=7$, $7\times7$, $\ell=7$) & 30.87 & 15.33 & 18.61 & 68.48 & 2.95 & 27.25\\
cal-7L ($n=7$, $7\times7$, $\ell=7$) & 32.51 & 19.05 & 8.39 & 70.12 & 1.03 & 26.22\\  
autoenc-3L  ($n=5$, $5\times5$, $\ell=3$) & 29.68 & 10.62 & 18.62 & 67.25 & 2.81 & 25.80\\ 
cal-3L ($n=5$, $5\times5$, $\ell=3$)& 30.74 & 18.95 & 6.07 & 67.64 & 1.71 & 25.02\\  
cal-1L ($n=5$, $5\times5$, $\ell=1$) & 29.71 & 10.34 & 9.31 & 61.92 & 0.12 & 22.28\\  
autoenc-1L  ($n=5$, $5\times5$, $\ell=1$) & 27.27 & 12.79 & 8.21 & 52.40 & 0.22 &  20.18 \\
 \bottomrule
\end{tabu}
\end{center}}
\end{table}

%% file: conc.tex
This paper proposes a learning theory that formalizes the problem of learning visual
features from streams of visual data. The theory, which is based on the  ergodic assumption of
the video signal, provides a reformulation of classic unsupervised learning schemes based
on mutual information. In particular, the convolutional filters are derived within the
framework of the principle of least action that nicely parallels the methods
derived from statistical learning. A distinctive contribution of this paper is the  
idea of enforcing motion coherence joined with information-based indices to learn convolutional architectures by an on-line computational scheme.
In addition to the promising experimental results, this paper gives first insights to
address ten questions for a theory of vision that are stated at the very beginning. 
We are currently working towards the extension to deep networks of the 
same theory with the purpose of constructing a visual agent that interacts 
with the environment. The underlying assumption is that the injection of motion 
invariance will play a crucial role for a dramatic reduction of supervisions. 
Unlike what is reported in this paper, more than invariance with respect 
to object movement, we have been experimenting the invariance with respect to
the focus of attention on the basis of a model proposed in~\cite{zanca2019gravitational}. Interestingly, since our focus of attention model mostly 
tracks moving objects there are clear similarities with the model reported in this 
paper. However, the micro-saccades that characterizes the focus of attention 
in presence of still images also lead to enforcing motion invariance, thus 
enriching dramatically the amount of information that drives the learning process. 



%% file: motion.tex
\label{motion-invariance-var-append}
The motion invariance term can be written as
\[\begin{split}\omega(\varphi)=&
\frac{1}{2}\int dzd\tau\, f(z,\tau)\Bigl(\int d\xi\, \textrm{C}_m(z-\xi)
\partial_\tau \varphi_{km}(\xi,\tau)\\
&+(\partial_\tau \textrm{C}_m(z-\xi,\tau)+
v_\alpha \partial_\alpha \textrm{C}_m(z-\xi,\tau))\varphi_{km}(\xi,\tau)\Bigr)^2,
\end{split}\]
or equivalently
\[\begin{split}\omega(\varphi)=&
\frac{1}{2}\int d\tau d\xi d\zeta\,
\partial_\tau \varphi_{km}(\xi,\tau) \textrm{W}_{ml}(\xi,\zeta,\tau)
\partial_\tau \varphi_{kl}(\zeta,\tau)\\
&+2\varphi_{km}(\xi,\tau)\textrm{Y}_{ml}(\xi,\zeta,\tau)\partial_\tau \varphi_{kl}
(\zeta,\tau)
+\varphi_{km}(\xi,\tau) \textrm{H}_{ml}(\xi,\zeta,\tau)\varphi_{kl}(\zeta,\tau),
\end{split}\]
where $\textrm{W}_{ml}(\xi,\zeta,\tau)=\int dz\, f(z,\tau) \textrm{C}_m(z-\xi,\tau)
\textrm{C}_l(z-\zeta,\tau)$, while  $\textrm{Y}_{ml}(\xi,\zeta,\tau)=
\int dz\, f(z,\tau)[\partial_\tau \textrm{C}_m(z-\xi,\tau)+
v_\alpha \partial_\alpha \textrm{C}_m(z-\xi,\tau)] \textrm{C}_l(z-\zeta,\tau)$ and
$H_{ml}(\xi,\zeta,\tau)=
\int dz\, f(z,\tau)[\partial_\tau \textrm{C}_m(z-\xi,\tau)+
v_\alpha \partial_\alpha \textrm{C}_m(z-\xi,\tau)][\partial_\tau \textrm{C}_l(z-\zeta,\tau)+
v_\beta \partial_\beta C_l(z-\zeta,\tau)]$.
Since we can always assume that at $t=0$ and $t=T$ that the video $\textrm{C}$ with its
derivative is identically zero, we automatically have
$\textrm{W}_{ml}(\xi,x,0)\equiv \textrm{W}_{ml}(\xi,x,T)\equiv \textrm{Y}_{ml}(\xi,x,0)\equiv
\textrm{Y}_{ml}(\xi,x,T)\equiv \textrm{H}_{ml}(\xi,x,0)\equiv \textrm{H}_{ml}(\xi,x,T)\equiv 0$.
This property saves us from having boundary terms coming from
the integration by parts that we need to perform when we compute the
variation of this term.

We can now compute the first variation of the functional with respect to
$\varphi_{ij}(x,t)$:
\[\begin{split}
\frac{\delta\omega(\varphi)}{\delta\varphi_{ij}(x,t)}=
\int d\xi[&-\textrm{W}_{lj}(\xi,x,t)\partial^2_t \varphi_{il}(\xi,t)\\
&+(\textrm{Y}_{jl}(x,\xi,t)-\textrm{Y}_{lj}(\xi,x,t)-\partial_t \textrm{W}_{lj}(\xi,x,t))
\partial_t\varphi_{il}(\xi,t)\\
&+(\textrm{H}_{lj}(\xi,x,t)-\partial_t \textrm{Y}_{lj}(\xi,x,t))\varphi_{il}(\xi,t)].
\end{split}\]
By the following identification
\[\Xi_{jk}(x,\xi,t)=-\textrm{W}_{jk}(\xi,x,t),\quad
\Upsilon_{jk}(x,\xi,t)=\textrm{H}_{kj}(\xi,x,t)-\partial_t \textrm{Y}_{kj}(\xi,x,t),
\]
\[\Pi_{jk}(x,\xi,t)=
\textrm{Y}_{jk}(x,\xi,t)-\textrm{Y}_{kj}(\xi,x,t)-\partial_t \textrm{W}_{kj}(\xi,x,t),\]
the above expression is indeed the result stated in Eq.~(\ref{MotionELT}).


%% file: gauss.tex
\label{Gauss-appendix}
First of all define $L_\sigma^m:=\sum_{n=0}^m
(-1)^n(\sigma^{2n}/ 2^n n!) {d^{2n}/ dx^{2n}}$ and consider the induced function
$\rho_\sigma(x):= L_\sigma^m G_\sigma(x)$.
Now notice that as $\sigma\to 0$ $L\to 1$ and $G_\sigma\to \delta$ so
that we automatically have $LG=\delta$. It is also immediate to see that
$\int \rho_\sigma(x)\, dx=1$.
Let us now consider for any $\delta>0$

\def\|#1|{\includegraphics[width=0.45\hsize]{#1}}

 \begin{figure}
		\hbox to\hsize{\hfil\|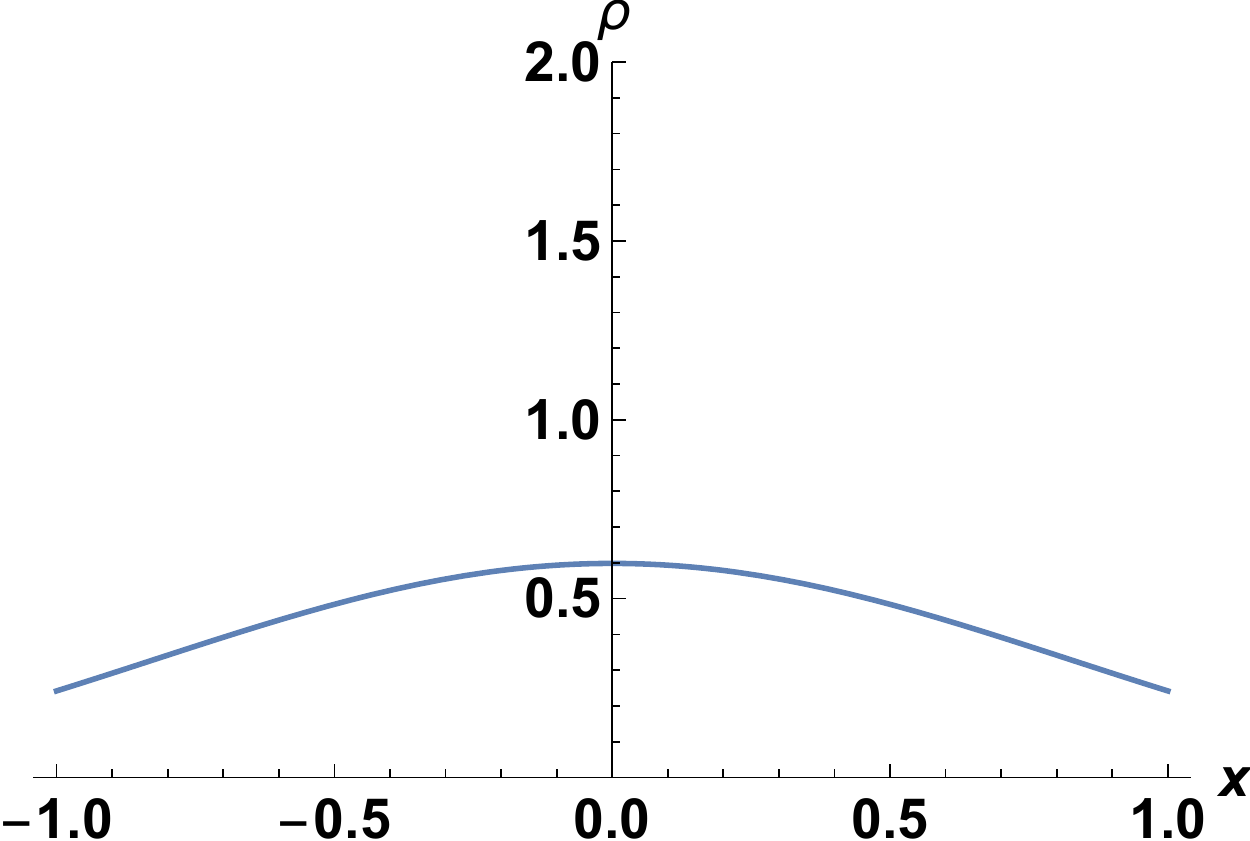|\hfil\|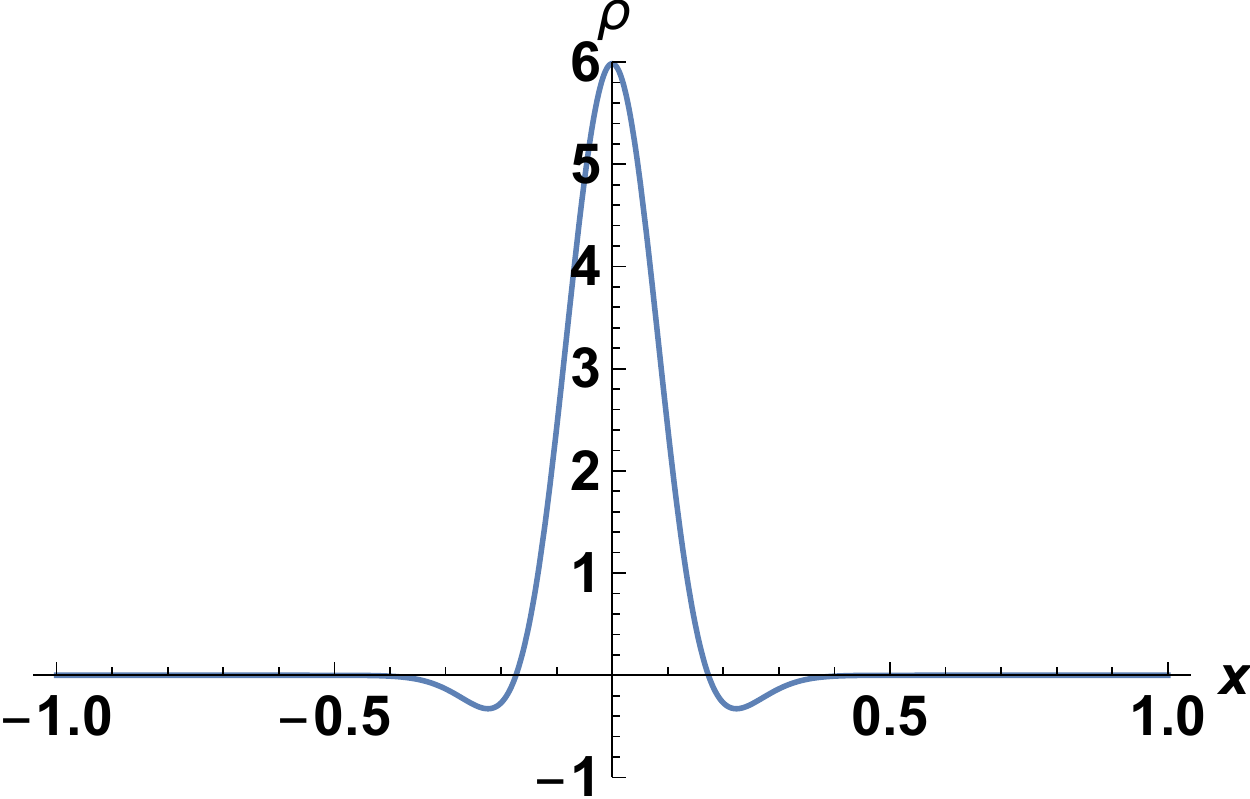|\hfil}
		\hbox to\hsize{\hfil (a)\hfil (b)\hfil}
		\bigskip
		\hbox to\hsize{\hfil\|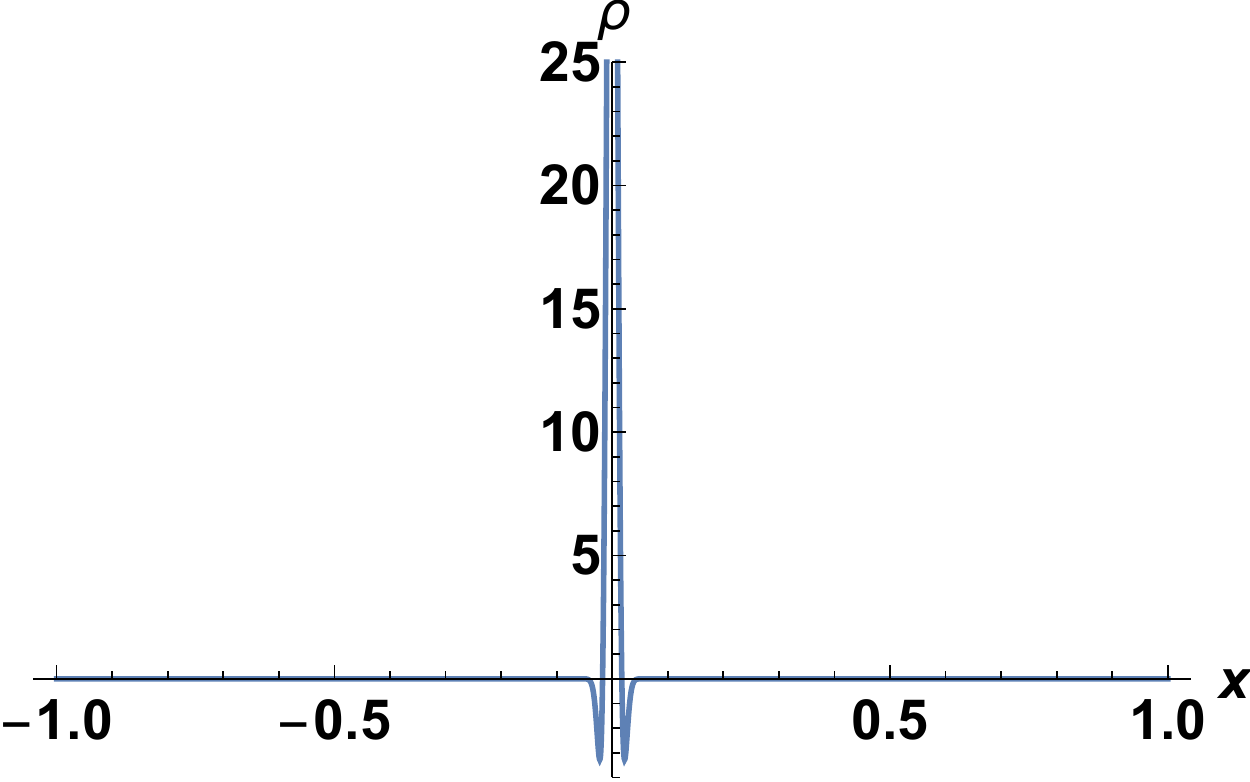|\hfil\|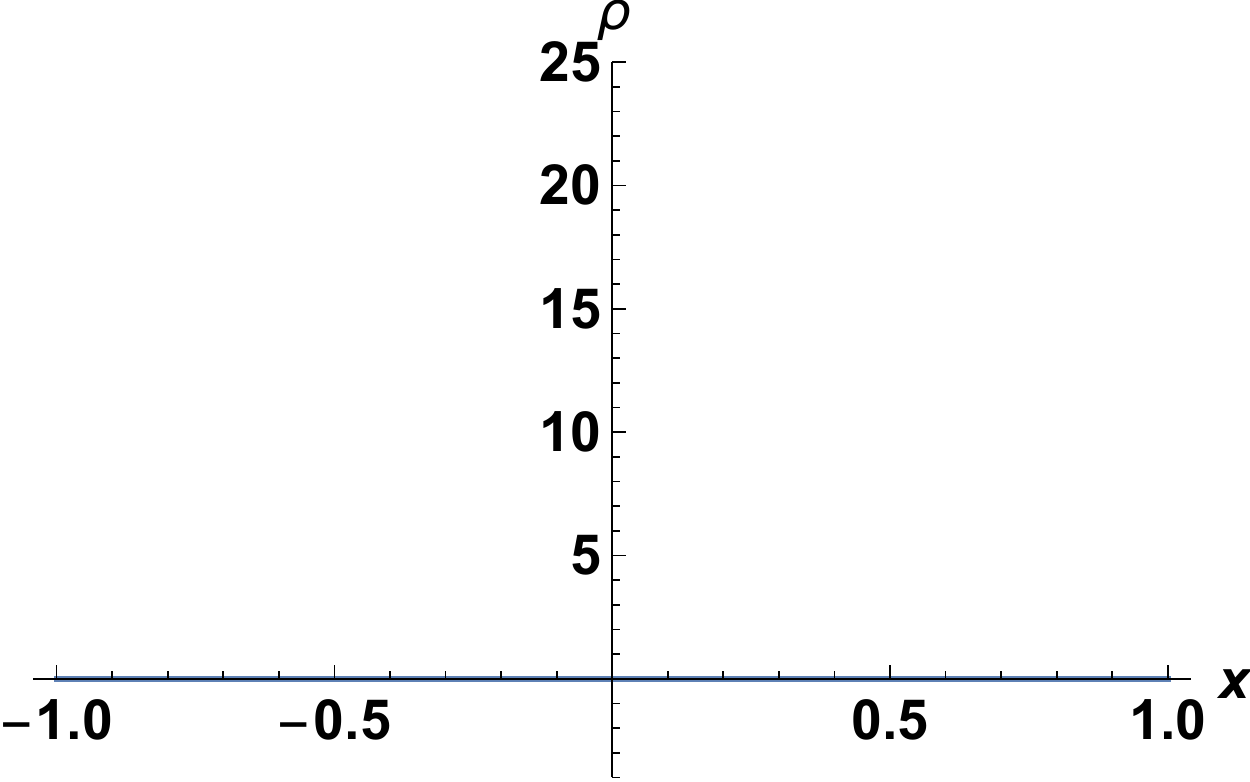|\hfil}
		\hbox to\hsize{\hfil (c)\hfil (d)\hfil}
		\caption{Plots of $\rho_{\sigma}$ when $m=1$ for various values of $\sigma$:
		$\sigma=1$ in (a), $\sigma=0.1$ in (b), $\sigma=0.01$ in (c) and $\sigma=0.001$ in (d).} 
	\label{Gauss}
\end{figure}

$$\lim_{\sigma\to 0}\int_{\vert x\vert \ge\delta} \vert\rho_\sigma(x)\vert
\, dx=
\lim_{\sigma\to 0}\sum_{n=1}^m{\sigma^{2n}\over 2^n n!} 
\int_{\vert x\vert \ge\delta} \left\vert
{d^{2n} \over dx^{2n}}G_\sigma(x)\right\vert\, dx;
$$
notice that the sum starts at $n=1$ because, since the
gaussian is a mollifier, we already known that
$\lim_{\sigma\to 0}\int_{\vert x\vert \ge\delta} e^{-x^2/2\sigma^2}/
\sqrt{2\pi \sigma^2}\, dx=0$.
Now, let us consider the Hermite polynomial:
$$H_n(x)=(-1)^n e^{x^2}{d^n e^{-x^2}\over dx^n},$$
which satisfies
$${d^n G_\sigma(x)\over dx^n}=(-1)^n {H_n(x/\sqrt{2}\sigma)\over
(\sqrt{2}\sigma)^n} G_\sigma(x).$$
If we change the variable $y=x/\sigma$, we have
\[\begin{split}
\sigma^{2n}\int_{\vert x\vert \ge\delta}
\left\vert{d^{2n} \over dx^{2n}}G_\sigma(x)\right\vert\, dx
&={1\over 2^n}
\int_{\vert x\vert \ge\delta}\left\vert H_{2n}(x/\sqrt{2}\sigma)
G_\sigma(x)\right\vert\,dx\\
&={1\over 2^n}
\int_{\vert y\vert \ge\delta/\sigma}\left\vert H_{2n}(y/\sqrt{2}) G_1(y)\right\vert\,dy,\end{split}\]
that goes to $0$ as $\sigma\to0$

To sum up, we can state the following lemma for $\rho_\sigma$.

\begin{lemma}
The family of functions
$\{\rho_\sigma\}_{\sigma>0}
\subset C^\infty(\bbR)$ has the following properties
\medskip
\item{\it i. } $\sup_{x\in \bbR} \vert \rho_\sigma(x)\vert <\infty$;
\item{\it ii. } $\int_{\bbR} \rho_\sigma(x)\, dx=1$;
\item{\it iii. } $\lim_{\sigma\to 0}\int_{\vert x\vert \ge\delta} \vert\rho_\sigma(x)\vert\, dx=0$ for every $\delta>0$.
\item{\it iv.} $\int_{\bbR} |\rho_\sigma(x)|\, dx<\infty$
\end{lemma}

\medskip\noindent
In view of this Lemma the following theorem holds:

\begin{theorem}
The sequence  $\langle\rho_\sigma\rangle$
converges in ${\cal D}'$ to the $\delta$ function as $\sigma\to 0$:
$$\lim_{\sigma\to0}\int \rho_\sigma(x) \varphi(x)\, dx=\varphi(0),
\qquad \forall \varphi\in C^\infty_0(\bbR).$$
\end{theorem}

\begin{proof}
Because of {\it ii.} we set
$$I_\sigma:=\int\rho_\sigma(x)\varphi(x)\, dx-\varphi(0)
=\int\rho_\sigma(x)\bigl(\varphi(x)-\varphi(0)\bigr)\, dx,$$
so that for any fixed $\delta>0$:
$$I_\sigma=\int_{\vert x\vert\le \delta}
\rho_\sigma(x)\bigl(\varphi(x)-\varphi(0)\bigr)\, dx+
\int_{\vert x\vert>\delta}
\rho_\sigma(x)\bigl(\varphi(x)-\varphi(0)\bigr)\, dx;$$
if we let $\alpha_\sigma^\delta$ be the first integral, and
$\beta_\sigma^\delta$ the second one, we have the following bounds
\[\begin{split}
&\vert\alpha_\sigma^\delta\vert\le
\sup_{\vert x\vert\le\delta}\vert\varphi(x)-\varphi(0)\vert
\int_{\vert x\vert\le \delta}\vert \rho_\sigma(x)\vert\, dx
\le K\sup_{\vert x\vert\le\delta}\vert\varphi(x)-\varphi(0)\vert\equiv
A_\delta,\\
&\vert\beta_\sigma^\delta\vert
\le \sup_{\vert x\vert>\delta}\vert\varphi(x)-\varphi(0)\vert
\int_{\vert x\vert> \delta}\vert \rho_\sigma(x)\vert\, dx
\le 2\Vert\varphi\Vert_{L^\infty(\bbR)}
\int_{\vert x\vert >\delta} \vert\rho_\sigma(x)\vert\, dx.
\end{split}\]
Now because of the continuity in $x=0$ of $\varphi$ and because of
property {\it iii.} of Lemma~A we have that $\lim_{\delta\to0} A_\delta=0$, and
$\lim_{\sigma\to0}\beta_\sigma^\delta=0$ for every $\delta$ positive. Then
$$\maxlim_{\sigma\to 0} \vert I_\sigma\vert\le
A_\delta+\maxlim_{\sigma\to 0} \vert\beta_\sigma^\delta\vert=A_\delta,$$
now taking the limit $\delta\to 0$ we finally obtain $\lim_\sigma\to0
\vert I_\sigma\vert=0$.
\end{proof}

%% file: vect.tex
\label{vectorization-app}
On the discrete retina the activations assume the form (we do not
explicitly write the time dependence on the time)
\[
  A_{ix_1x_2}= \varphi_{ij\xi_1\xi_2}
  C_{j(x_1-\xi_1)(x_2-\xi_2)}.
  \]
Now let us define $\gamma^x=(C_{1(x_1-1)(x_2-1)},C_{1(x_1-1)(x_2-2)},\dots
C_{m(x_1-\ell)(x_2-\ell)})$ and $\chi^i:=(\varphi_{i111},\varphi_{i112},\dots,
\varphi_{im\ell\ell})$. Then 
\[A_{ix}\equiv A_{i x_1 x_2}=\chi^i_\alpha \gamma^x_\alpha\]

First of all let us analyze the motion-invariance term. If we let $\zeta^x$
be the vector that for each pixel $x$ on the retina collects the
components of the discretization of the term
$v(x,t)\cdot \nabla_xC_j(x-\xi,t)$ with respect to the indexes $\xi$ and  $j$, 
then the part of the Lagrangian relative
to the motion invariance term can be written as
\[\frac{1}{2}\int_0^T h(t) g_x\Bigl(
(\chi^i_\alpha\gamma^x_\alpha)\,\dot{}\,+\chi_\alpha^i\zeta_\alpha^x
\Bigr)^2.\]
The square in the previous equation, once expanded,  gives:
\begin{equation}\begin{split}
g_x\Bigl(
(\chi^i_\alpha\gamma^x_\alpha)\,\dot{}\,+\chi_\alpha^i\zeta_\alpha^x
\Bigr)^2=&
\chi^i_\alpha\bigl(g_x(\dot \gamma^x_\alpha\dot\gamma^x_\beta
+\zeta^x_\alpha\zeta^x_\beta +2\dot\gamma^x_\alpha\zeta^x_\beta)\bigr)
\delta_{ij}\chi^j_\beta\\
&+2\chi^i_\alpha\bigl(g_x(\dot \gamma^x_\alpha\gamma^x_\beta
+\zeta^x_\alpha\gamma^x_\beta)\bigr) \delta_{ij}\dot\chi^j_\beta\\
&+\dot\chi^i_\alpha\bigl(g_x\gamma^x_\alpha\gamma^x_\beta
\bigr) \delta_{ij}\dot\chi^j_\beta\\
=&\chi^i_\alpha O_{\alpha \beta}
\delta_{ij}\chi^j_\beta
+2\chi^i_\alpha N_{\alpha\beta}\delta_{ij}\dot\chi^j_\beta
+\dot\chi^i_\alpha M_{\alpha\beta}\delta_{ij}\dot\chi^j_\beta.
\end{split}\label{MI-lin}
\end{equation}
Where we have defined
$O_{\alpha\beta}:=\bigl(g_x(\dot \gamma^x_\alpha\dot\gamma^x_\beta
+\zeta^x_\alpha\zeta^x_\beta +2\dot\gamma^x_\alpha\zeta^x_\beta)\bigr)$,
$N_{\alpha\beta}:=\bigl(g_x(\dot \gamma^x_\alpha\gamma^x_\beta
+\zeta^x_\alpha\gamma^x_\beta)\bigr)$ and
$M_{\alpha\beta}:=\dot\chi^i_\alpha\bigl(g_x\gamma^x_\alpha\gamma^x_\beta
\bigr) \delta_{ij}\dot\chi^j_\beta$.

Given $A\in \R^{m\times n}$ and $B\in\R^{n\times k}$ and having defined
the vectorization operation as follows
\[\Vec(A)=(a_{11}, a_{21},\dots, a_{m1},a_{m2},a_{12},a_{22},
\dots, a_{mn})',\]
this two identities holds
\begin{enumerate}
\item $\Vec(A B)=(B'\otimes I_m)\Vec(A)$;
\item $\Tr(A'B)=\Vec(A)\cdot \Vec(B)$.
\end{enumerate}
Using these two identities we can rewrite the terms in Eq.~(\ref{MI-lin})
as follows:
\[
\begin{split}
&\chi^i_\alpha O_{\alpha \beta}
\delta_{ij}\chi^j_\beta=\Tr(\chi'\chi O)=\Vec(\chi)\cdot\Vec(\chi O)=
\Vec(\chi)\cdot(O'\otimes I_{m\ell^2})\Vec(\chi);
\\
&\chi^i_\alpha N_{\alpha\beta}\delta_{ij}\dot\chi^j_\beta
=\Tr(\chi'\dot \chi N)=\Vec(\chi)\cdot\Vec(\dot\chi N)=
\Vec(\chi)\cdot(N\otimes I_{m\ell^2})\Vec(\dot\chi);\\
&\dot\chi^i_\alpha M_{\alpha\beta}\delta_{ij}\dot\chi^j_\beta=
\Tr(\dot\chi'\dot\chi M)=\Vec(\dot\chi)\cdot\Vec(\dot\chi M)=
\Vec(\dot\chi)\cdot(M'\otimes I_{m\ell^2})\Vec(\dot\chi).
\end{split}
\]
Once we define  $q:=\Vec(\chi)$,
$O^\natural:=(O'\otimes I_{m\ell^2})$,
$N^\natural:=(N'\otimes I_{m\ell^2})$ and $M^\natural:=(M'\otimes I_{m\ell^2})$
we eventually have
\[\chi^i_\alpha O_{\alpha \beta}
\delta_{ij}\chi^j_\beta=q\cdot O^\natural q,\quad
\chi^i_\alpha N_{\alpha\beta}\delta_{ij}\dot\chi^j_\beta=
q\cdot N^\natural \dot q,\quad
\dot\chi^i_\alpha M_{\alpha\beta}\delta_{ij}\dot\chi^j_\beta=
\dot q\cdot M^\natural
\dot q.\]

The form of $U(q,C)$ is instead straightforward, it follows directly from the
definition.
